\documentclass[runningheads]{llncs}

\usepackage[mobile]{eccv}

\usepackage{eccvabbrv}

\usepackage{graphicx}
\usepackage{booktabs}
\usepackage{array}
\usepackage[accsupp]{axessibility}  

\usepackage[pagebackref,breaklinks,colorlinks,citecolor=eccvblue]{hyperref}

\usepackage{orcidlink}

\begin{document}

\title{iComMa: Inverting 3D Gaussian Splatting for Camera Pose Estimation via Comparing and Matching} 

\titlerunning{iComMa}

\author{Yuan Sun\inst{1,\dagger} \and
Xuan Wang\inst{2,\dagger} \and
Yunfan Zhang\inst{1} \and Jie Zhang\inst{1} \and Caigui Jiang\inst{1} \and \\ Yu Guo\inst{1,*} \and Fei Wang\inst{1} \thanks{$^{\dagger}$Contribute equally. $^{*}$Corresponding author.}}

\authorrunning{Y. Sun et al.}

\institute{Xi'an Jiaotong University \and
Ant Group
}

\maketitle

\begin{abstract}
We present a method named iComMa to address the 6D camera pose estimation problem in computer vision. Conventional pose estimation methods typically rely on the target's CAD model or necessitate specific network training tailored to particular object classes. Some existing methods have achieved promising results in mesh-free object and scene pose estimation by inverting the Neural Radiance Fields (NeRF). However, they still struggle with adverse initializations such as large rotations and translations. To address this issue, we propose an efficient method for accurate camera pose estimation by inverting 3D Gaussian Splatting (3DGS). 
Specifically, a gradient-based differentiable framework optimizes camera pose by minimizing the residual between the query image and the rendered image, requiring no training. An end-to-end matching module is designed to enhance the model's robustness against adverse initializations, while minimizing pixel-level comparing loss aids in precise pose estimation. Experimental results on synthetic and complex real-world data demonstrate the effectiveness of the proposed approach in challenging conditions and the accuracy of camera pose estimation.
\end{abstract}   
\begin{figure}[t]
    \centering
    \includegraphics[width=\textwidth]{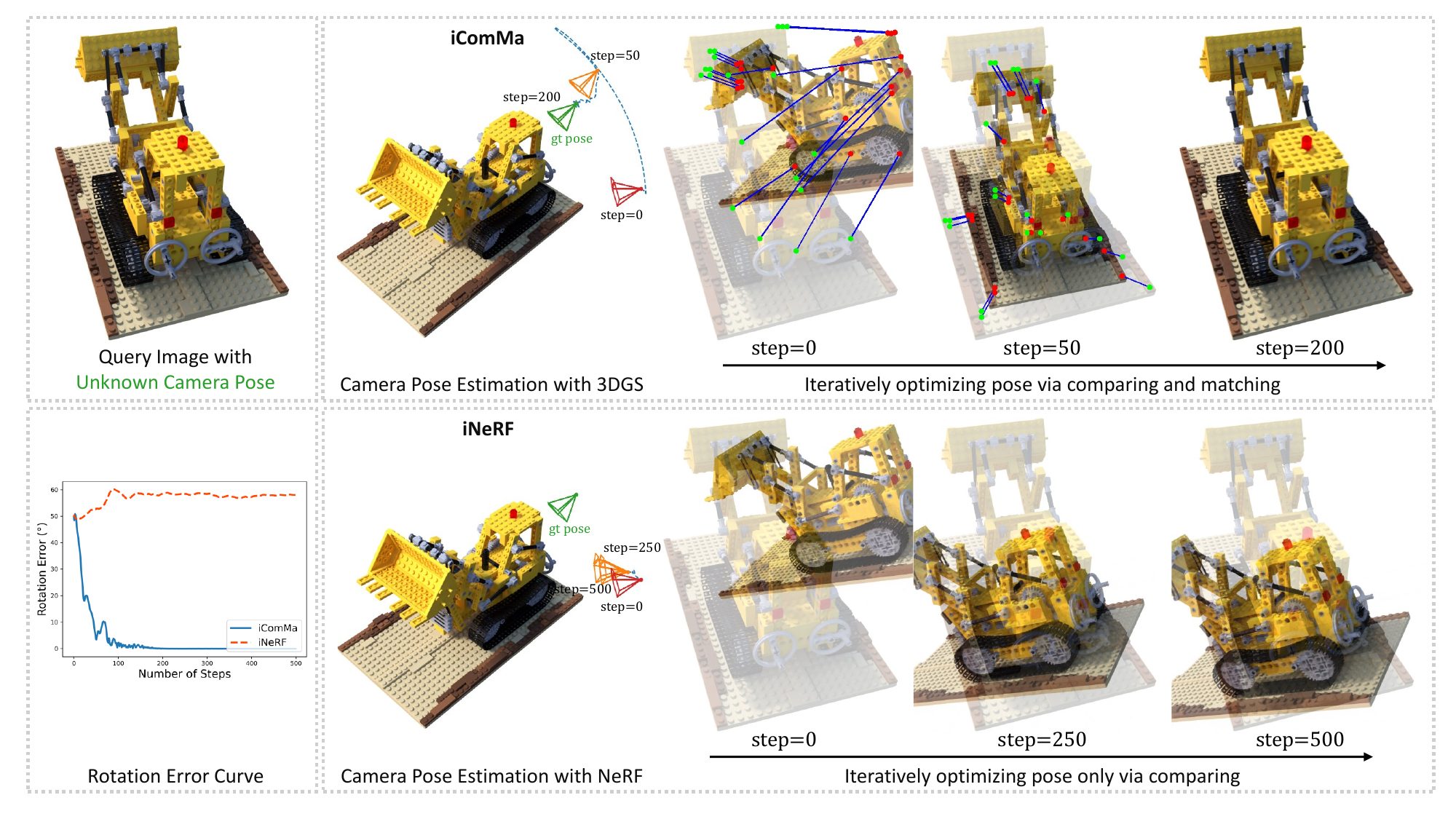}
    \caption{ Given a query image with an \textcolor[RGB]{44, 160, 44}{unknown camera pose}, iComMa accurately estimates it by inverting 3D Gaussian Splatting from a \textcolor[RGB]{214, 39, 40}{known initial pose} (step=0). The gradient information inherent in the differences between the query image and the rendered image (which are  overlaid in the above figure, with a higher degree of overlap indicating more accurate pose estimation) is utilized for iteratively optimizing the camera pose. Compared to iNeRF\cite{yen2021inerf}, the proposed method not only employs pixel-to-pixel comparing but also utilizes 2D keypoints matching, which are connected by \textcolor[RGB]{0, 0, 255}{blue}  lines in the above figure. As a result, our method is capable of precisely estimating camera poses even under poor initial conditions, such as large angular deviations.}
    \label{fig_teaser}
\end{figure}

\section{Introduction}
\label{sec:intro}

Six-degree-of-freedom (6DoF) pose estimation is crucial in various areas like robotics, Simultaneous Localization and Mapping (SLAM), augmented reality, and virtual reality. Common pose estimation methods often rely on detailed geometric models related to the target object\cite{labbe2020cosypose,peng2019pvnet,tekin2018real,xiang2017posecnn}. In recent years, there has been a shift in scholarly focus towards category-level pose estimation\cite{ahmadyan2021objectron,chen2020category,lin2022keypoint,lin2022single}. This change aims to reduce dependence on CAD models for target objects or scenes, leading to notable advancements. Category-level pose estimation involves learning from various instances within a category to find similarities in appearance and shape. Despite their effectiveness, these methods require separate data collection and training for each category, adding complexity to pose estimation. Some approaches also use traditional Perspective-n-Point (PnP) algorithms, combining them with neural networks to match three-dimensional and two-dimensional points\cite{sarlin2019coarse,sun2022onepose,he2022onepose++,shugurov2022osop}, ultimately determining relative poses. The accuracy of these methods depends on the effectiveness of the matching process and a precise understanding of the point cloud. The accuracy of the correspondence between 2D and 3D information significantly influences the capabilities of pose estimation models.

Represented  by Neural Radiance Fields (NeRF)\cite{wang2021nerf}, the application of differentiable rendering methods to articulate three-dimensional scenes has evinced substantial efficacy. The concept of target pose estimation grounded in this paradigm has garnered noteworthy scholarly attention\cite{yen2021inerf,li2023nerf,lin2021barf}. These methods utilize neural radiance fields to articulate spatial information in three dimensions. Subsequently, they prognosticate the object's camera pose by establishing correspondences between the two-dimensional image and the implicit representation of the neural field, or by discerning disparities between the rendered image and the query image. In contradistinction to conventional matching-based methods, this approach iteratively refines the pose based on the extant three-dimensional scene representation, obviating the necessity for supplementary network training tailored to a singular target category. This iterative refinement culminates in heightened accuracy and efficacy within certain scenarios. However, its applicability is markedly circumscribed by initial conditions. In arduous scenarios, such as significant angular deviations, generating precise gradient information through local appearance comparisons is difficult, thereby impeding effective pose estimation.

Our objective is to devise an easily applicable RGB-only approach for pose estimation, relying solely on gradient information without the need for network training. To achieve this, we employ 3D Gaussian Splatting \cite{kerbl20233d} for initial scene reconstruction requiring only images with pose information as input without any 3D geometric models. Additionally, we explore how to complete the entire camera pose optimization process under the 3D Gaussian expression. We extensively assess the strengths and limitations of render-and-compare and matching methods, with the aim of proposing a more rational framework capable of maintaining precise pose estimation even in challenging scenarios. Specifically, we compute both comparing and matching losses between the image rendered from the current pose via Gaussian Splatting and the query image. The former involves pixel-to-pixel comparisons, enabling fine-grained pose predictions through iterative optimization. This portion of backpropagation does not involve neural networks, providing a speed advantage. Acknowledging the advantages of traditional pose estimation matching methods in addressing complex scenarios, we design an end-to-end matching module that penalizes inaccurate pose predictions based on the correspondence between 2D keypoints. Its differentiability ensures seamless integration into our method. 

In summary, we present a novel end-to-end framework that combines two paradigms of image-based camera pose estimation: the render-and-compare strategy and feature matching methods.  Experiments demonstrate that our method  reaches  the best of both worlds. It achieves both reduced reliance on good initialization in render-and-compare approaches  and enhanced robustness in handling noisy feature detection and poorly-textured scenarios compared to the feature matching methods.  This enables it to converge rapidly to more accurate results even when there is a large bias in the initialization. Both qualitative and quantitative evaluations illustrate that the proposed  iComMa outperforms the state-of-the-art techniques significantly. Simultaneously, inheriting the efficient rendering pipeline from 3D  Gaussian Splatting allows our method surpass the baseline approach (iNeRF) in time efficiency by almost tenfold.

\section{Related Work}
\label{sec:formatting}
\subsection{Matching-Based Pose Estimation}

The pose estimation methods based on feature matching are prevalent techniques in the field of computer vision\cite{collet2009object,martinez2010moped,tang2012textured,li2018deepim,goodwin2022zero,fan2023pope}. These methods rely on matching identical feature points or feature descriptors between different image frames or 3D models, followed by utilizing these matched points to calculate the camera's pose. Initially, this process required manually designing matching features. In recent years, remarkable progress has been achieved by introducing neural networks for feature extraction\cite{ni2023pats,chen2022aspanformer,jiang2021cotr}. Architectures such as SuperGlue\cite{sarlin2020superglue}, LoFTR\cite{sun2021loftr}, and MatchFormer\cite{wang2022matchformer} leverage distinct Transformer structures, meticulously considering the global information of images and the potential correspondences between image pairs, resulting in significant matching outcomes. Subsequently, the PnP-RANSAC method is employed to compute relative camera poses, yielding impressive estimation results. However, these methods predominantly focus on 2D-2D matching relationships and fail to fully exploit the three-dimensional information of the scene.\par

Moreover, there are methods that concentrate on matching between images and target point clouds or 3D models\cite{castro2023posematcher,goodwin2022zero,fan2023pope,sun2022onepose,he2022onepose++}, achieving notable outcomes. Nevertheless, these methods depend on dense point cloud information or other high-quality target models and are subject to the effectiveness of the matching process, thus exhibiting certain limitations in achieving precise pose estimation.

\subsection{Pose Estimation with Neural Radiance Fields}

In recent years, approaches based on Neural Radiance Fields (NeRF) \cite{wang2021nerf} have demonstrated substantial advancements in representing three-dimensional scenes \cite{hedman2018deep, bian2023nope, fridovich2022plenoxels, xu2022point, barron2022mip}. By capitalizing on the distinctions between rendered and real images, these methodologies train neural networks to articulate the color and density information of the target scene as a function of spatial position within the scene, thereby achieving exceptional expressive capabilities for intricate three-dimensional scenes. Several endeavors leverage this paradigm to undertake tasks related to pose estimation and Simultaneous Localization and Mapping (SLAM) \cite{labbe2022megapose, huang2022neural, chng2022garf, sucar2021imap, zhu2022nice}. The iNeRF method \cite{yen2021inerf} employs the assumed camera pose to generate an image through rendering, subsequently comparing pixel differences with the query image. It then utilizes the acquired gradient information to iteratively refine the current camera pose until the rendered image aligns with the query. Nerf-pose \cite{li2023nerf} adopts NeRF's implicit representation of three-dimensional scenes and trains a pose regression network to establish correspondences between 2D and 3D. Nerfels \cite{avraham2022nerfels} constructs a locally dense and globally sparse three-dimensional scene representation via local modeling of feature parts, subsequently employing it for pose estimation tasks. While these methodologies achieve precise pose estimation through pixel-level comparative losses, they face challenges in convergence within complex and demanding scenarios, particularly when a significant mismatch exists between rendered and query images, thereby impeding accurate pose estimation.

\begin{figure*}
    \centering
    \includegraphics[width=\textwidth]{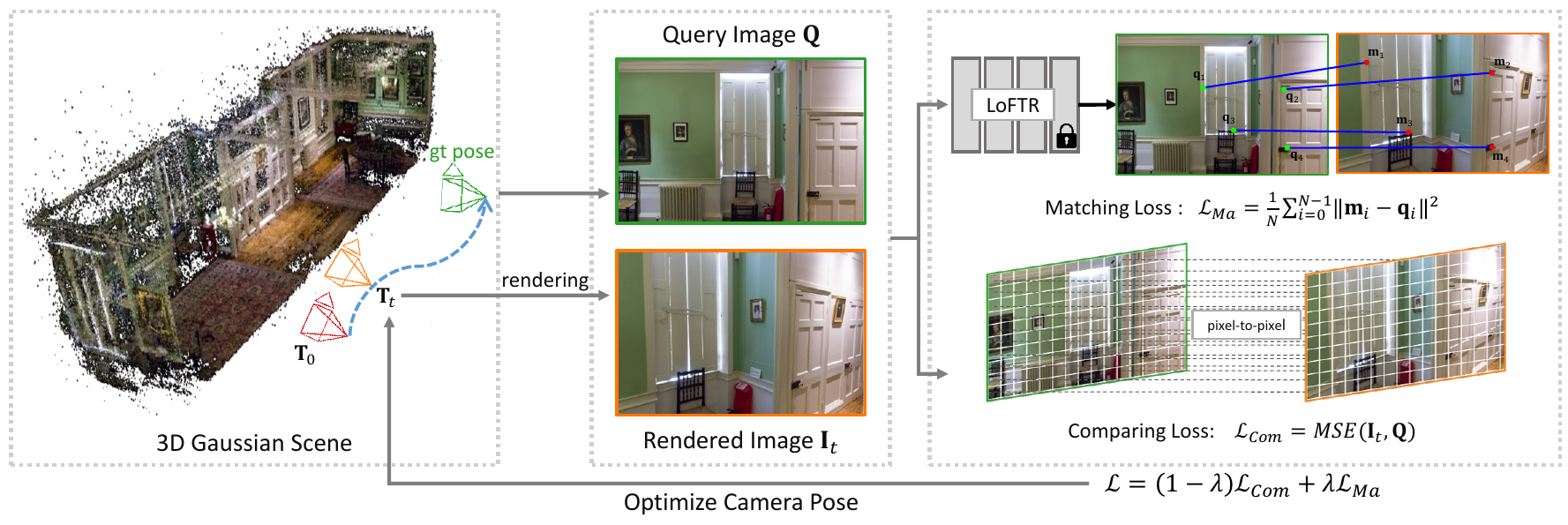}
    \caption{\textbf{Overview of iComMa.} Given an \textcolor[RGB]{214, 39, 40}{initial camera pose}, iComMa iteratively optimizes to estimate the \textcolor[RGB]{44, 160, 44}{ground truth pose} associated with the query image. For the $t$-th optimization step, we first render the image corresponding to the camera pose $\mathbf{T}_t$ using 3D Gaussian Splatting. Subsequently, we compute the residuals between the rendered image and the query image, which include the matching loss $\mathcal{L}_{Ma}$ obtained from the end-to-end matching module and the per-pixel comparing loss $\mathcal{L}_{Com}$. The entire framework is differentiable, enabling the optimization of camera poses by utilizing the gradients derived from minimizing the residuals.}
   \label{fig:overview}
\end{figure*}
\section{Method}

An overview of our approach is depicted in \cref{fig:overview}. Given a trained 3D representation of the scene, which are parameterized by $\mathrm{\Theta}$, and a known camera pose, our objective is to estimate the camera pose $\mathbf{\mathbf{T} } $ on an query observation $\mathbf{Q} $. We formulate the problem as follows:
\begin{equation}
\hat{\mathbf{T} } =\mathop{\arg\min}\limits_{\mathbf{T} \in SE(3)}\mathcal{L\mathrm{(\mathbf{T} |\mathbf{Q},\Theta } )}.  \label{eq}   
\end{equation}

To achieve our objective, we adopt the 3D Gaussian as the representation of the three-dimensional scene, owing to its exceptional performance in both expressiveness and rendering speed. Given the estimated camera pose $\mathbf{T} \in SE(3) $, we utilized a fixed 3D Gaussian $\mathcal{R} $ to render a corresponding image observation.  
By incorporating the geometric coordinates and pixel disparities between the rendered and query images, our approach demonstrates robustness even under challenging initial conditions. We introduce a matching loss function denoted as $\mathcal{L}_{Ma}$, which incrementally optimizes the camera pose for the target scene through image matching. This is beneficial for addressing complex scenes, such as when the initial camera rotation angle is extremely large. Additionally, we employ a rendering-based comparing loss named $\mathcal{L}_{Com}$ to ensure that, in the final stages of pose optimization, our method attains a high level of accuracy. The iterative update of the camera pose $\mathbf{\mathbf{T} } $ by minimize the combined loss $\mathcal{L}$ yields remarkable outcomes. This joint optimization framework provides a potent solution to the pose estimation problem, ensuring both accuracy and robustness in complex scenarios.

\subsection{Camera Pose Optimization by Inverting 3D Gaussian Splatting}

We employ the 3D Gaussian function as the representation of the scene, benefiting from its differentiability and efficient rasterization when projected onto a 2D plane, facilitating rapid $\alpha$-blending for rendering purposes. The fast rasterization of 3DGS profits from its utilization of CUDA and explicit parameter derivative computations. Similarly, we also derive the gradient information of the camera pose $\mathbf{T}$ explicitly. Specifically, the camera pose is passed as input into the 3D Gaussian backward process and participates in the correlation operations of the transformation matrix and the Gaussian 2D covariance matrix. Due to the differentiability of the entire process, according to the chain rule, we can obtain gradient information of the loss function with respect to the camera pose. Please refer to the supplementary materials for specific details.

To ensure that each iterative step's $\mathbf{T}_t$ remains within SE(3), inspired by iNerf \cite{yen2021inerf}, we parameterize the variation in camera pose as lie algebra $\tau \in se(3)$ with six degrees of freedom. Given the initial camera pose $\mathbf{T}_0$, the camera pose at the $t$-th iteration $\mathbf{T}_t=f(\tau) \mathbf{T}_0$, where $f(\cdot)$ transforms the lie algebra $se(3)$ to the special euclidean group $SE(3)$.

The overall loss function, composed of comparing loss $\mathcal{L}_{Com}$ and matching loss $\mathcal{L}_{Ma}$, is defined as follows:
\begin{equation}
\mathcal{L}=(1-\lambda) \mathcal{L}_{Com} + \lambda \mathcal{L}_{Ma},
\end{equation}
where $\lambda$ is a balancing coefficient belonging to the interval $\left[0, 1\right]$. In the following two subsections, we will introduce the detailed designs of the two components of the loss function.

\subsection{Pixel-to-Pixel Comparing Loss for Pose Estimation}
Direct pixel-level comparison between images emerges as a potent tool for detecting subtle variances in camera poses. Such an approach strives for a more refined level of pose prediction. We introduce a loss function defined as follows:
\begin{equation}
\mathcal{L}_{Com} = MSE(\mathbf{I}_t, \mathbf{Q}),
\end{equation}
where \( MSE \) denotes the Mean Squared Error, and \( \mathbf{I}_t \) and \( \mathbf{Q} \) respectively represent the rendered image corresponding to the current pose $\mathbf{T}_t$ and the query image. By aligning at the pixel level, fine-grained prediction of pose is achieved.

\subsection{End-to-end Matching Loss for Pose Estimation}

Traditional rendering-based methods often encounter difficulties in certain scenarios, such as excessively large rotation angles or significant translation biases. Moreover, these methods typically exhibit a reduced performance on real-world data. To address these issues, we introduce an end-to-end matching loss, $\mathcal{L}_{Ma}$, which uses the Euclidean distance between matching points as a metric to quantify the difference between two poses. Unlike conventional rendering methods, our matching loss directly measures images based on geometric positions, aligning more closely with the fundamental principles of pose calculation. Intuitively, the greater the similarity in the positions of the matching points across two images, the more closely aligned the corresponding poses are. 

When obtaining the image rendered via 3D Gaussian splatting at the current optimized pose, we employ LoFTR \cite{sun2021loftr}, a detector-free local feature matching method, to identify the corresponding feature points between the rendered and query images. Under predefined confidence conditions, LoFTR identifies matching point pairs \(\mathbf{\left\{m, q\right\}}\) between the two images, along with their pixel coordinates. It is important to note that these pixel coordinates are expressed as normalized coordinates relative to the dimensions of the respective images. The camera pose is optimized by employing an $\ell_2$ loss on the coordinates of each point, until the discrepancies in the 2D positions of matching points become sufficiently small:
\begin{equation}
\mathcal{L}_{Ma} = \frac{1}{N} \sum_{i=0}^{N-1} \left\| \mathbf{m}_{i} - \mathbf{q}_{i} \right\|^{2}, \label{eq}
\end{equation}
where \(\mathbf{m}_{i}\) and \(\mathbf{q}_{i}\) represent points in the set of matching pairs, and \(N\) is the number of such pairs. By directly measuring the positions of these matching points on their respective images, we anticipate effective completion of pose estimation, even under challenging initial settings or in complex scenes.

\subsection{Implementation Details}
\subsubsection{Optimization Strategy:}
The backpropagation of matching loss, due to the involvement of neural networks, affects the speed of pose estimation. To expedite the optimization process, we partition it into two stages. In the initial stage of optimization, our loss function is denoted by $\mathcal{L}$, aiming to optimize the pose to an approximate accuracy by incorporating gradient information generated from matching points. Intuitively, a smaller difference in 2D matching points implies a smaller difference in poses. Thus, when the $\mathcal{L}_{Ma}$ is less than a threshold value $\epsilon$, we deactivate the matching module and only utilize the comparing loss to make final adjustments to the camera pose. By the way, due to the normalization process applied to $\mathcal{L}_{Ma}$, setting the threshold is straightforward. Empirically, $\epsilon=0.001$ is applicable to all types of datasets.
\subsubsection{Optimizer and Learning Rate Schedule:} The Adam optimizer is employed, with hyperparameters $\beta_1$ and $\beta_2$ fixed at 0.9 and 0.999, respectively. The initial learning rate $\alpha_0$ is set to 0.05. Particularly, during the initial stage of pose optimization, we maintain this relatively high learning rate unchanged, aiming to swiftly optimize a pose with minimal error. Subsequently, when estimating the pose solely using the render-and-compare strategy, we gradually decrease the learning rate to ensure precise pose estimation. 
Specifically, at step $t$ during this stage of optimization, the learning rate is as follows:
\begin{equation}
\alpha_t=\alpha_0 0.6^{t / 50}.
\end{equation}

\section{Experiment}

In this section, we substantiate the superiority of our proposed method, particularly under challenging conditions, through quantitative experiments comparing against iNeRF and matching-based pose estimation methods. Furthermore, ablation experiments provide an intuitive demonstration of the efficacy of the iComMa components.
 
\subsection{Experimental Setup}
\subsubsection{Datasets:} In order to comprehensively evaluate the performance of the proposed method across various scenarios, a diverse array of datasets is employed. Specifically, eight NeRF's synthetic datasets \cite{wang2021nerf}, characterized by generated single-object scenes, are utilized. Additionally, eight LLFF-type datasets \cite{wang2021nerf} are incorporated, featuring a limited number of scene images captured using a forward-facing handheld cellphone. Furthermore, nine 360-degree unbounded scene datasets are obtained from Mip-NeRF360 \cite{barron2022mip}, along with scenes from \textit{Playroom} and \textit{DrJohnson} sourced from the Deep Blending dataset \cite{hedman2018deep}, representing $360^\circ$ scene datasets. In LLFF datasets and  $360^\circ$ scene datasets, individual instances exhibit considerable variations in translation scale. For ease of reporting, following NeRF\cite{wang2021nerf} and Mip-Nerf360\cite{barron2022mip}, we normalize the translation component using the boundary information. Consequently, in subsequent experimental results, the translation component is represented solely by values without units.

\subsubsection{Baseline Methods:} In Section \ref{sec_cwinerf}, we conduct comparative experiments with the render-and-compare method, iNeRF. It is noteworthy that due to the fact that iNeRF, based on the basic NeRF\cite{wang2021nerf} model, cannot effectively reconstruct $360^\circ$ scenes, we refer to the intrinsic mechanism of iNeRF and perform pose estimation by inverting Mip-Nerf360\cite{barron2022mip}. This variant is denoted as $\mathrm{iNeRF}^{\dagger}$. In Section \ref{sec_rpe}, we quantitatively compare our proposed method with several matching-based pose estimation methods, namely LightGlue \cite{lindenberger2023lightglue}, MatchFormer \cite{wang2022matchformer}, and LoFTR \cite{sun2021loftr}. 
These three methods all first obtain the matching point pairs between two images and then compute the essential matrix using the RANSAC method to obtain the relative camera poses.

\subsubsection{Metrics:} In the pose estimation community, a common metric is to report the proportion of correctly predicted poses, where a correctly predicted pose is defined as having rotation and translation errors below certain thresholds \cite{yen2021inerf,sun2022onepose}. In Section \ref{sec_rpe}, we follow previous work \cite{sun2021loftr,lindenberger2023lightglue} by using the Area Under the Curve (AUC) to measure algorithm performance.
\begin{figure*}[h!]
\centering
\setlength{\tabcolsep}{1pt} 
\begin{tabular}{ccc}

\includegraphics[width=0.31\linewidth]{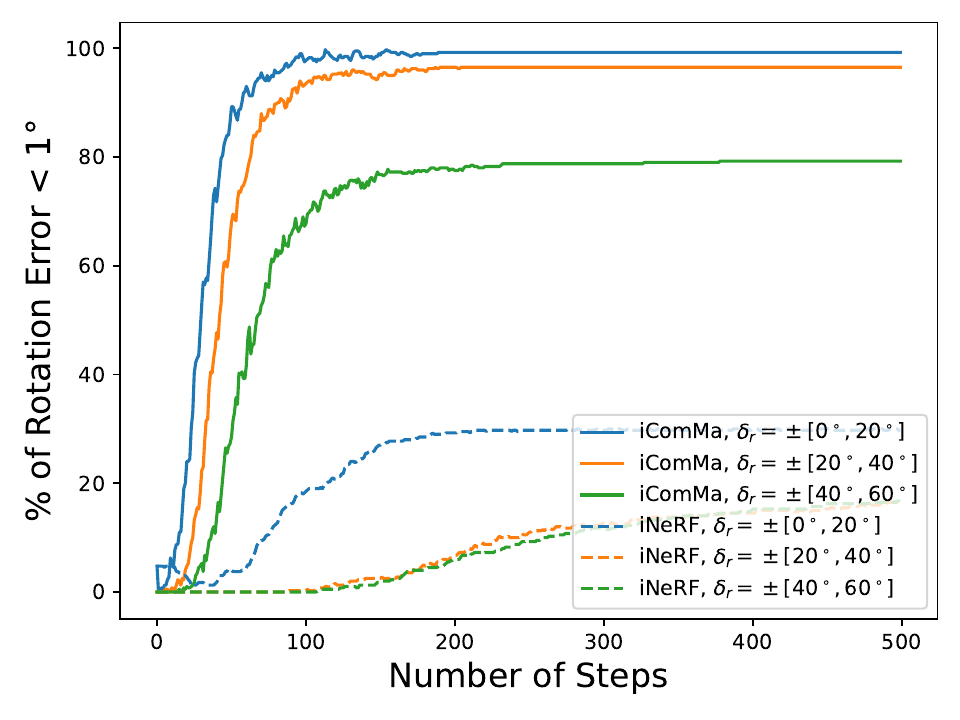} & 
\includegraphics[width=0.31\linewidth]{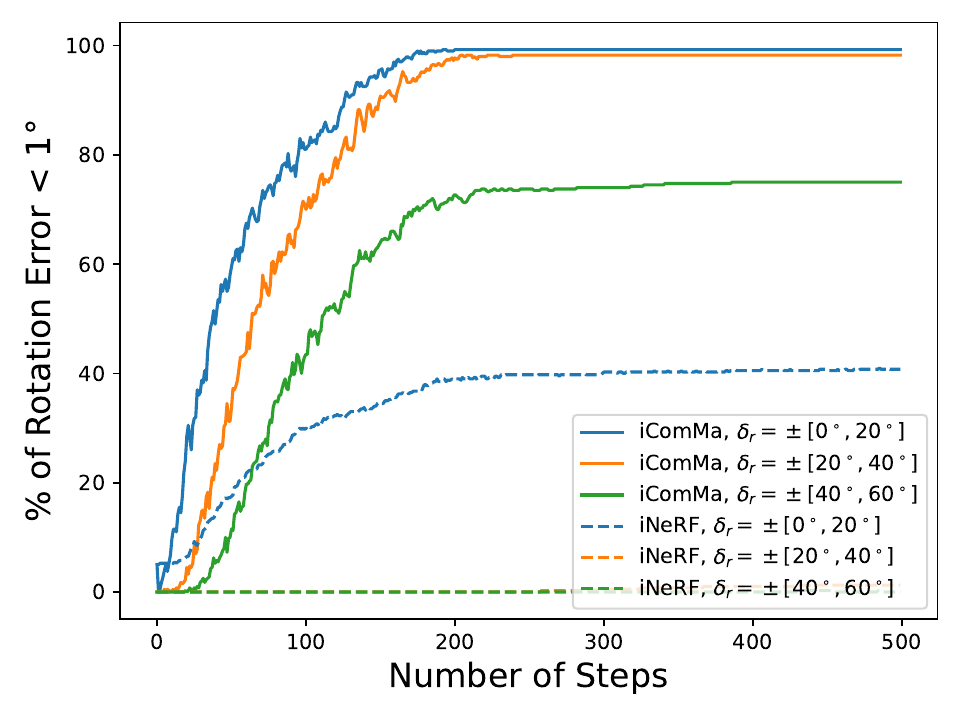} & 
\includegraphics[width=0.31\linewidth]{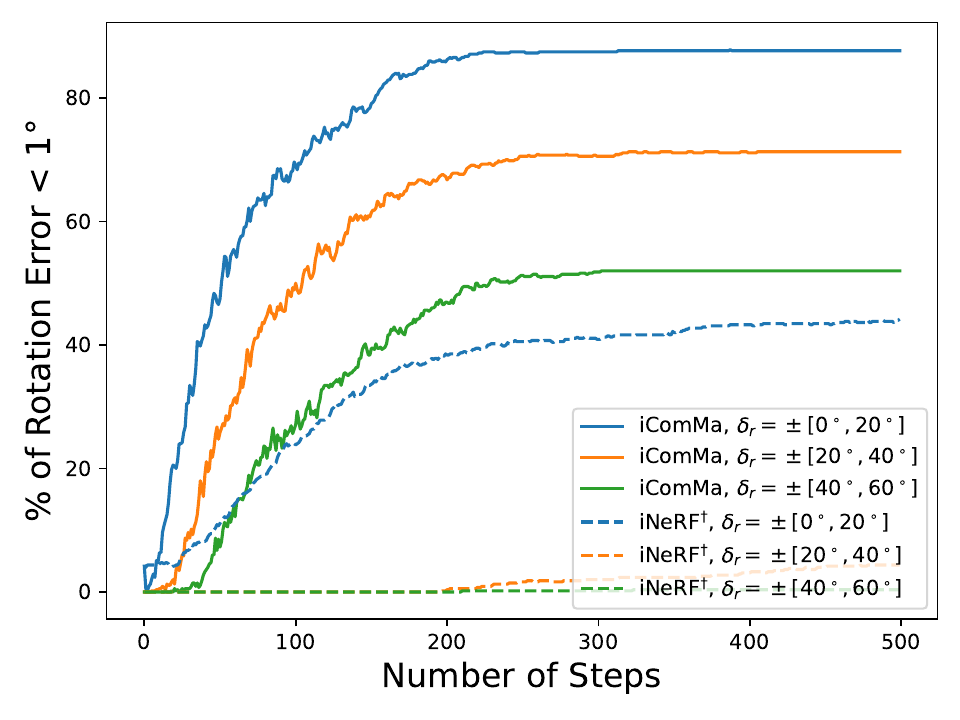} \\
(a) synthetic & (b) LLFF & (c) $360^\circ$ scene \\[2pt]

\includegraphics[width=0.31\linewidth]{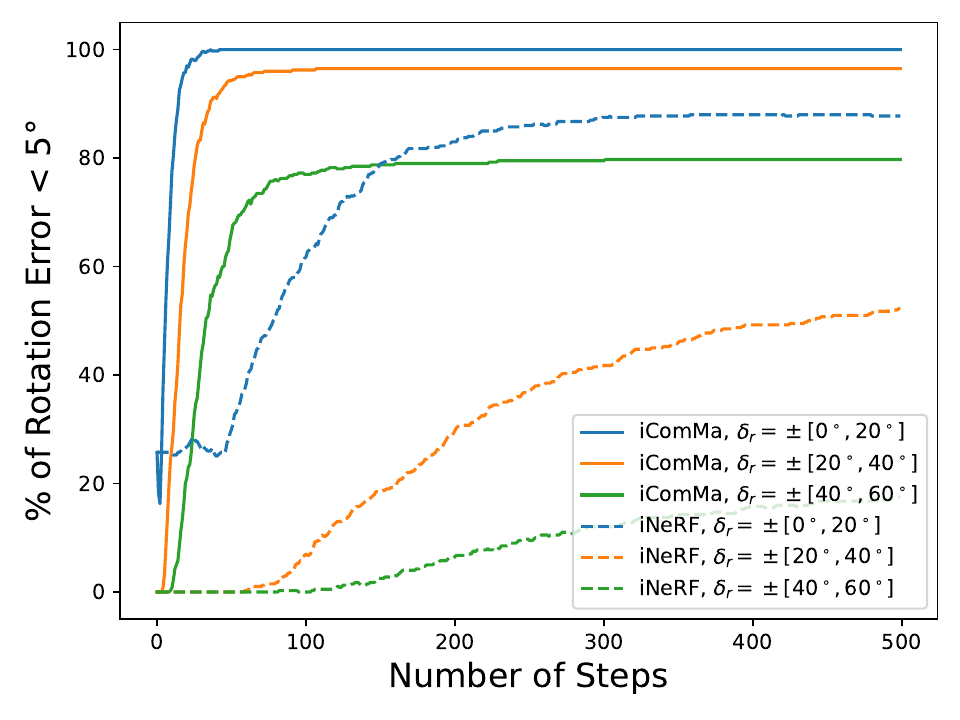} & 
\includegraphics[width=0.31\linewidth]{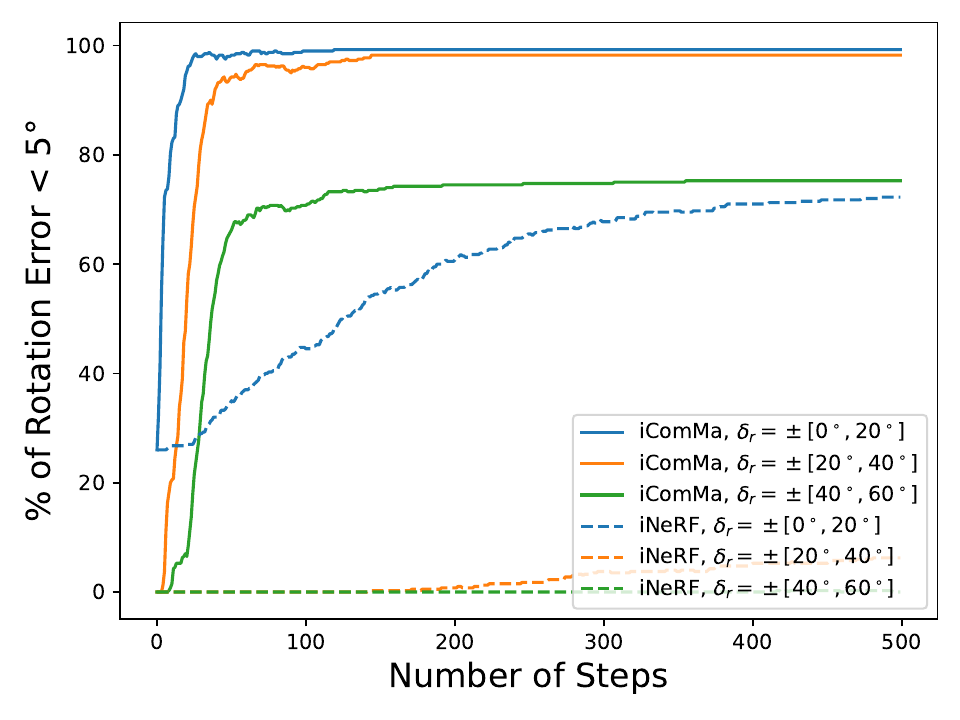} & 
\includegraphics[width=0.31\linewidth]{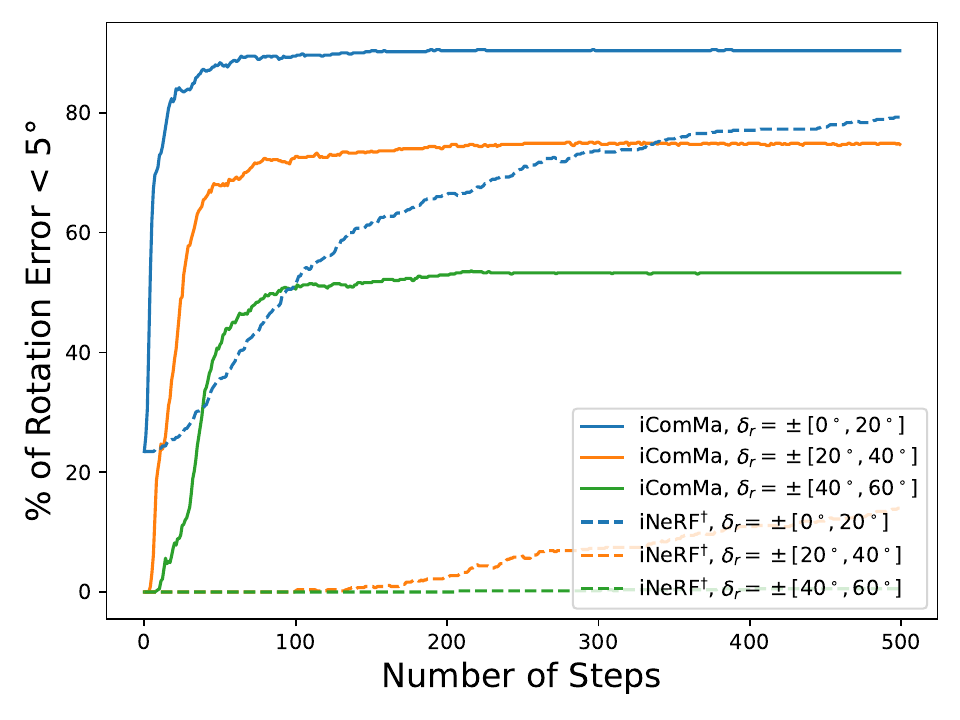} \\
(d) synthetic & (e) LLFF & (f) $360^\circ$ scene \\[2pt]

\includegraphics[width=0.31\linewidth]{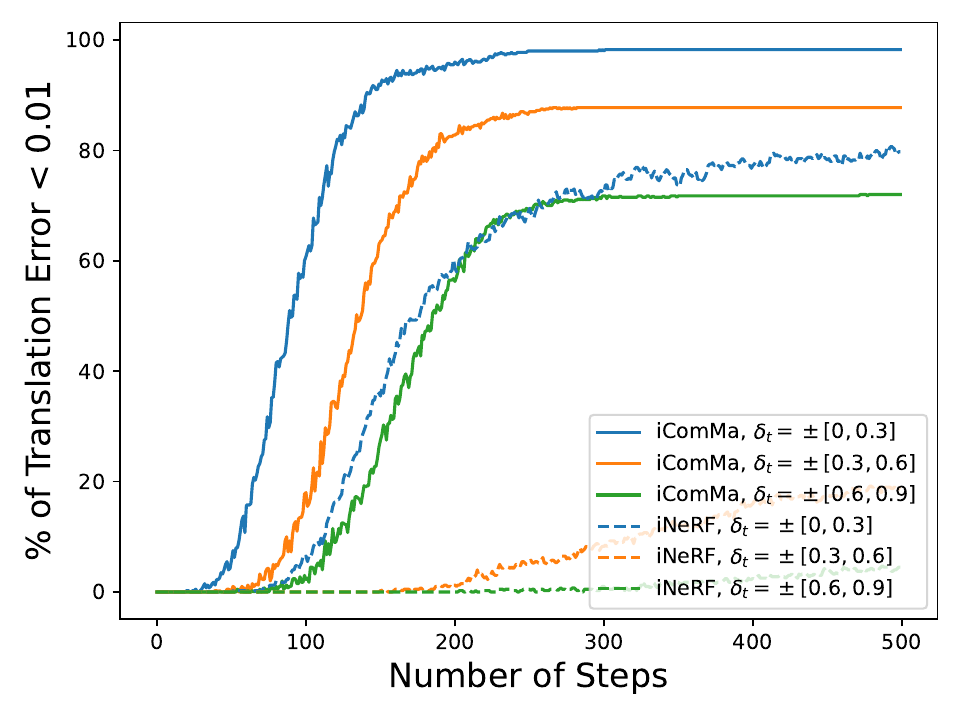} & 
\includegraphics[width=0.31\linewidth]{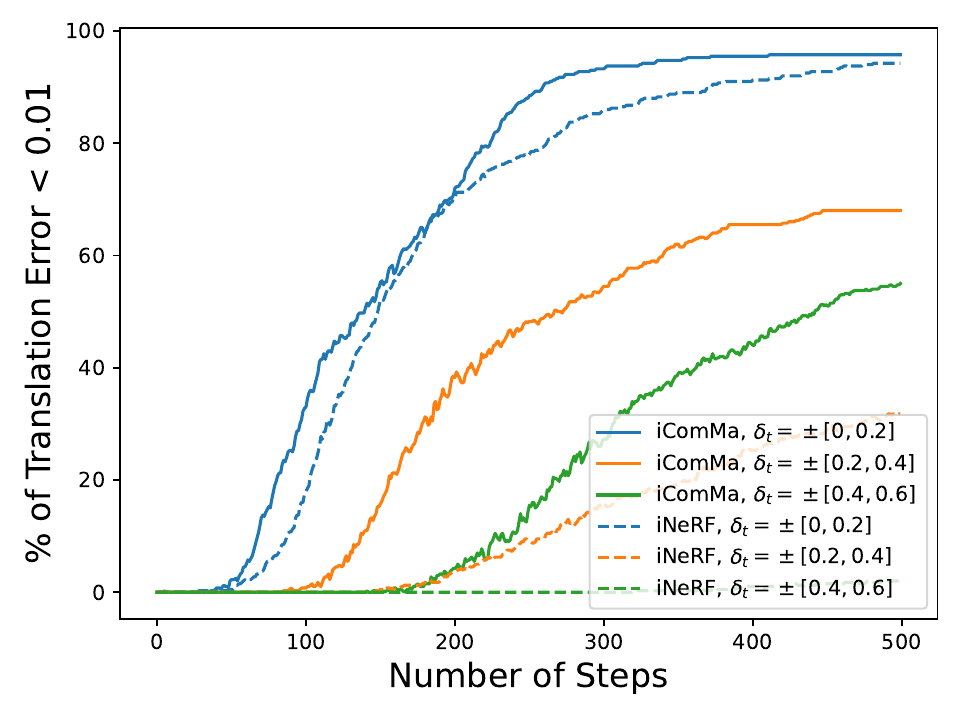} & 
\includegraphics[width=0.31\linewidth]{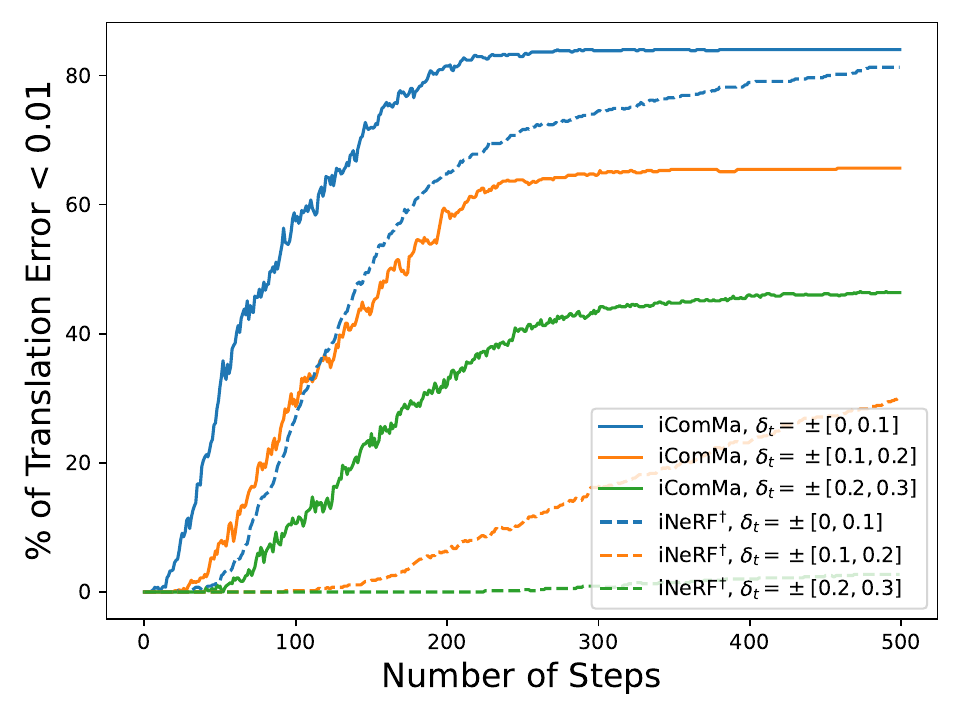} \\
(g) synthetic & (h) LLFF & (i) $360^\circ$ scene \\[2pt]

\includegraphics[width=0.31\linewidth]{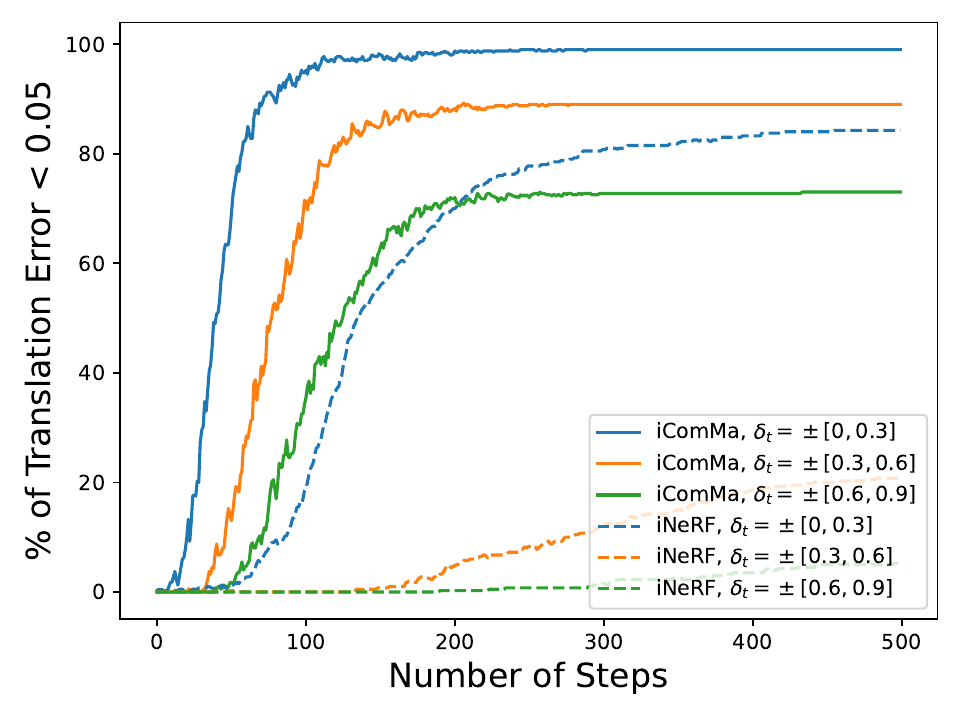} & 
\includegraphics[width=0.31\linewidth]{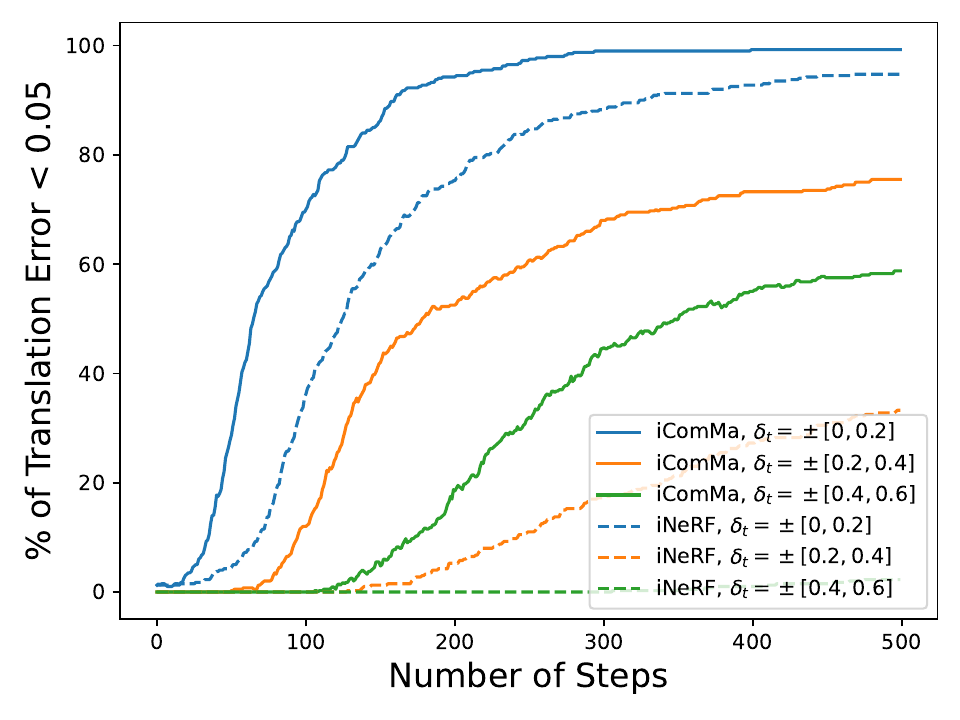} & 
\includegraphics[width=0.31\linewidth]{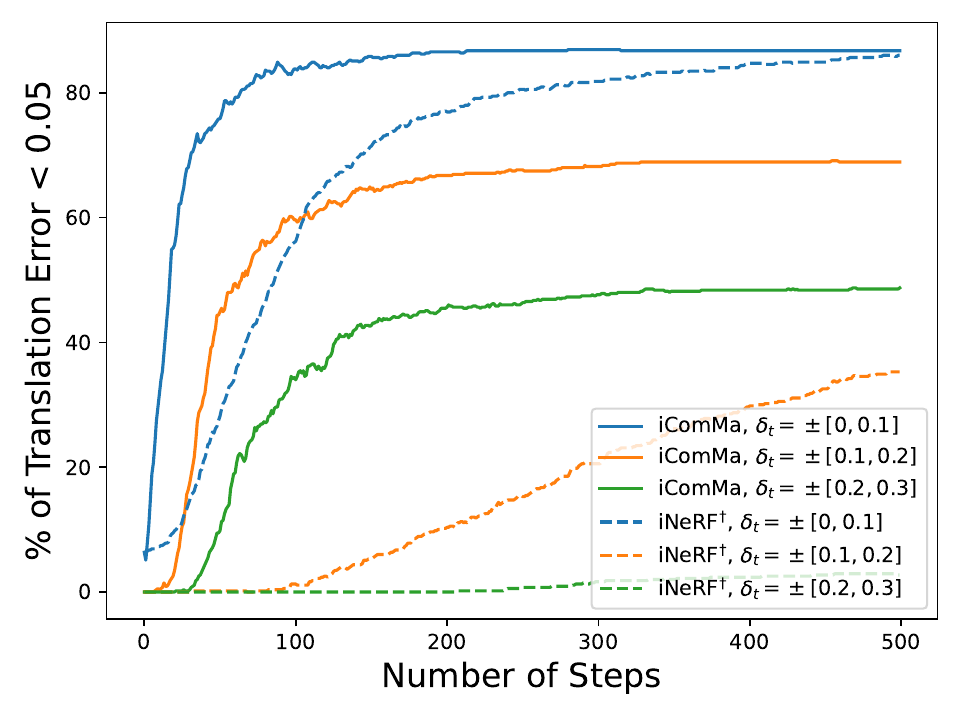} \\
(j) synthetic & (k) LLFF & (l) $360^\circ$ scene \\[2pt]

\end{tabular}
\caption{Quantitative Comparison with iNeRF. Different columns represent different types of datasets: the first column represents 8 synthetic datasets, the second column represents 8 forward-facing LLFF datasets, while the third column displays 11 complex $360^\circ$ scene datasets. As for the curves, solid lines represent the results of iComMa, while dashed lines represent those of iNeRF and $\mathrm{iNeRF}^{\dagger}$. The color of the curves indicates different degrees of initialization conditions. Due to space limitation, only the main results are provided. For more details, please refer to the supplementary material.}
\label{fig_com_inerf}
\end{figure*}

\begin{figure}[bth!]
    \centering
    {\scriptsize
    \begin{tabular}{c|ccc|c}
        \textbf{Initial Pose} & \multicolumn{3}{c|}{ \textbf{iComMa}} &  \textbf{iNeRF} \\
        \hline
        \includegraphics[width=0.19\textwidth]{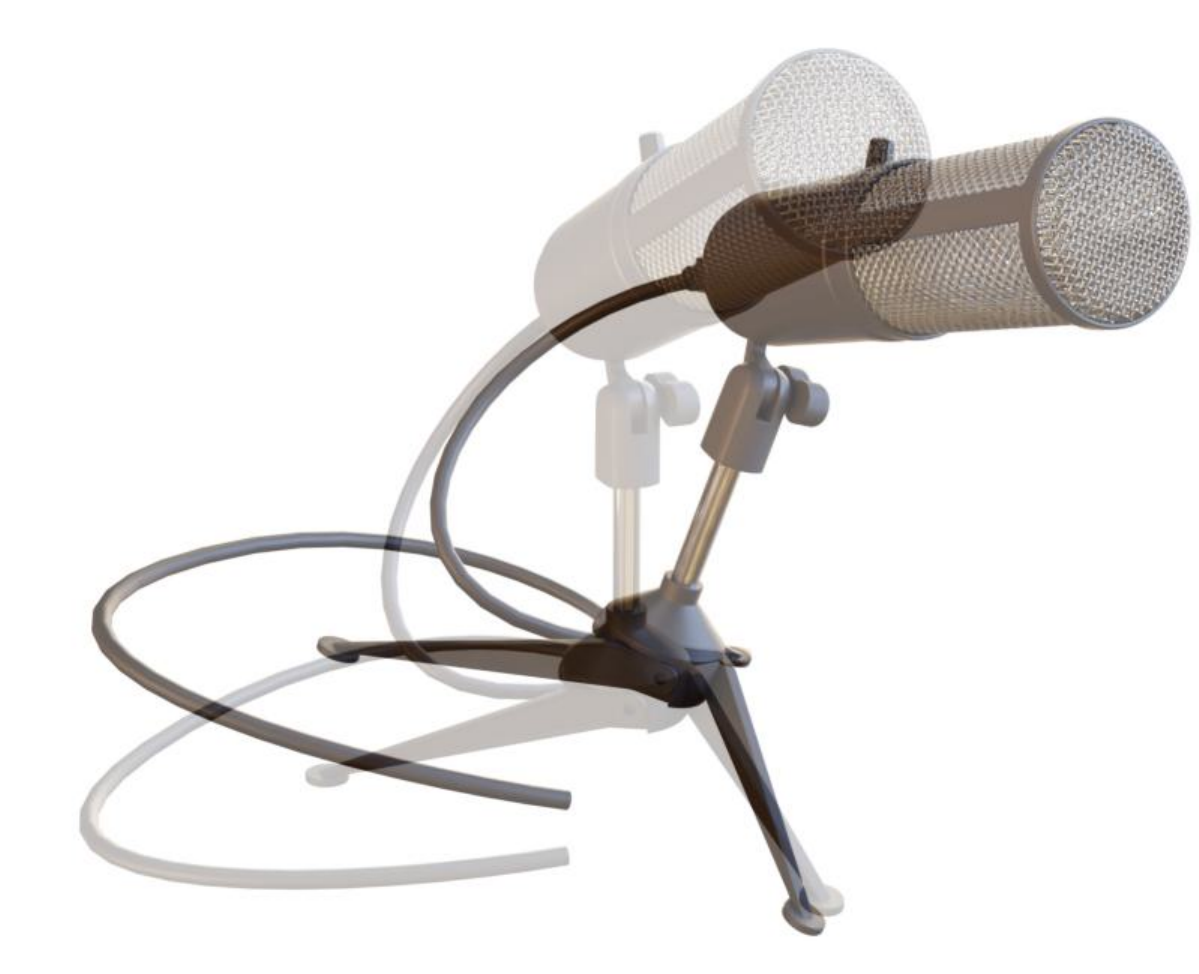} &
        \includegraphics[width=0.19\textwidth]{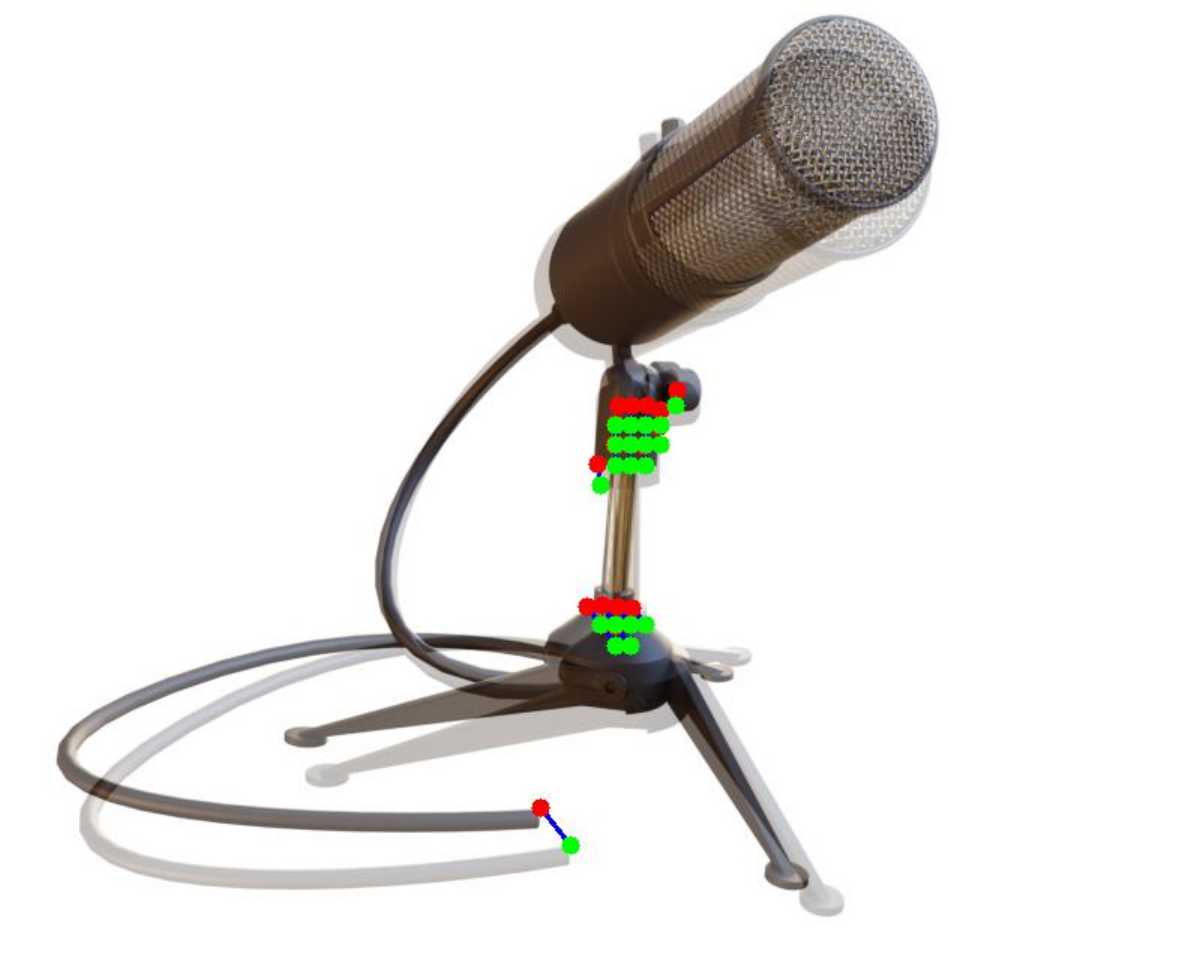} &
        \includegraphics[width=0.19\textwidth]{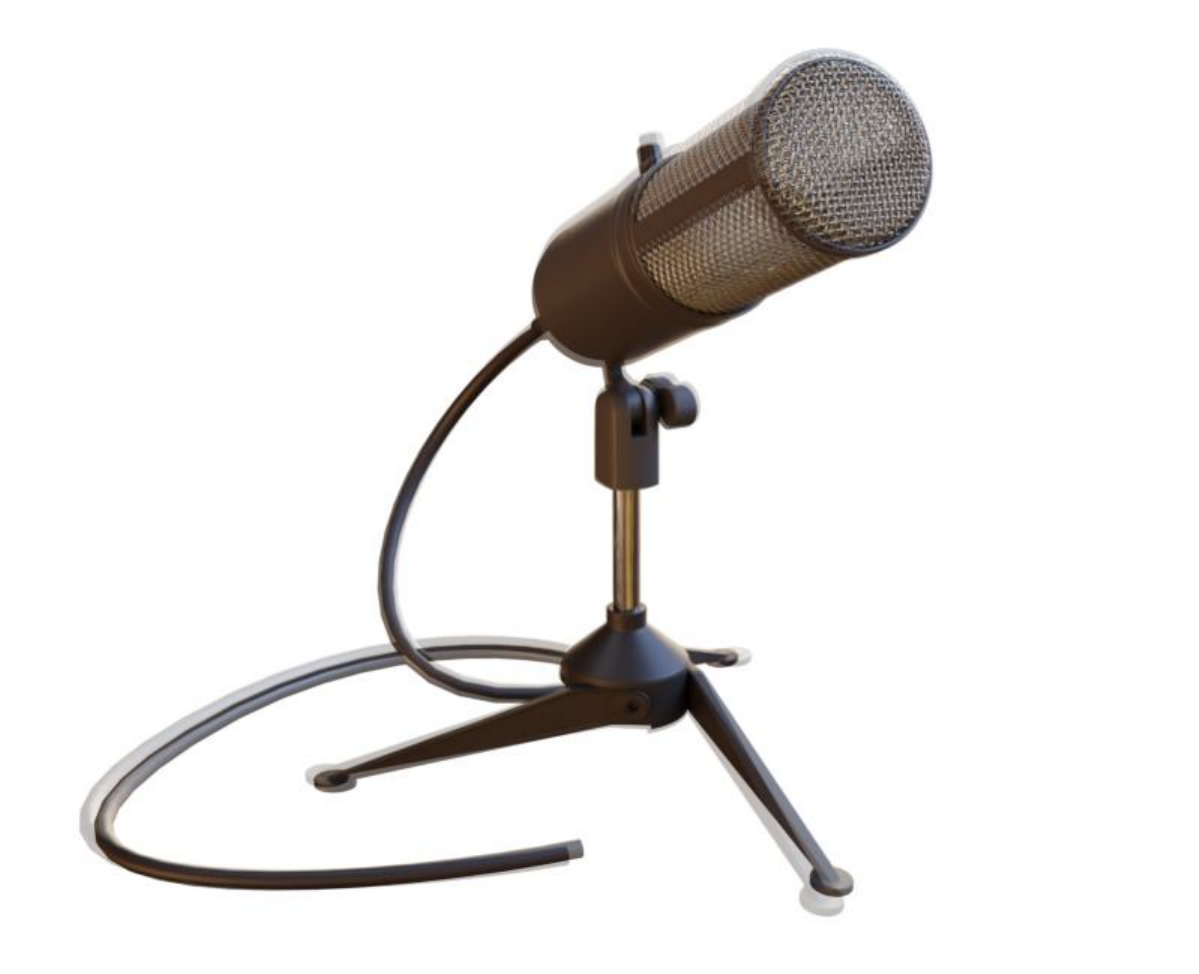} &
        \includegraphics[width=0.19\textwidth]{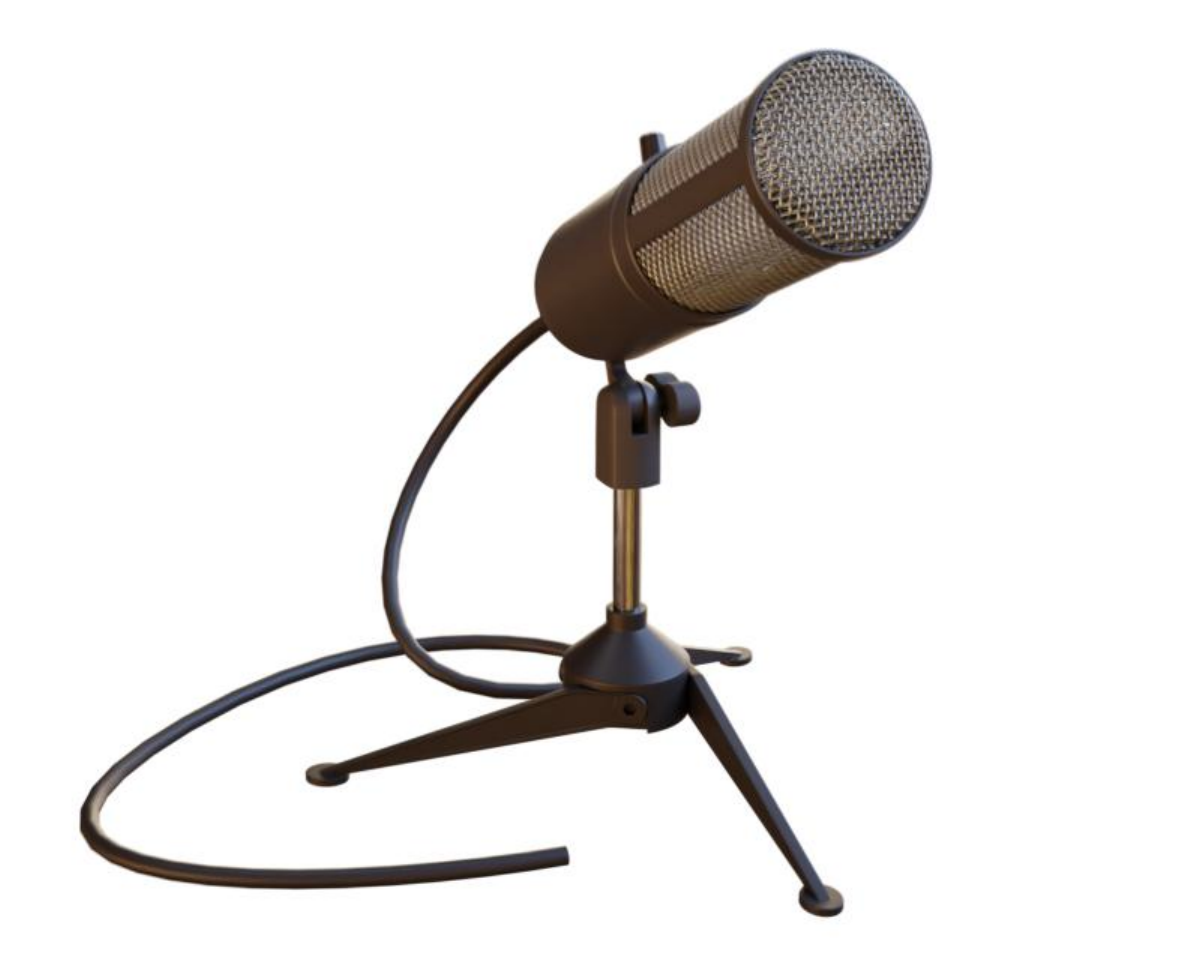} &
        \includegraphics[width=0.19\textwidth]{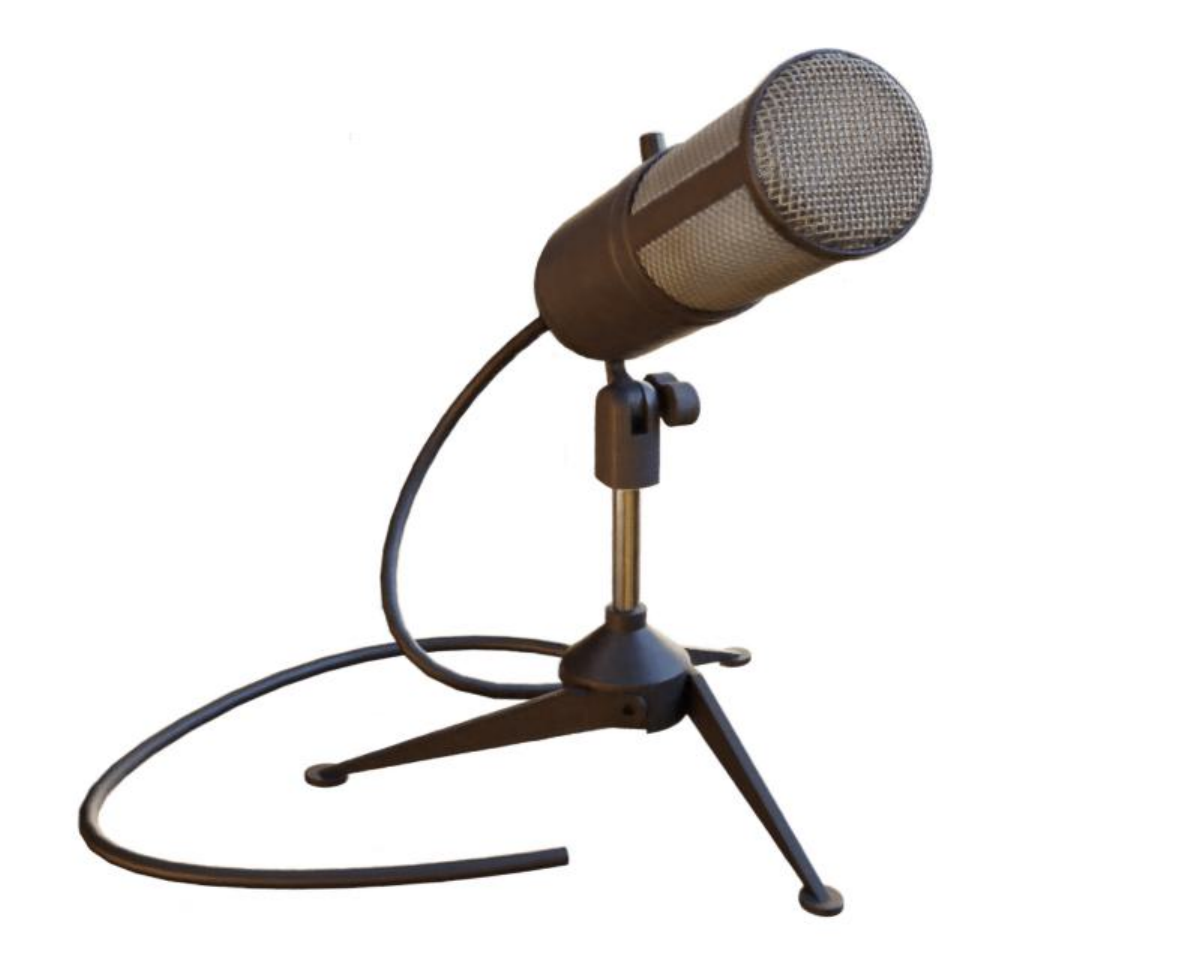} \\
        step=0 & step=5 & step=20 & step=150 & step=200  \\
        \includegraphics[width=0.19\textwidth]{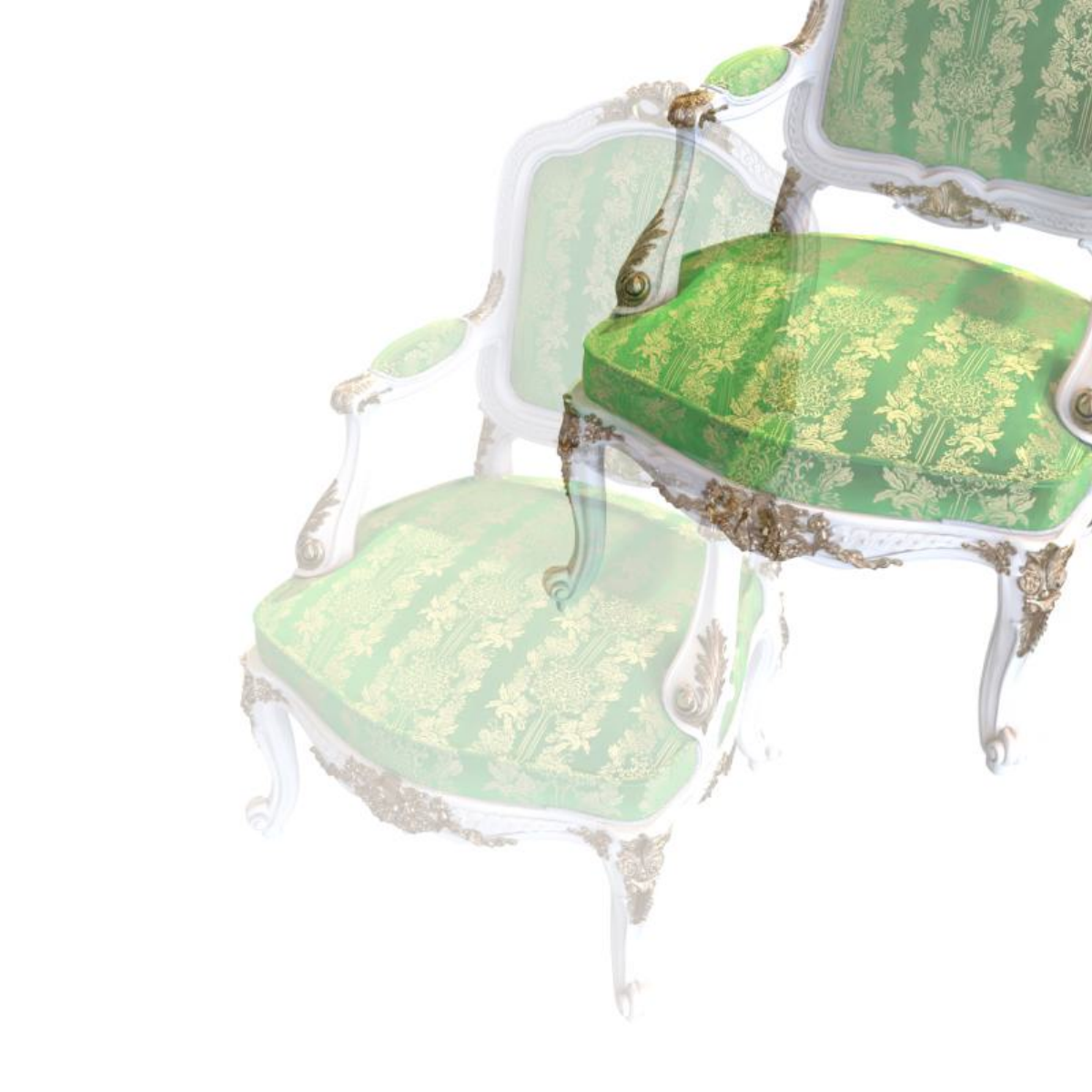} &
        \includegraphics[width=0.19\textwidth]{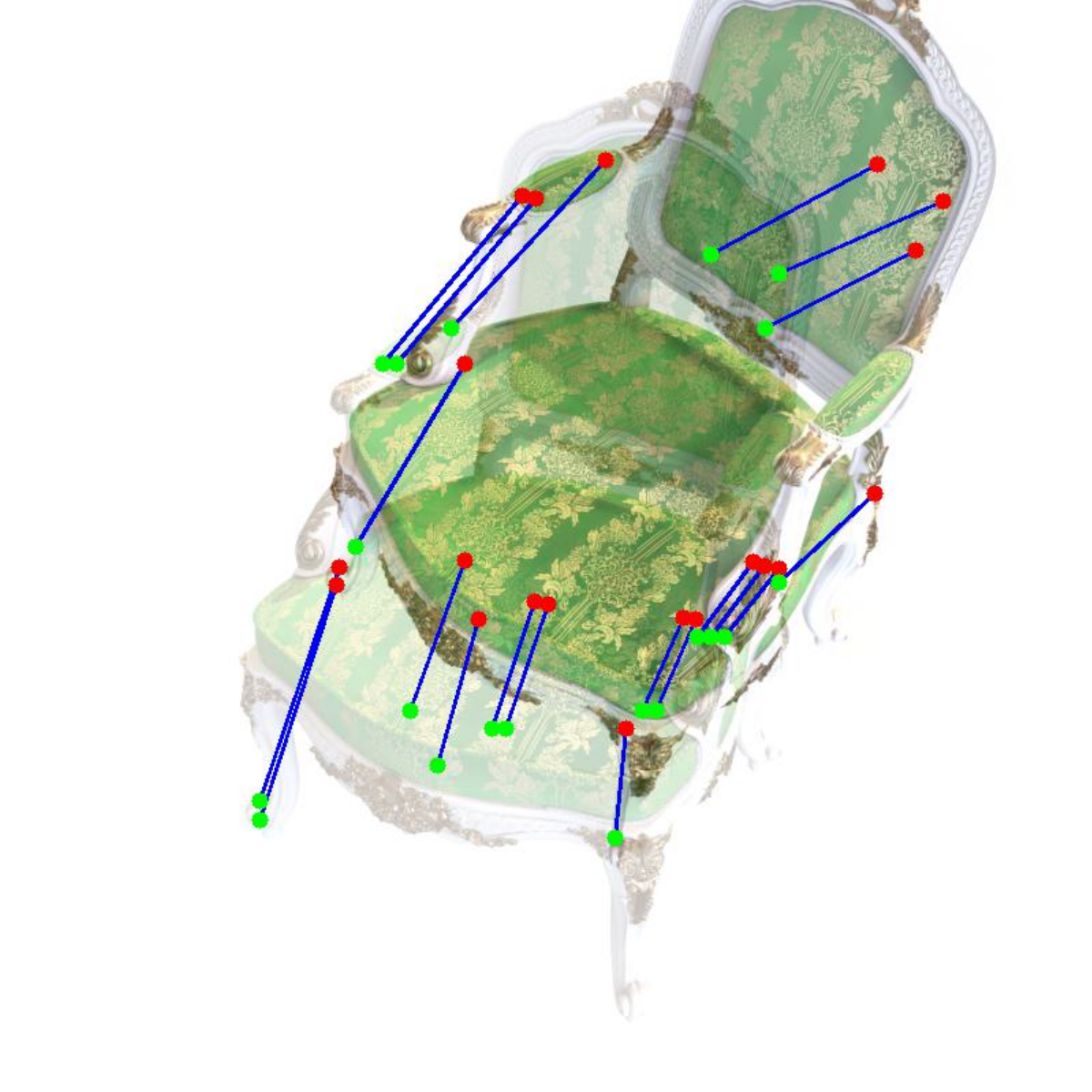} &
        \includegraphics[width=0.19\textwidth]{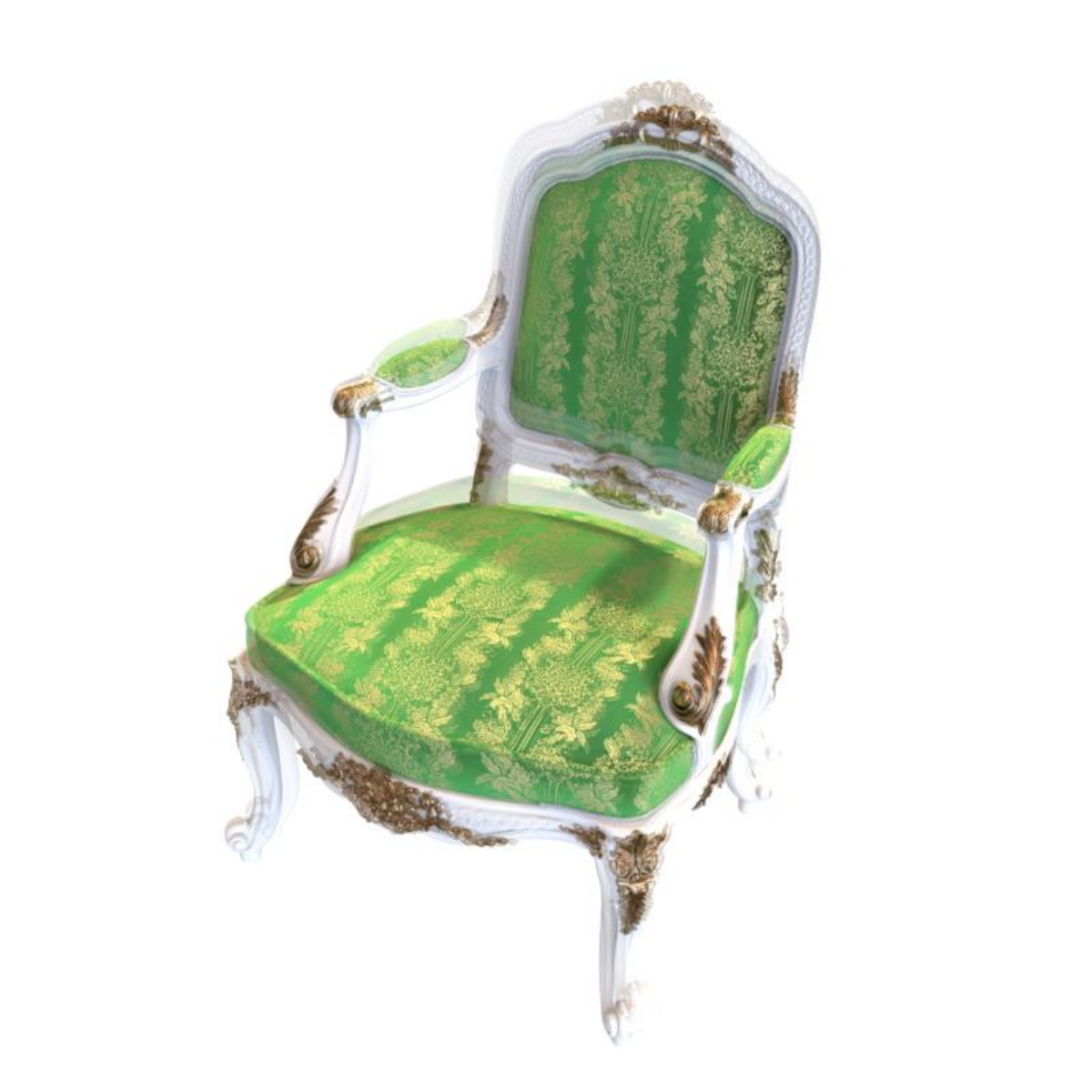} &
        \includegraphics[width=0.19\textwidth]{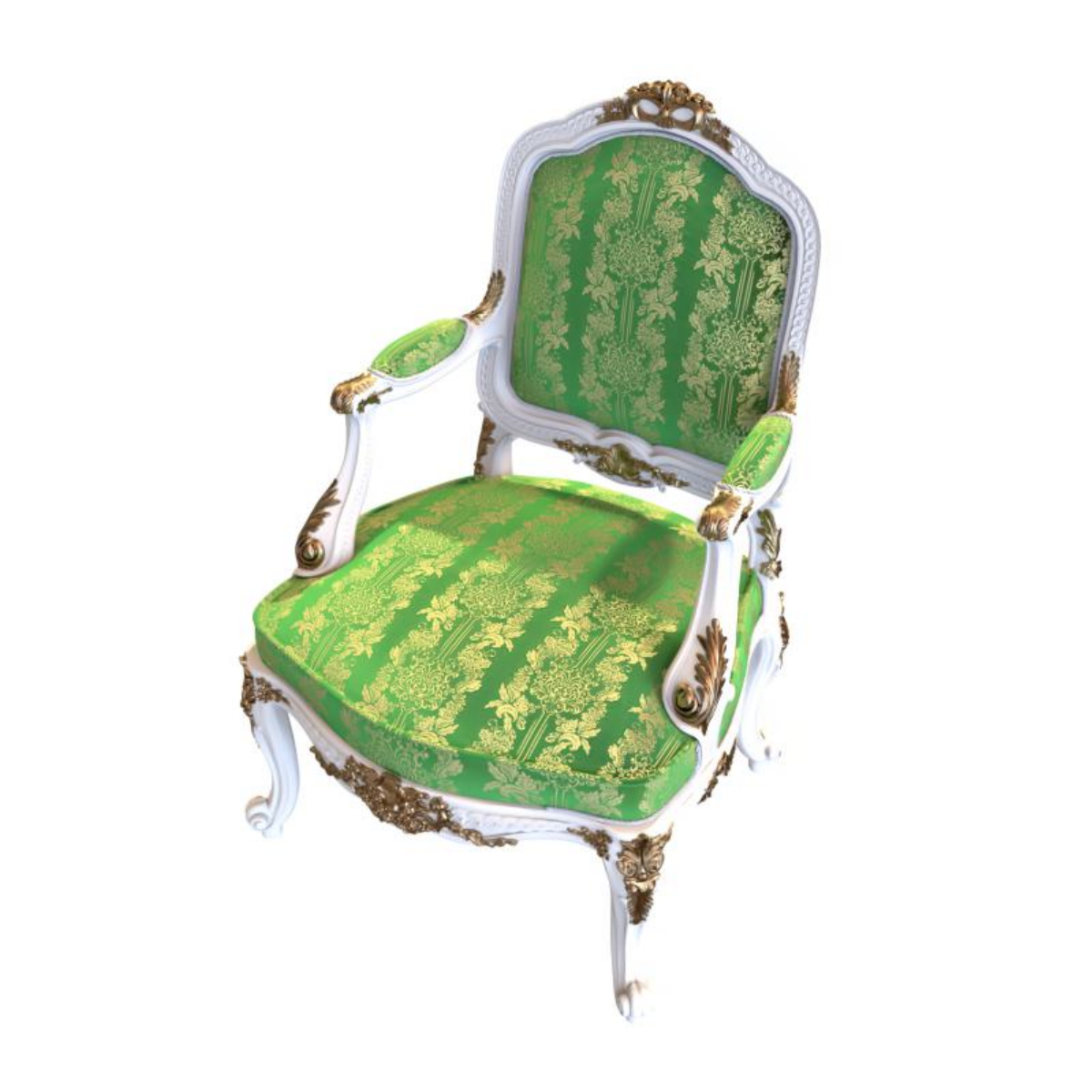} &
        \includegraphics[width=0.19\textwidth]{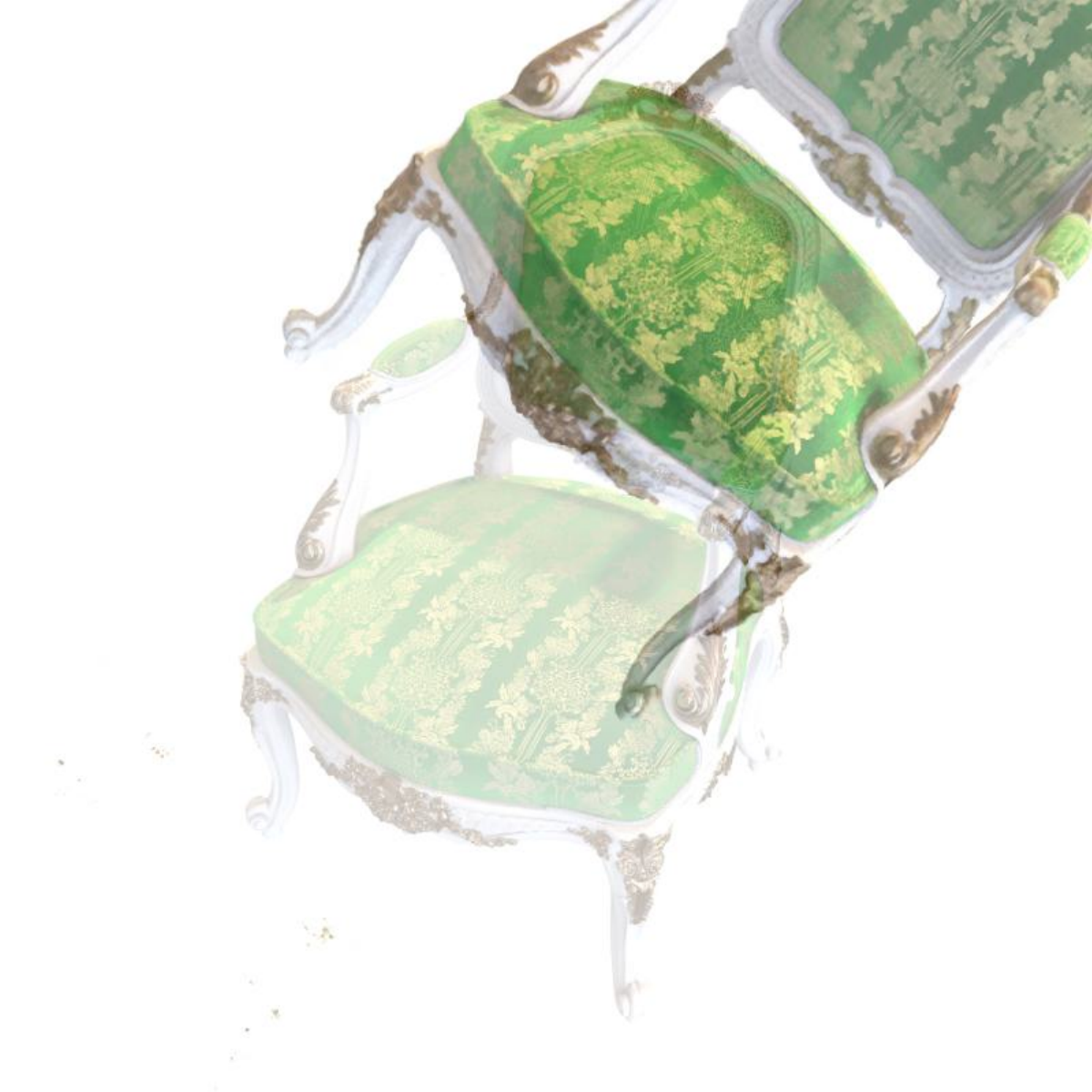} \\
         step=0 & step=10 & step=50 & step=200 & step=1000  \\
        \includegraphics[width=0.19\textwidth]{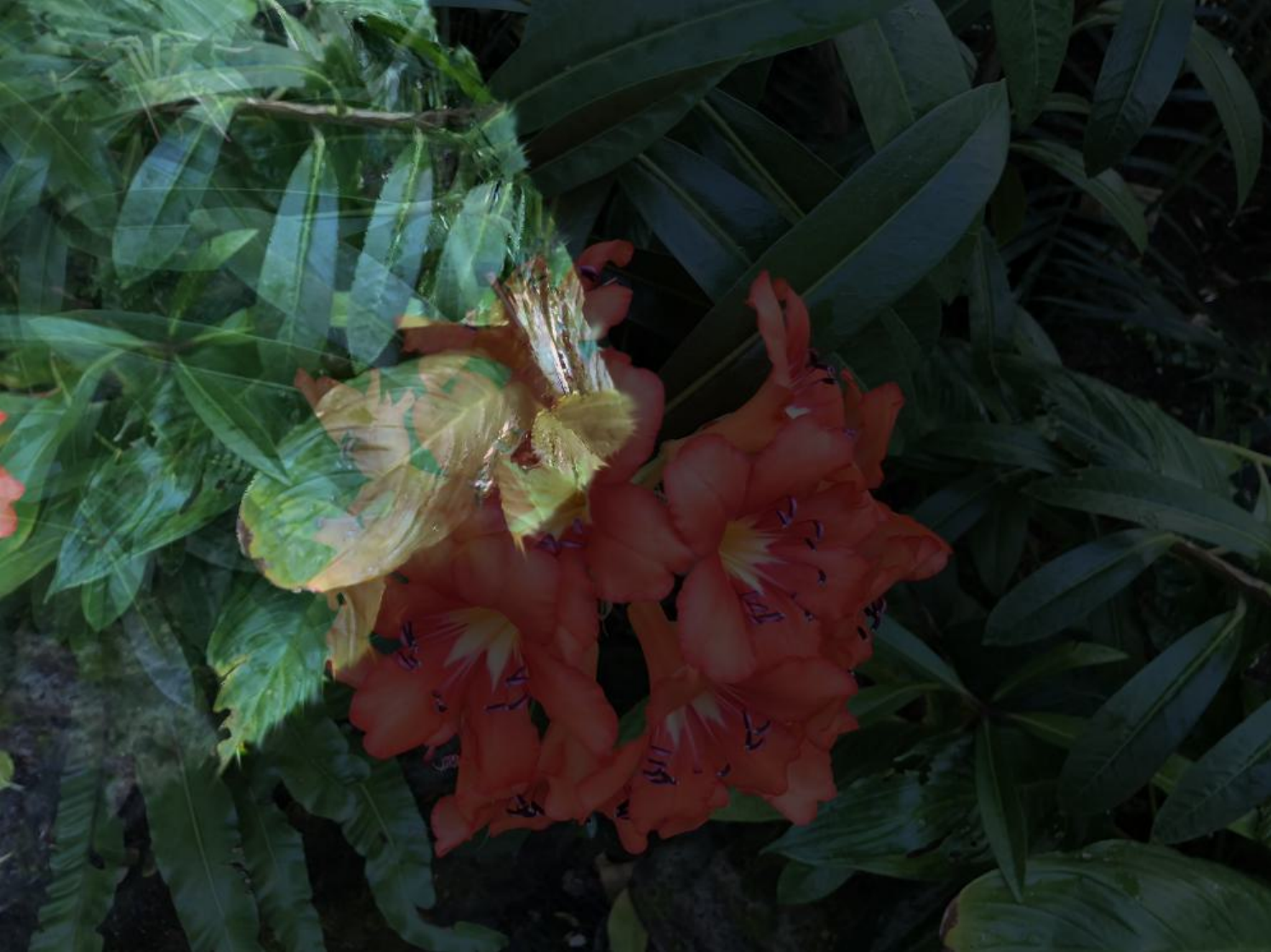} &
        \includegraphics[width=0.19\textwidth]{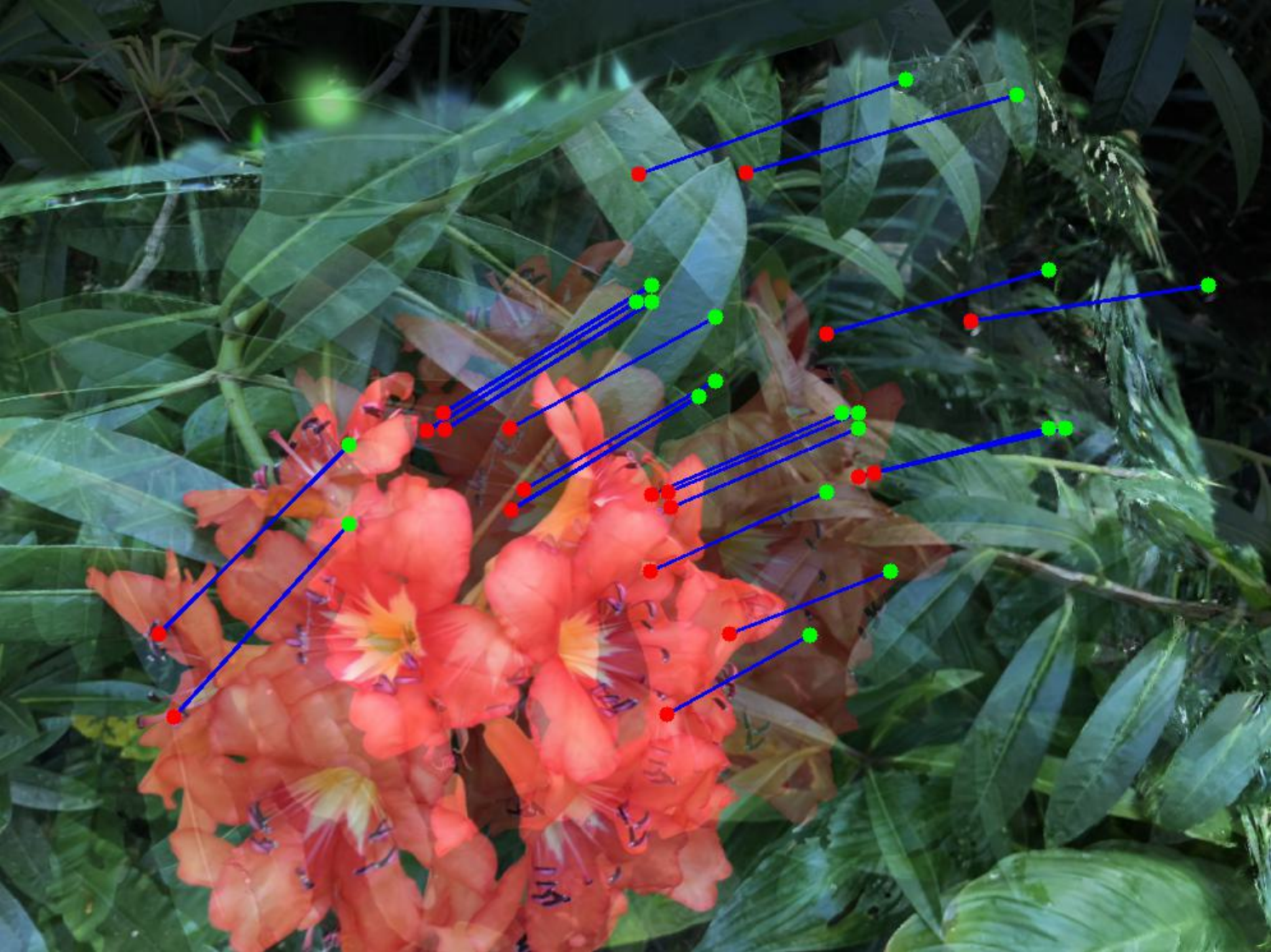} &
        \includegraphics[width=0.19\textwidth]{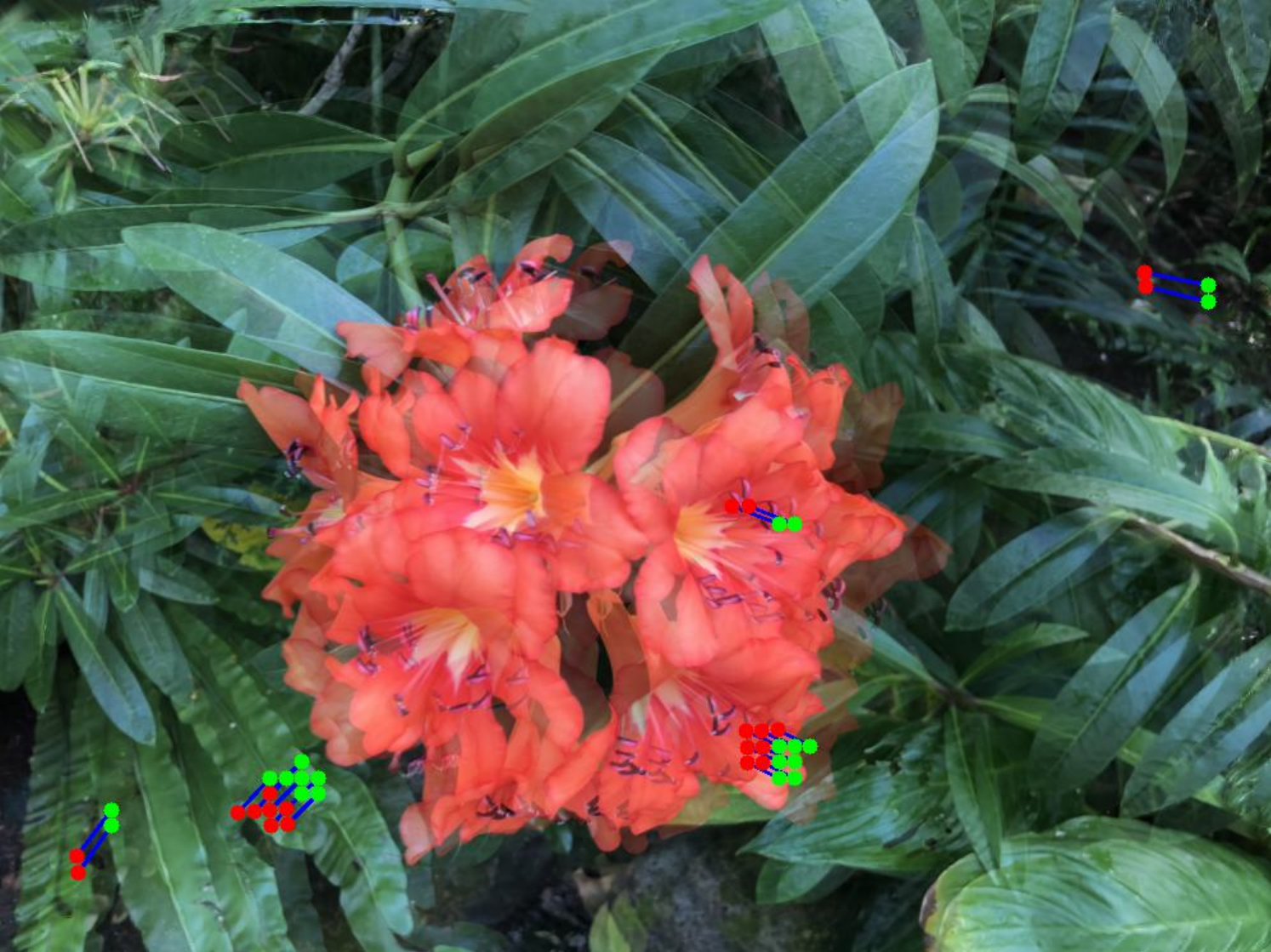} &
        \includegraphics[width=0.19\textwidth]{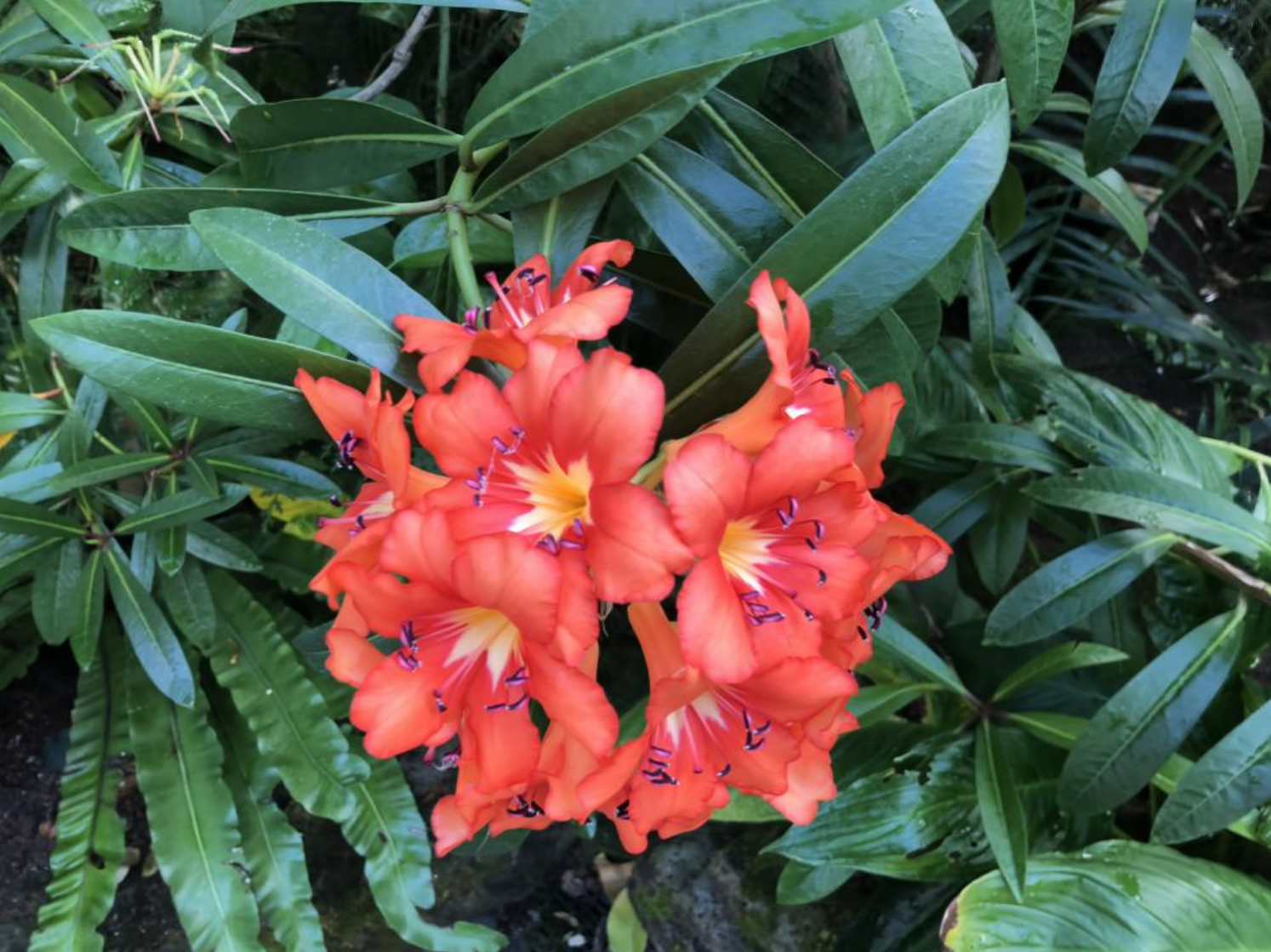} &
        \includegraphics[width=0.19\textwidth]{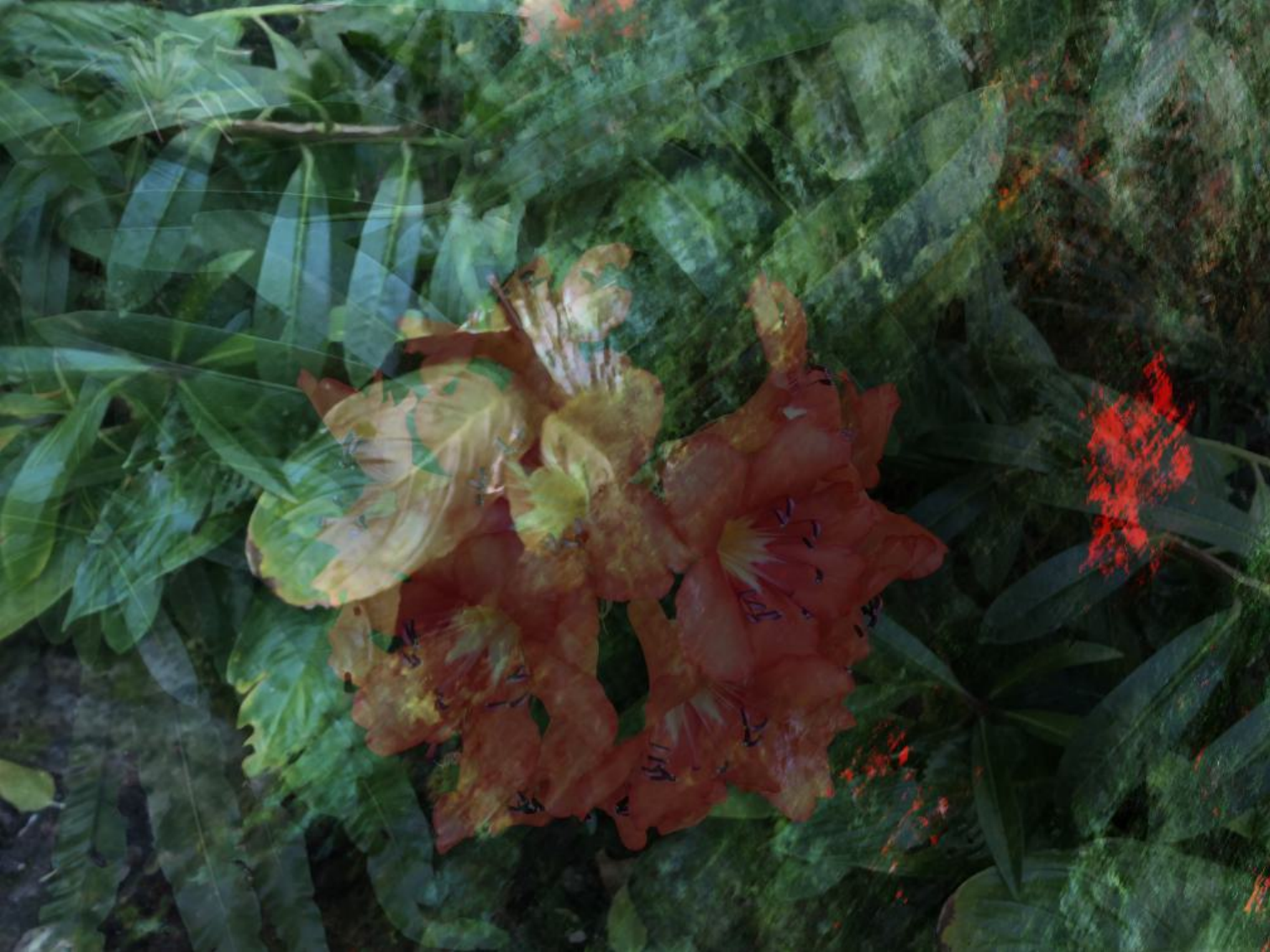} \\
         step=0 & step=10 & step=30 & step=200 & step=1000 \\
        \includegraphics[width=0.19\textwidth]{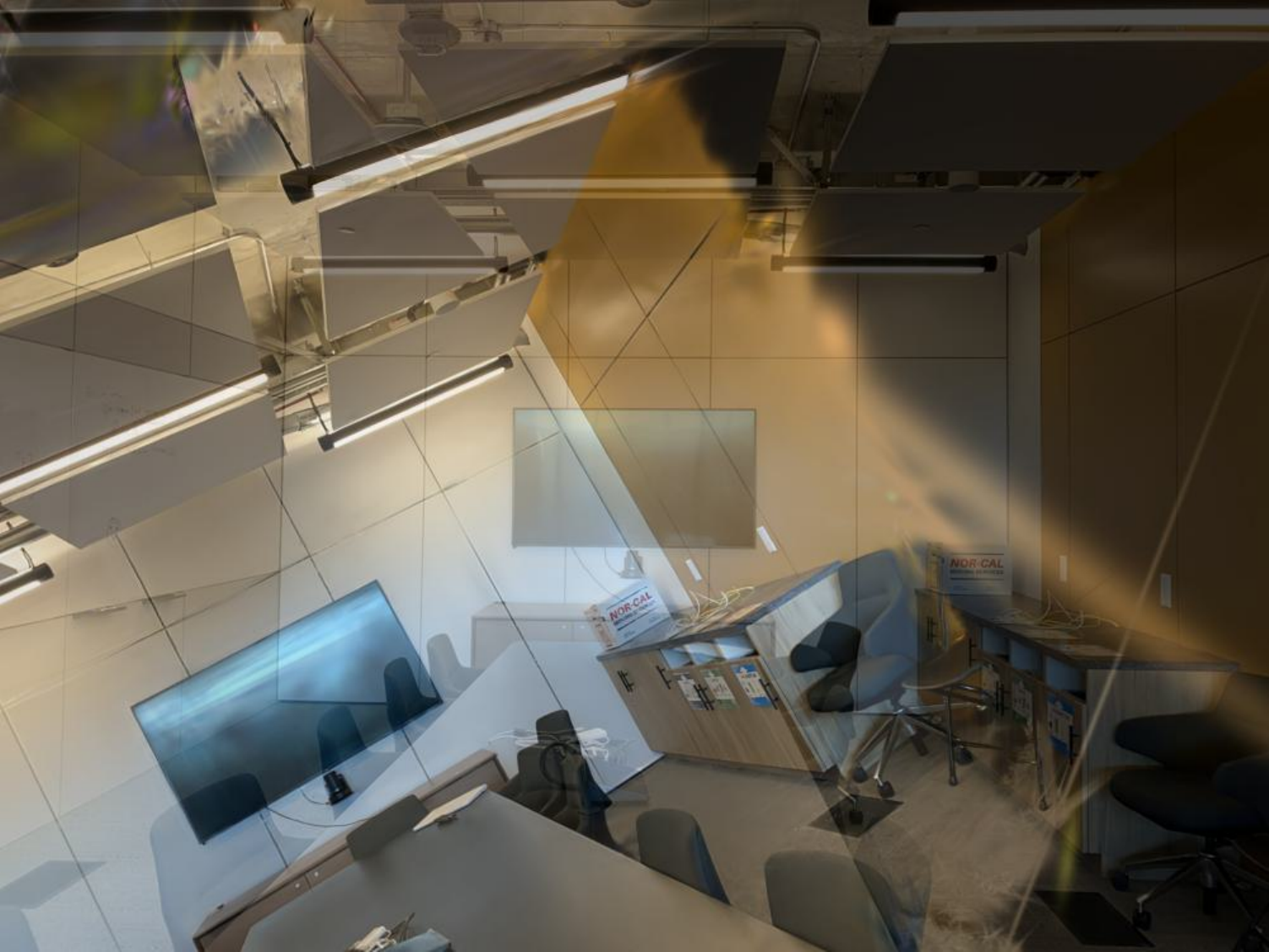} &
        \includegraphics[width=0.19\textwidth]{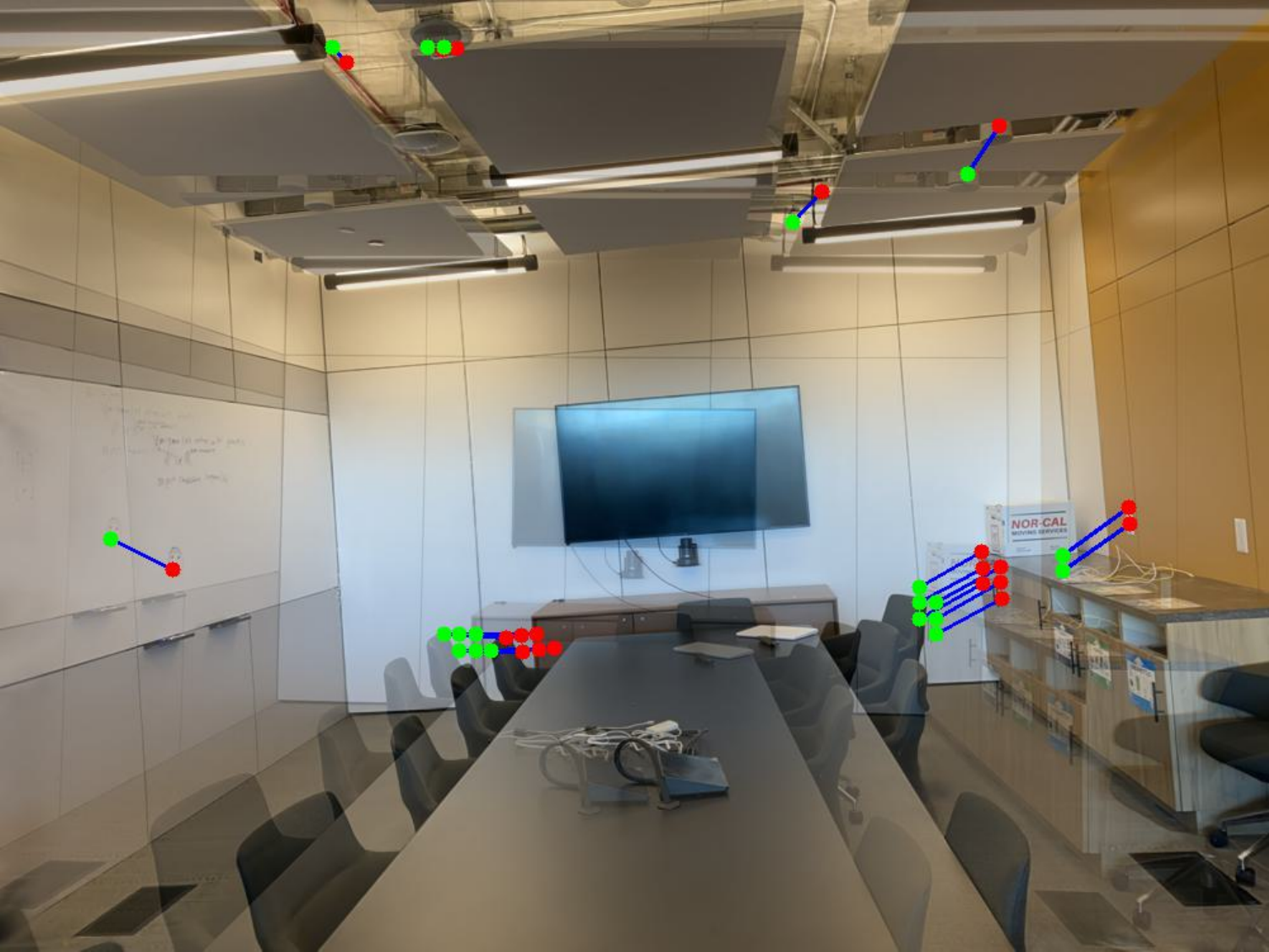} &
        \includegraphics[width=0.19\textwidth]{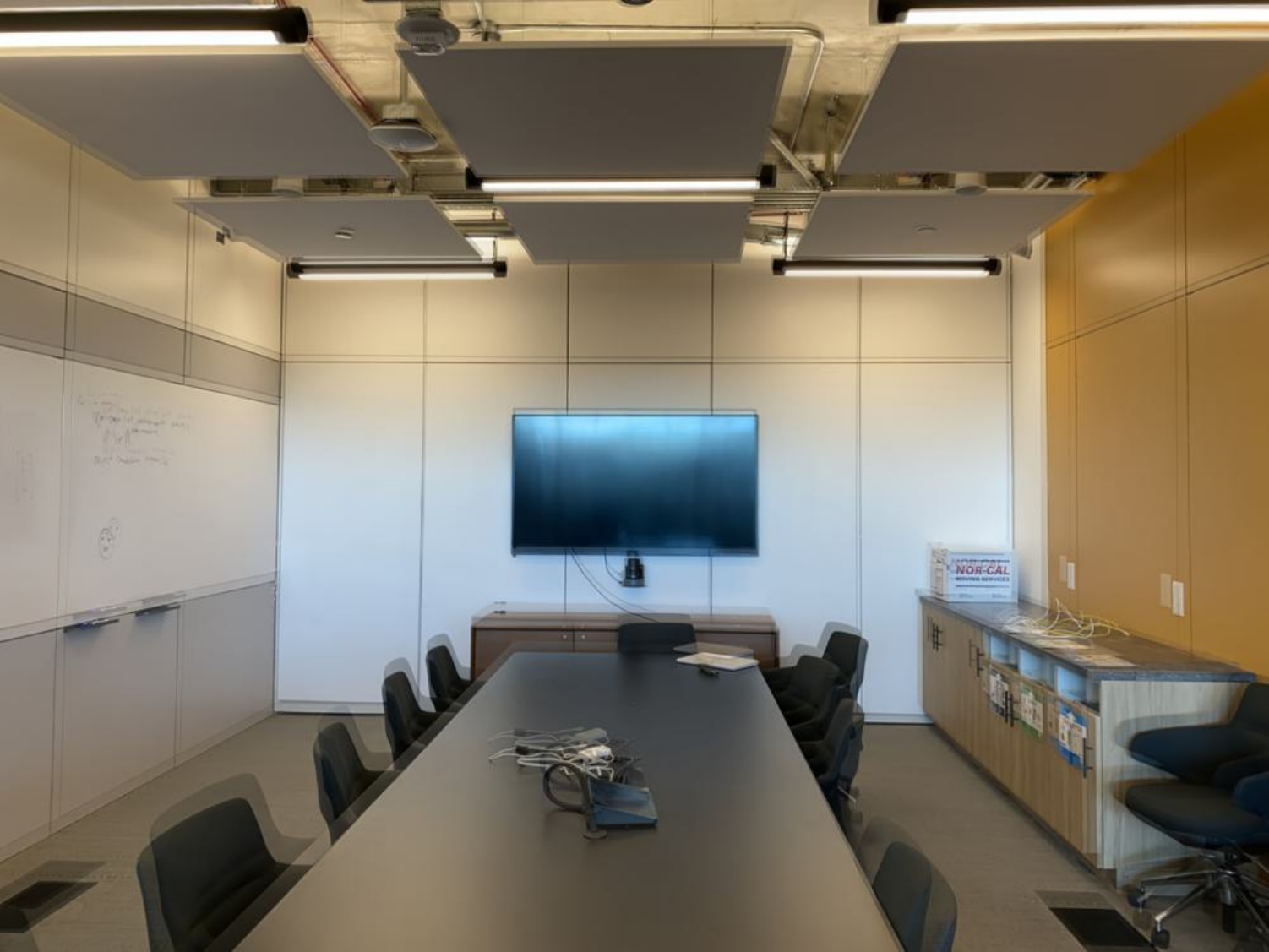} &
        \includegraphics[width=0.19\textwidth]{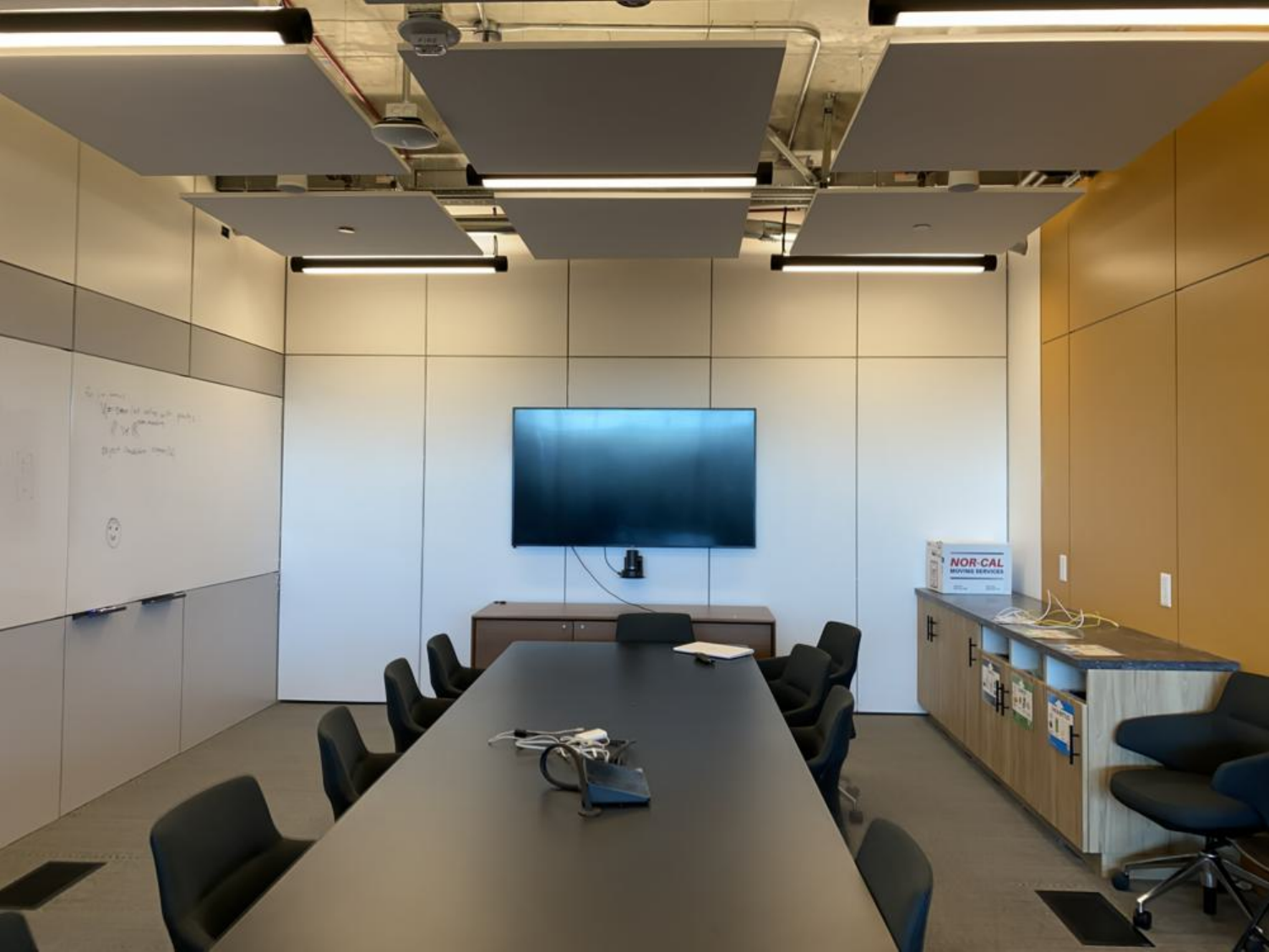} &
        \includegraphics[width=0.19\textwidth]{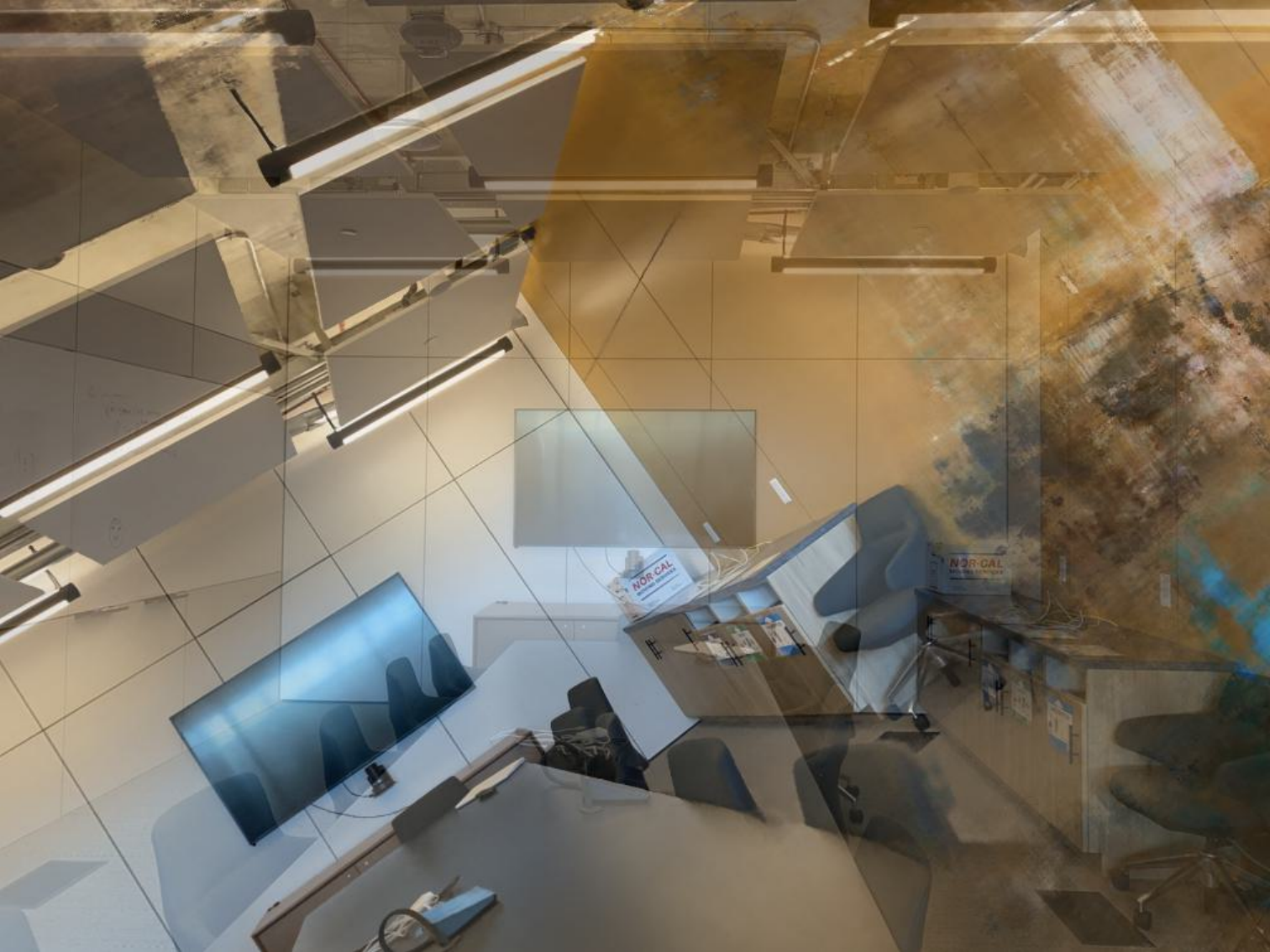} \\
        step=0 & step=10 & step=50 & step=200 & step=1000 \\  
        \includegraphics[width=0.19\textwidth]{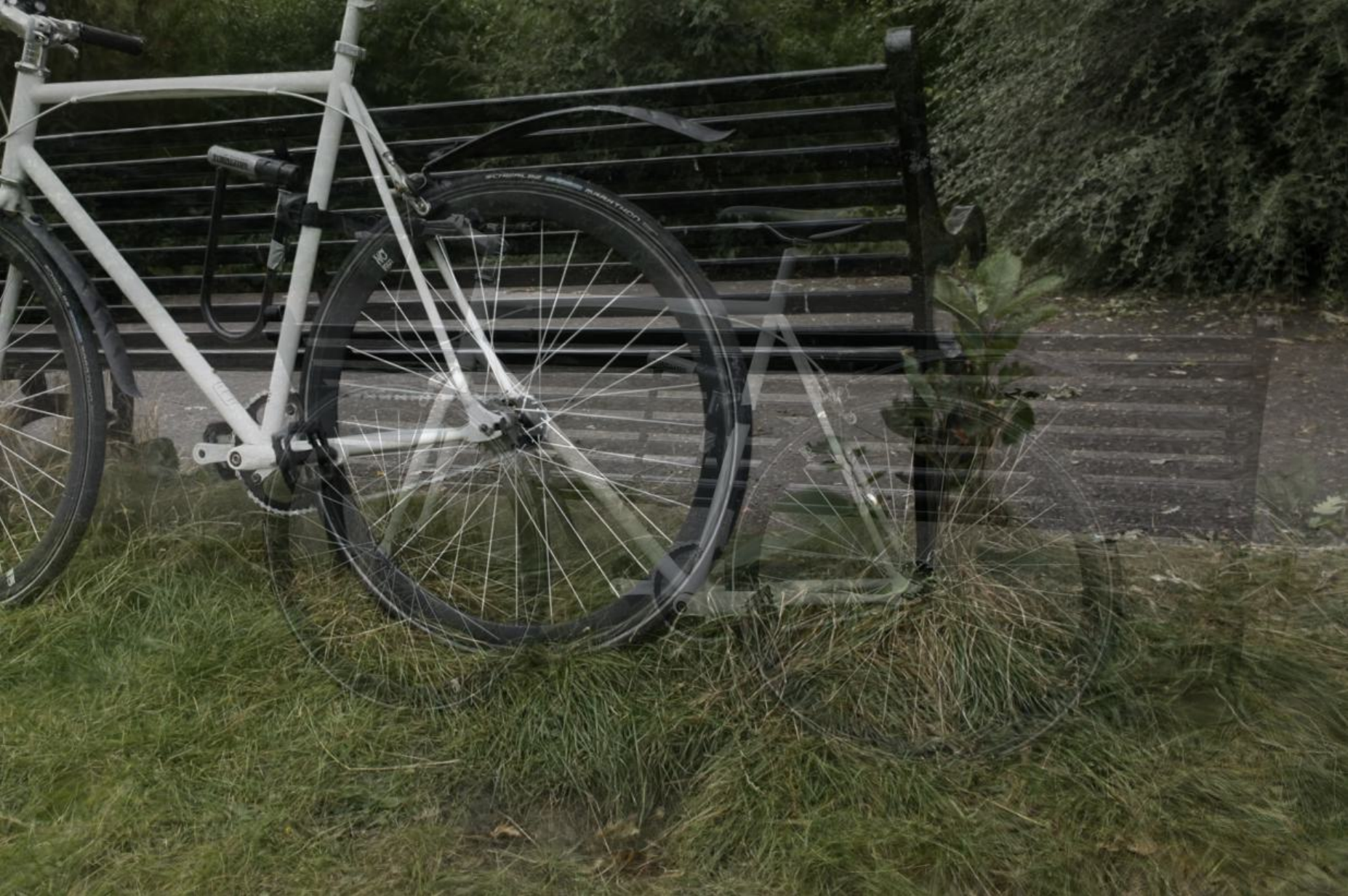} &
        \includegraphics[width=0.19\textwidth]{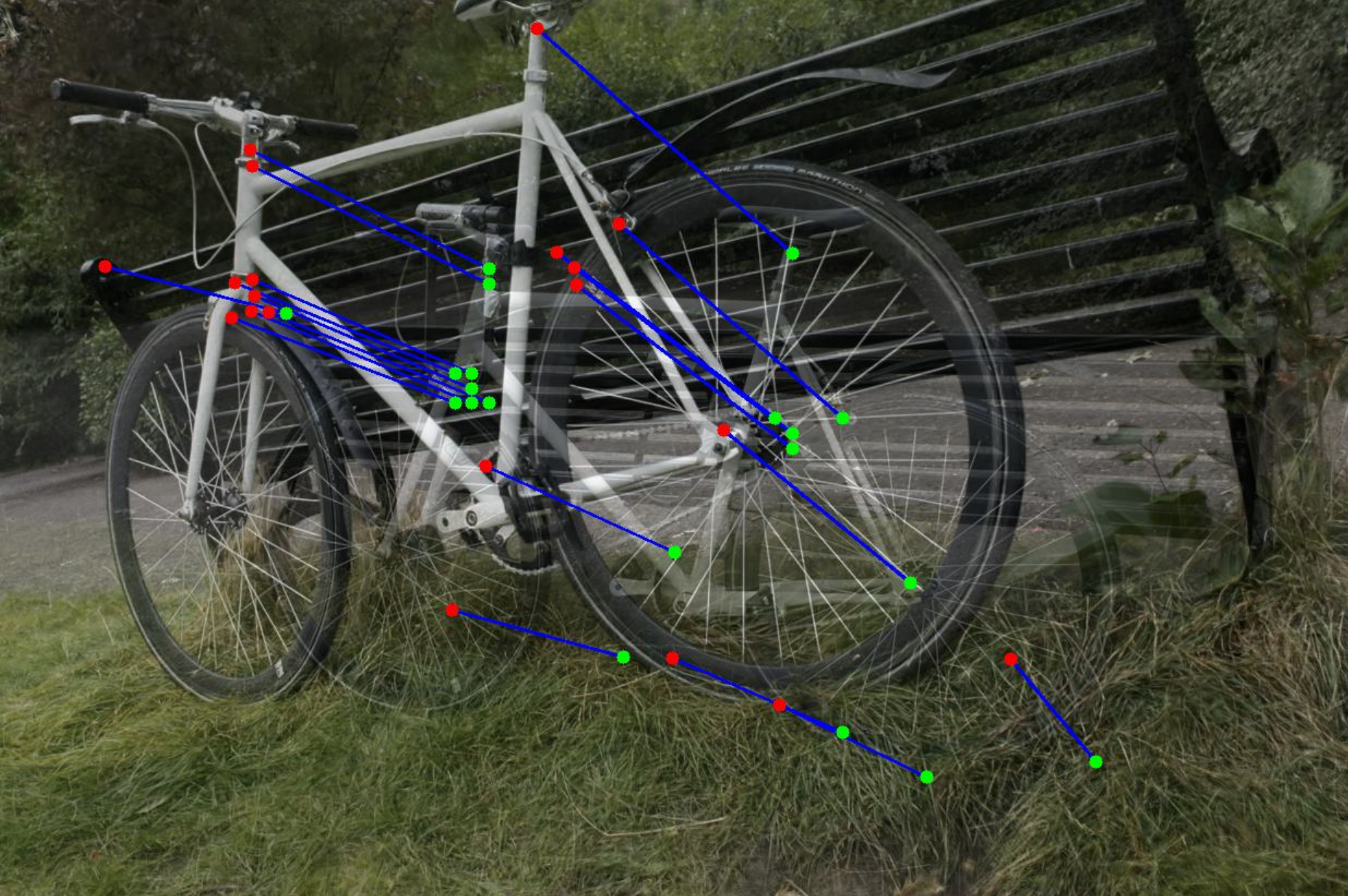} &
        \includegraphics[width=0.19\textwidth]{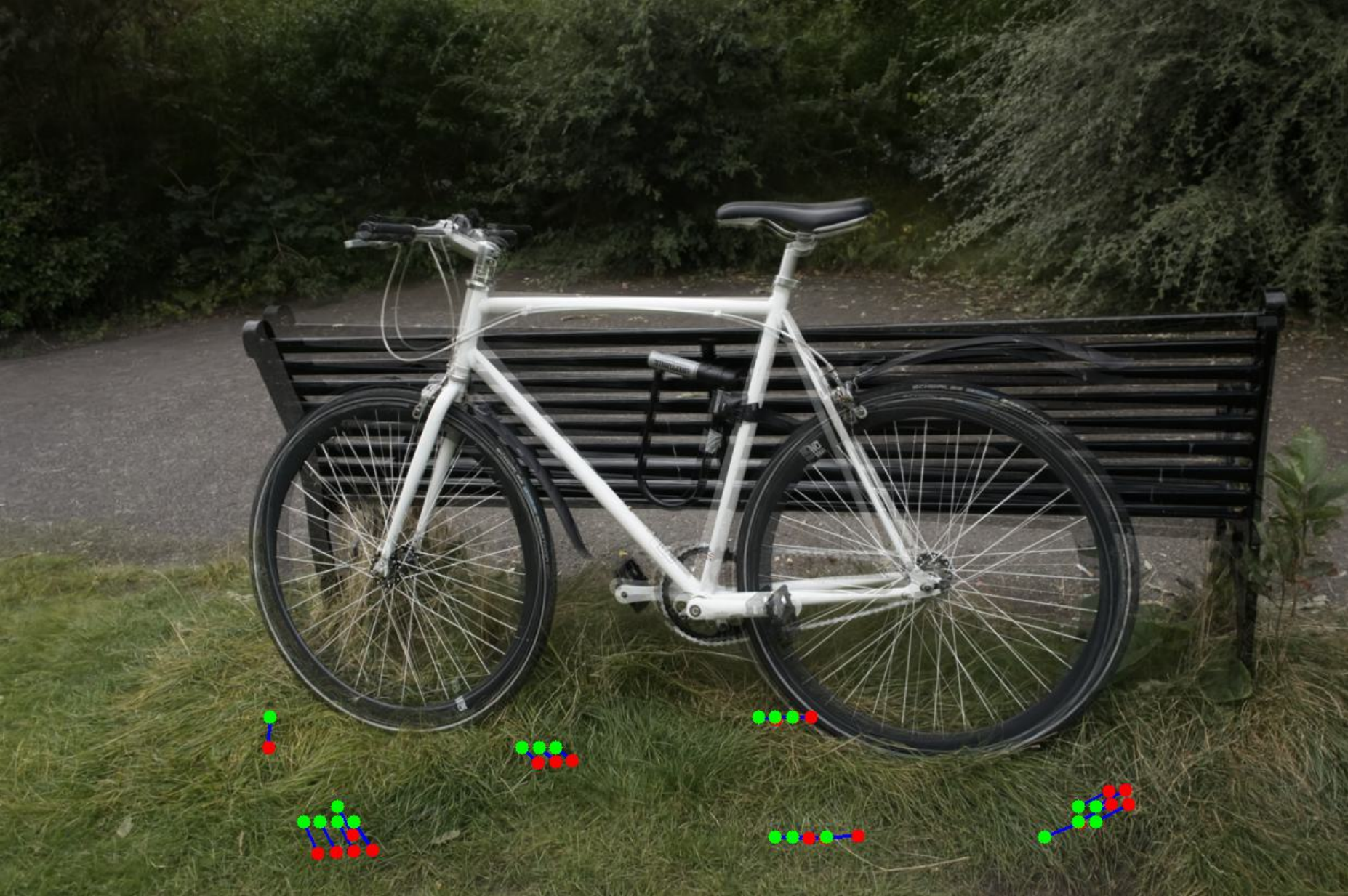} &
        \includegraphics[width=0.19\textwidth]{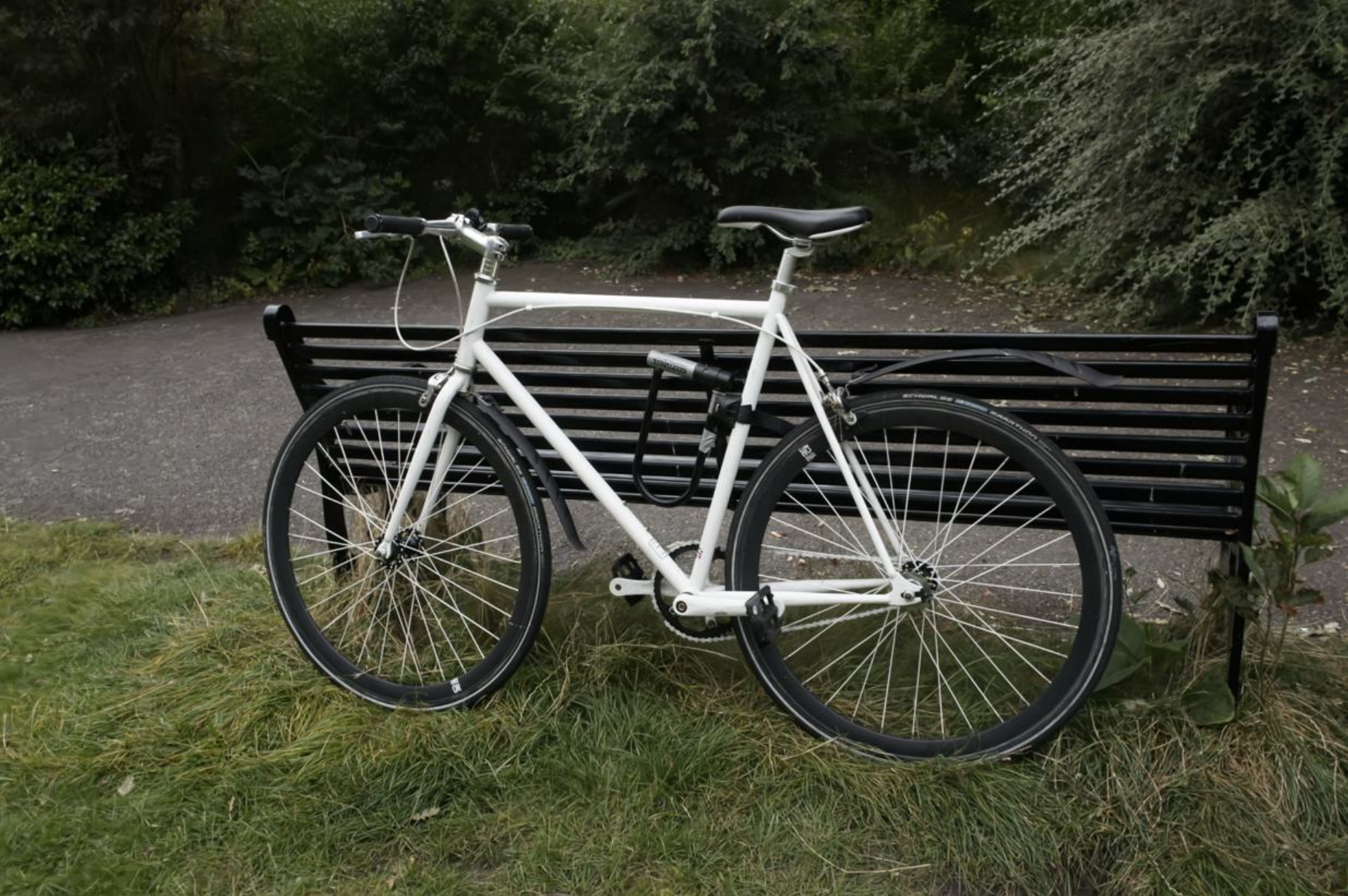} &
        \includegraphics[width=0.19\textwidth]{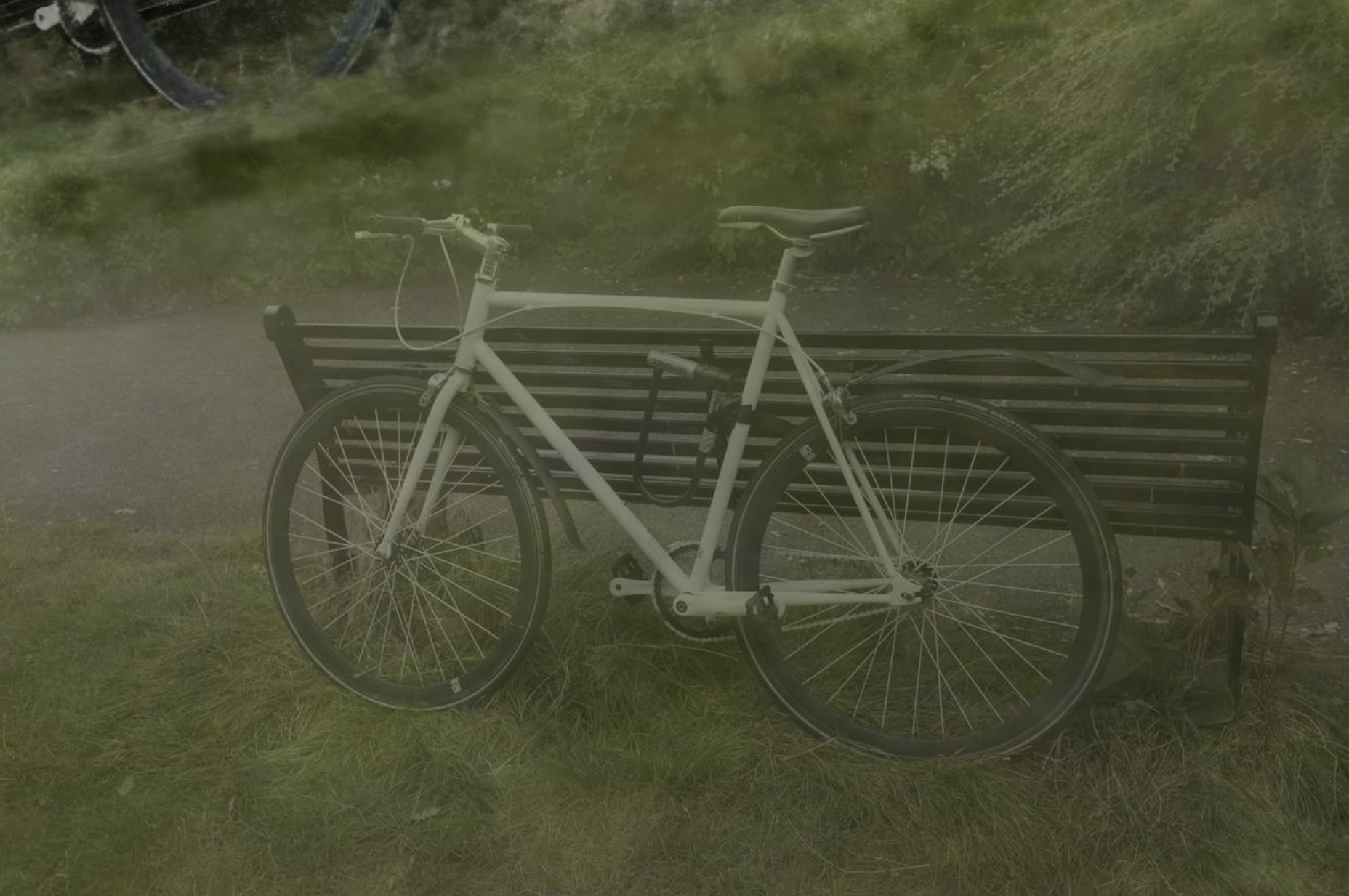} \\
        step=0 & step=10 & step=30 & step=300 & step=1000  \\
        \includegraphics[width=0.19\textwidth]{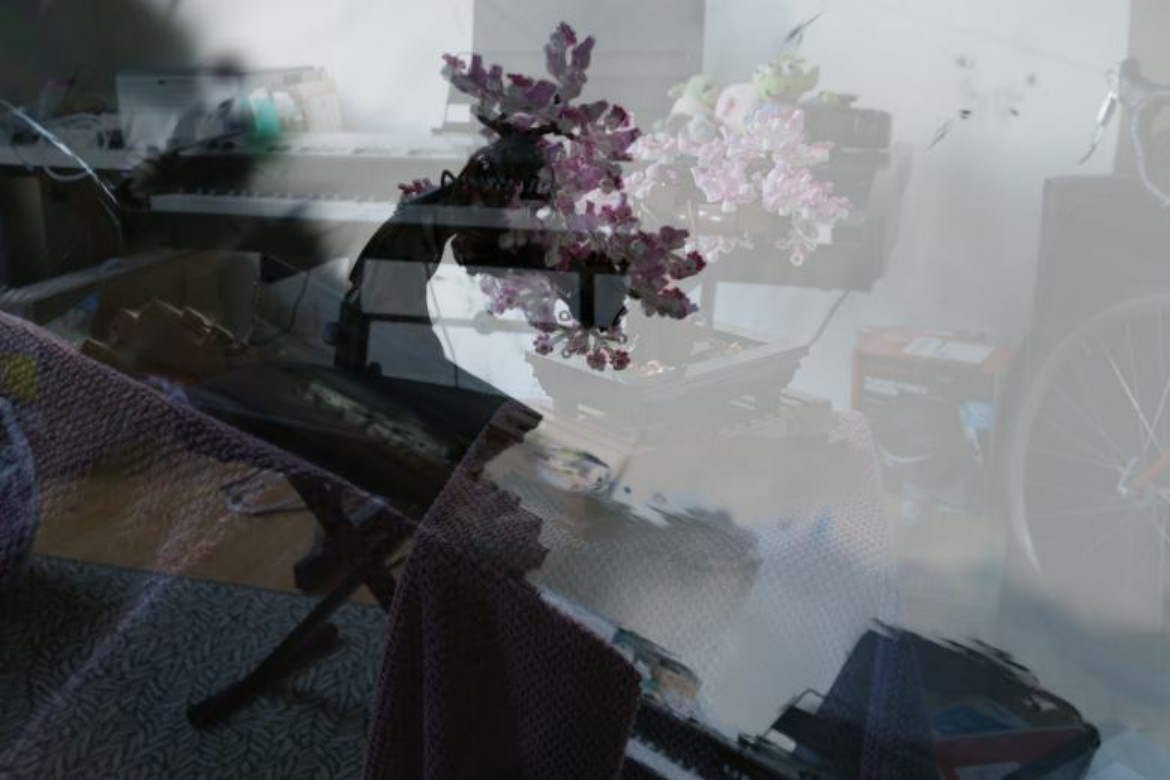} &
        \includegraphics[width=0.19\textwidth]{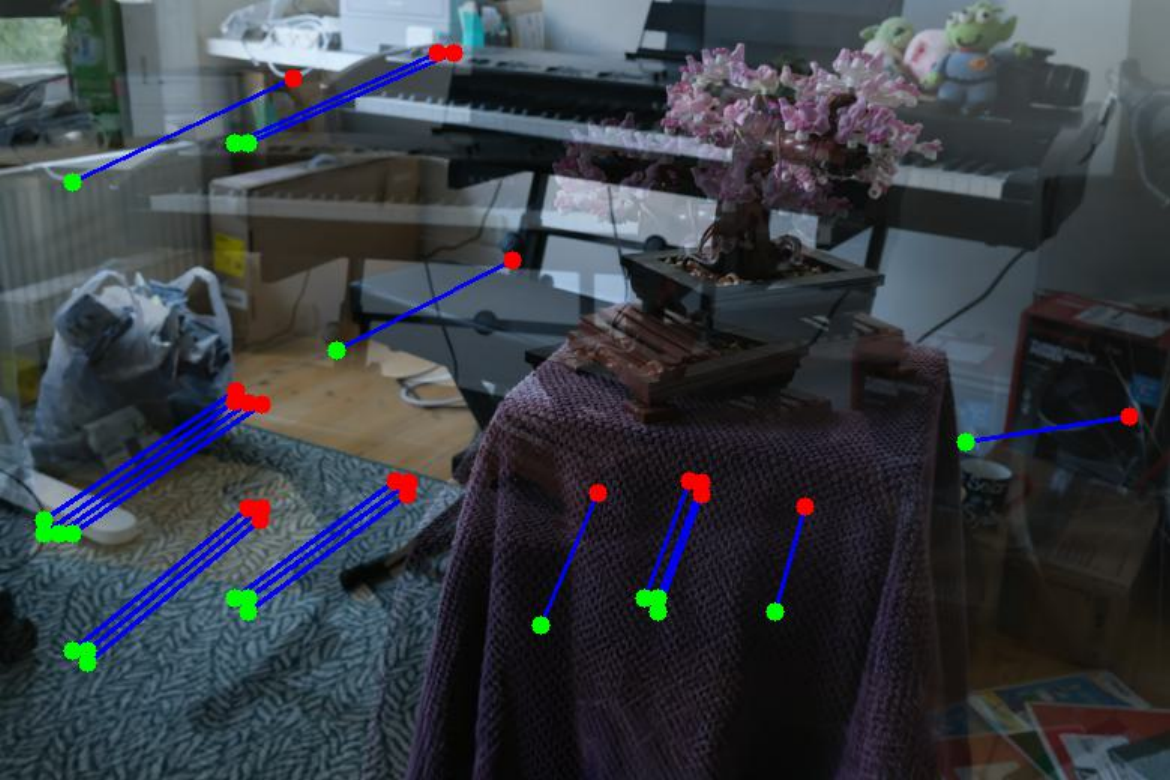} &
        \includegraphics[width=0.19\textwidth]{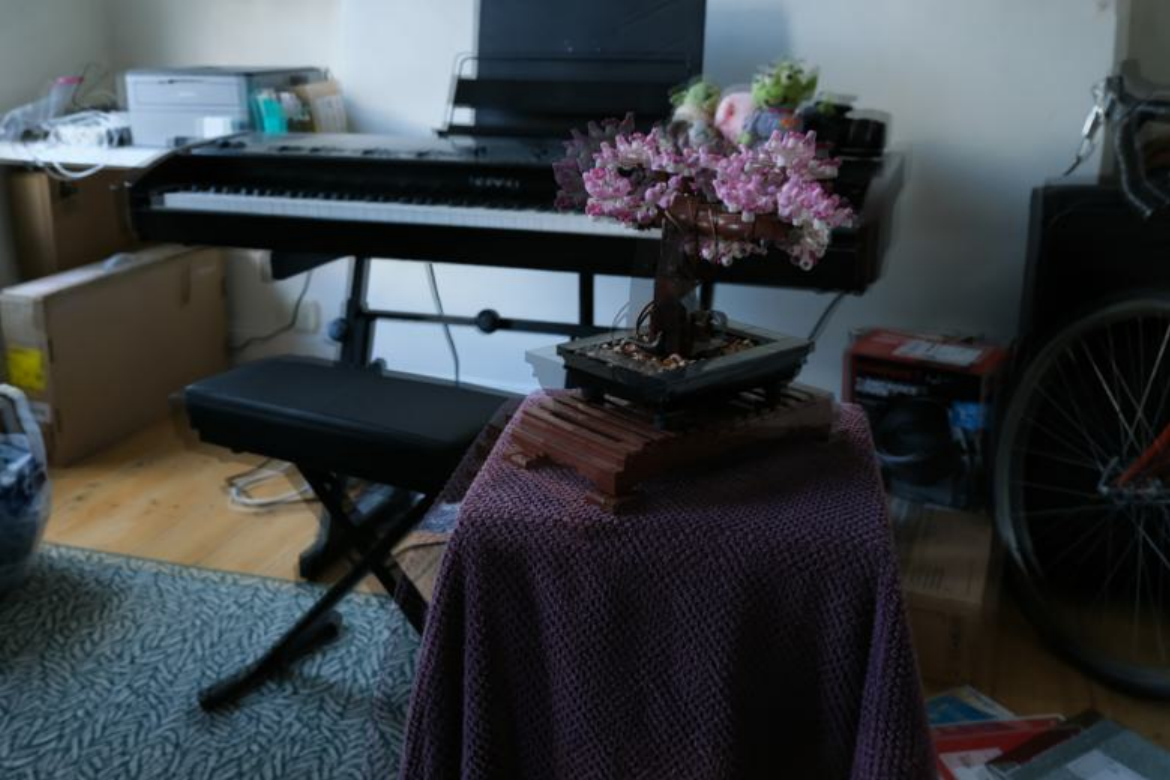} &
        \includegraphics[width=0.19\textwidth]{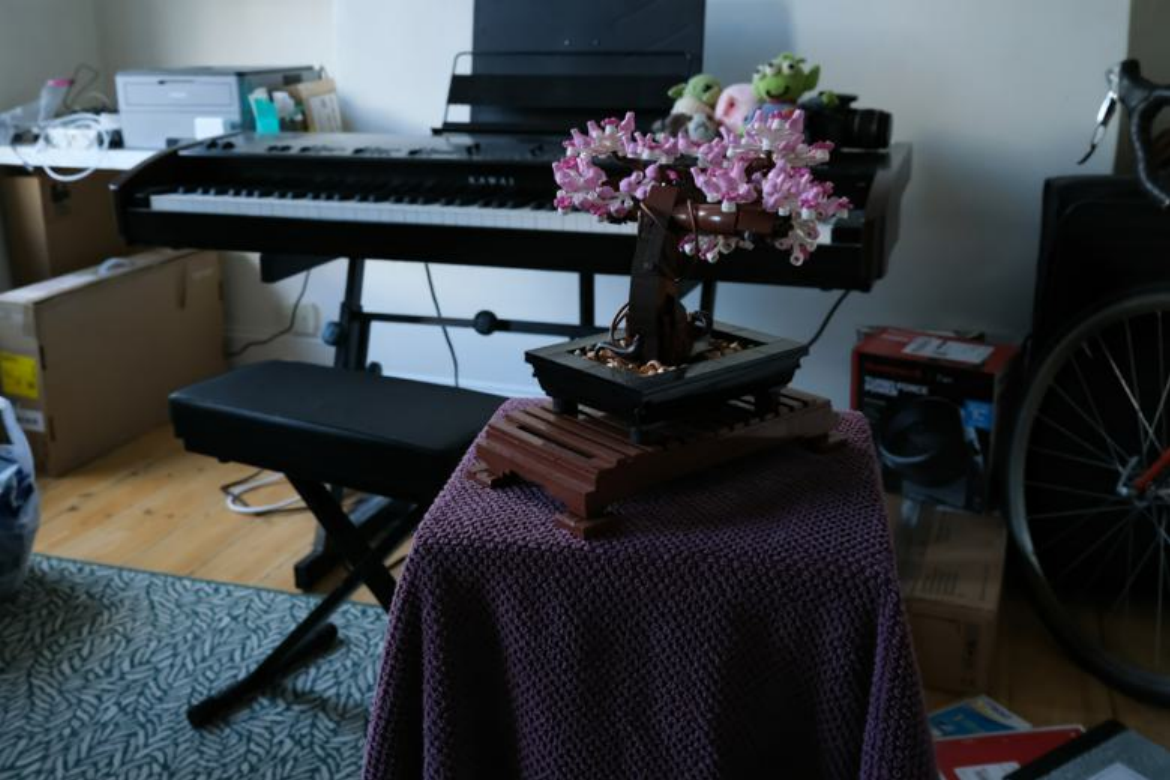} &
        \includegraphics[width=0.19\textwidth]{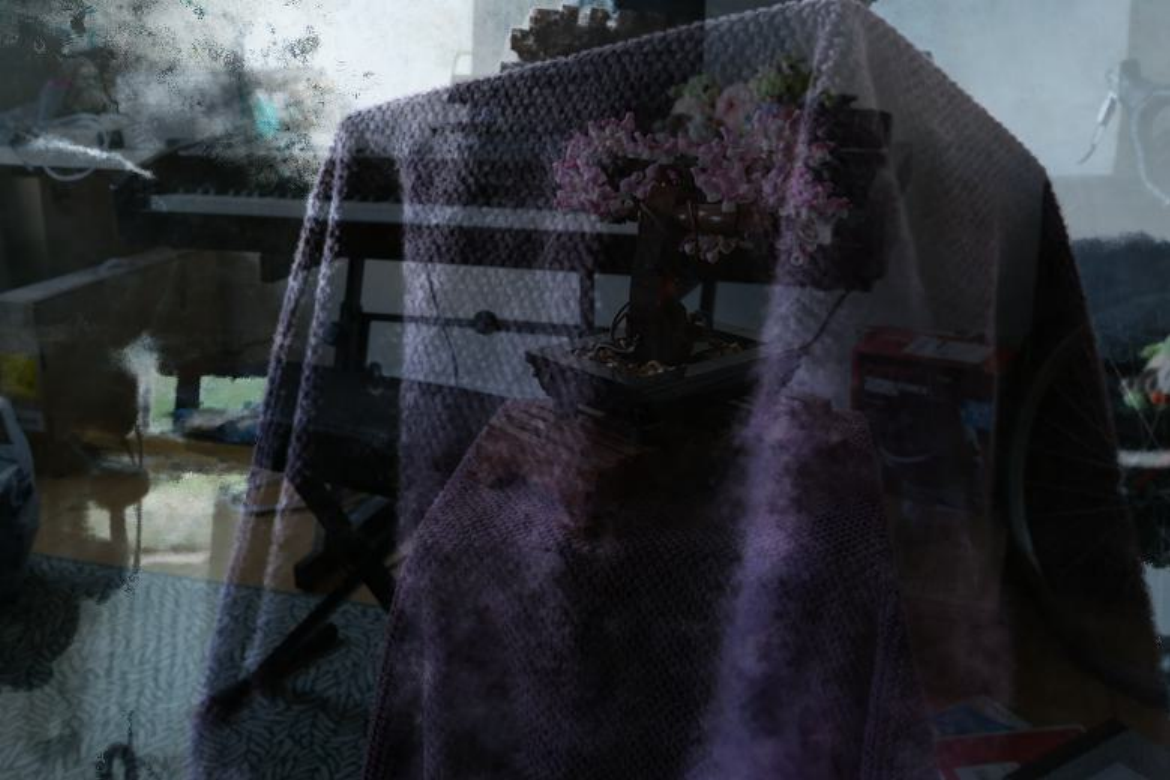} \\
        step=0 & step=20 & step=50 & step=200 & step=1000  \\
    \end{tabular}
    }
    \caption{Visualization results of pose estimation. Displaying the pose estimation results by overlaying query images and rendered images. It is noteworthy that for the initial poses, we only provide the rendering results from iComMa, as the initial poses for iNeRF are consistent. For iComMa, during the initial stages of pose optimization, we also visualize the top 20 keypoint matches with the highest confidence detected by LoFTR. Keypoints in the query image are depicted in \textcolor[RGB]{0, 255, 0}{green}, keypoints in the rendered image are depicted in \textcolor[RGB]{255, 0, 0}{red}, and \textcolor[RGB]{0, 0, 255}{blue} lines connect matching keypoints.}
    \label{fig_vis_com_inerf}
\end{figure}
\subsection{Comparison with iNeRF}
\label{sec_cwinerf}
 \subsubsection{Experimental Setting:}We investigate the performance of iNeRF and our method in iterative pose estimation across all datasets, considering various initial conditions. In all experiments, the main parameters of iNeRF are set as follows: batch\_size=2048, sampling\_strategy=interest\_regions. For each dataset, we randomly select five images. For each image, following the experimental setup of iNeRF, we initialize ten poses by randomly sampling an axis for rotation from the unit sphere. To facilitate a more nuanced comparison with iNeRF, we discretize the rotation angle and displacement bias into distinct intervals, aiming to evaluate the model's robustness across diverse conditions. Specifically, for the study of rotation, the initial rotation transformation $\delta_r$ is divided into three distinct degrees, with the specific interval values referenced in the legend of Fig. \ref{fig_com_inerf}. Additionally, we set the camera's translation along each axis within the range of $\pm\left[ 0, 0.05\right]$. Similarly, for the study of translation, after a rotation transformation of $\pm\left[ 0^\circ, 5^\circ \right]$ on the ground truth camera pose, three different degrees of translation transformation $\delta_t$ are applied. This meticulous categorization allowed for a comprehensive evaluation of the model's adaptability under various pose configurations. To ensure the fairness of the experiments, it is rigorously ensured that the inputs for all algorithms are consistent.

\subsubsection{Quantitative Comparison:} 
Generally, if the error is below $0.05$ meters and $5$ degrees, the pose estimation result is considered reliable, which is a widely accepted standard in the field \cite{yen2021inerf}. To conduct a more precise evaluation, we set the threshold for rotation error to $1$ degree and the threshold for translation error to $0.01$. Fig. \ref{fig_com_inerf} reports the performance of iComMa and iNeRF under different types of datasets. The horizontal axis represents the number of iterations, while the vertical axis indicates the proportion meeting a certain threshold condition.

For Fig. \ref{fig_com_inerf}, we have the following observations and analysis: Firstly, under challenging initialization conditions, iComMa demonstrates overwhelming superiority over iNeRF. Particularly, in LLFF datasets and complex $360^\circ$ scene datasets, iNeRF is basically unable to handle cases where initial rotation or translation magnitudes are excessively large, as illustrated in Fig. \ref{fig_com_inerf} (b) and (c), \etal. The proposed method exhibits varying degrees of robustness in addressing such situations across. This is attributed to the matching module of iComMa, which provides effective gradient information for pose optimization through the positional information of 2D keypoints. Secondly, in terms of pose estimation accuracy, iComMa also exhibits outstanding performance. For example, in synthetic datasets where $\delta_{r} \in \pm [0^\circ,20^\circ]$, at a threshold of $5^\circ$, iNeRF achieves a performance level of 80\%. However, when the evaluation criterion becomes stricter, set to $1^\circ$ as shown in Fig. \ref{fig_com_inerf} (a), this proportion sharply drops to approximately 30\%. In contrast, iComMa maintains a high success rate even under more stringent evaluation criteria.  On the one hand, this is benefited by the powerful scene representation and high-quality rendering capability of 3DGS. On the other hand, iNeRF is constrained by computational speed, as it only calculates the residuals for sampled pixels in each iteration, while iComMa compares all pixels, utilizing global information. Finally, iComMa requires fewer iterations to achieve a pose with small errors, which also gives it an advantage in terms of pose optimization speed.
\begin{table}[b]
\caption{Time Consumption Statistics.
 The \textbf{best} performance is indicated in bold.}
\label{table_time}
\centering
\begin{tabular}{@{}p{2.5cm}|*{3}{>{\centering\arraybackslash}p{1.5cm}}|*{3}{>{\centering\arraybackslash}p{1.5cm}}@{}}
\toprule
        & \multicolumn{3}{c}{$\delta_r \in \pm{[}10^\circ,20^\circ{]}$} & \multicolumn{3}{c}{$\delta_r \in \pm{[}20^\circ,40^\circ{]}$} \\ \midrule
        Methods & iNeRF     & iComMa     & $\mathrm{iComMa}_{c}$     & iNeRF       & iComMa       & $\mathrm{iComMa}_{c}$       \\
\hline
Success Rate (\%) & 90.5      & \bf 100        & \bf 100               & 65.5        & \bf 100          & 96            \\
Time (s) & 12.66     & 1.13       & \bf 0.75              & 23.33       & 1.57         & \bf 0.83          \\ \bottomrule
\end{tabular}
\end{table}

\subsubsection{Visualization Results:} Fig. \ref{fig_vis_com_inerf} visually shows  the results of pose estimation across different types of datasets. A higher degree of overlap between the query image and the rendered image indicates a more accurate pose estimation. It is evident that under small pose differences, as shown in the results of the \textit{mic} dataset, both iComMa and iNeRF can effectively perform pose estimation. However, when facing large differences in initial poses or complex scenes, iNeRF struggles to complete the task. In contrast, iComMa utilizes the information of matching points in the initial stage to optimize a pose with small errors in very few iterations. When the distance between 2D matching points is small, the strategy of utilizing comparing alone enables rapid and precise optimization of pose estimation. 

\subsubsection{Comparison of Computational Time:} 
We conduct timing statistics on pose estimation for four synthetic datasets, namely \textit{chair}, \textit{drums}, \textit{hotdog}, and \textit{lego}, each with images resolution of $800 \times 800$ pixels. Similarly, five ground truth poses are  randomly sampled, and ten initial transformations are applied to each pose. Two scales of $\delta_r$ are considered: $\pm [10^\circ,20^\circ]$ and $\pm [20^\circ,40^\circ]$, while $\delta_t \in \pm [0,0.2]$. A successful case is defined as one where the rotation error is less than $5^\circ$ and the translation error is less than 0.05. All experiments are conducted using a single NVIDIA RTX A6000 GPU. Table \ref{table_time} reports the proportion of successful pose estimations and the average time taken to achieve the aforementioned thresholds for successful samples. It is evident that iNeRF consumes approximately ten times more time than iComMa. Specifically, $\mathrm{iComMa}_{c}$, which computes only the comparing loss, achieves speeds of less than 1 second and maintains high accuracy even with small pose differences. Due to the explicitly computed gradients of poses during the render-and-compare process, rather than through automatic differentiation of neural networks, this stage exhibits a significantly faster computational speed compared to iNeRF.

\subsection{Relative Pose Estimation}
\label{sec_rpe}
\subsubsection{Experimental Setting:} In this study, we quantitatively compared our proposed method with several matching-based pose estimation methods, namely, LightGlue \cite{lindenberger2023lightglue}, MatchFormer \cite{wang2022matchformer}, and LoFTR \cite{sun2021loftr}. 
Pairs of images are chosen as inputs for relative pose estimation, and the rotational differences $\delta_r$ between these pairs are categorized into intervals of $\pm [0^\circ,20^\circ]$, $\pm [20^\circ,40^\circ]$, and $\pm [40^\circ,60^\circ]$. Fifty pairs of images are randomly selected for each scale within each dataset. Due to the limited number of samples provided by the LLFF dataset, we only utilized eight synthetic datasets and eleven complex $360^\circ$ scene datasets in this experiment. For matching-based method, we use RANSAC to compute the essential matrix from predicted matches, and then calculate the camera pose.
\subsubsection{Quantitative Comparison:}  In Table \ref{table_rpe_sy} and Table \ref{table_rpe_rw}, we report the AUC values of angle errors at thresholds of $5^\circ$, $10^\circ$, and $20^\circ$. Obviously, our method consistently outperforms three contrastive algorithms based on matching across various datasets and under different initialization conditions. Under challenging initialization conditions, the performance of the three comparative methods is notably deficient. Additionally, they also encounter challenges related to insufficient precision in pose estimation. For instance, on the synthetic dataset under the condition of $\delta_r\in \pm{[}0^\circ,20^\circ{]}$, LightGlue achieves a commendable AUC@20 of 86.55, yet its AUC@5 is merely 61.76. In contrast, iComMa maintains both these metrics above 90.

Using the RANSAC algorithm directly for computing relative pose imposes strict requirements on the accuracy of matched points. Particularly in scenarios with significant pose disparities, achieving accurate feature matching in the target scene becomes increasingly challenging. Furthermore, despite some degree of generalization exhibited by these matching point detection models, they still fail to universally apply to all scenes, resulting in decreased precision of matched points. Moreover, for a monocular camera with known intrinsic parameters, in the absence of CAD models, employing only RANSAC and acquired 2D matching points cannot determine the camera's relative translational component.

\begin{table}[tb]
\centering
\caption{Relative pose estimation results on synthetic datasets. The AUC of the angle error at $5^\circ$, $10^\circ$, and $20^\circ$ is reported. The \textbf{best} performance is highlighted in bold.}
\label{table_rpe_sy}
\resizebox{\textwidth}{!}{
\begin{tabular}{@{}l|ccc|ccc|ccc@{}}
\toprule
            & \multicolumn{3}{c}{$\delta_r\in \pm{[}0^\circ,20^\circ{]}$} & \multicolumn{3}{c}{$\delta_r\in \pm{[}20^\circ,40^\circ{]}$} & \multicolumn{3}{c}{$\delta_r\in \pm{[}40^\circ,60^\circ{]}$} \\
\hline
            & \scriptsize AUC@5      & \scriptsize AUC@10     & \scriptsize AUC@20     & \scriptsize AUC@5      & \scriptsize AUC@10      & \scriptsize AUC@20     & \scriptsize AUC@5      & \scriptsize AUC@10      & \scriptsize AUC@20     \\
\hline
MatchFormer & 39.65      & 55.92      & 70.94      & 21.09      & 33.16       & 44.56      & 6.08       & 12.48       & 19.44      \\
LightGlue   & 61.76      & 76.93      & 86.55      & 42.24      & 58.71       & 70.56      & 17.65      & 27.58       & 36.82      \\
LoFTR       & 59.27      & 75.34      & 85.56      & 31.85      & 45.98       & 58.41      & 8.07       & 13.93       & 19.77      \\
iComMa      & \bf 90.95      & \bf 94.02      & \bf 96.71      & \bf 74.65      & \bf 78.40       & \bf 82.97      & \bf 37.93      & \bf 39.87       & \bf 43.98  \\
\bottomrule
\end{tabular}
}
\end{table}
\begin{table*}[tb]
\centering
\caption{Relative pose estimation results on $360^\circ$ scene datasets. The AUC of the angle error at $5^\circ$, $10^\circ$, and $20^\circ$ is reported. The \textbf{best} performance is highlighted in bold.}
\label{table_rpe_rw}
\resizebox{\textwidth}{!}{
\begin{tabular}{@{}l|ccc|ccc|ccc@{}}
\toprule
            & \multicolumn{3}{c}{$\delta_r\in \pm{[}0^\circ,20^\circ{]}$} & \multicolumn{3}{c}{$\delta_r\in \pm{[}20^\circ,40^\circ{]}$} & \multicolumn{3}{c}{$\delta_r\in \pm{[}40^\circ,60^\circ{]}$} \\
\hline
            & \scriptsize AUC@5      & \scriptsize AUC@10     & \scriptsize AUC@20     & \scriptsize AUC@5      & \scriptsize AUC@10      & \scriptsize AUC@20     & \scriptsize AUC@5      & \scriptsize AUC@10      & \scriptsize AUC@20     \\
\hline
MatchFormer & 13.11      & 34.86      & 60.61      & 0.39       & 4.93        & 28.06      & 0.10       & 0.80        & 7.78       \\
LightGlue   & 13.77      & 37.92      & 64.25      & 0.65       & 5.48        & 31.51      & 0.23       & 1.40        & 10.32      \\
LoFTR       & 13.64      & 35.55      & 61.82      & 0.42       & 4.67        & 30.03      & 0.14       & 1.17        & 10.04      \\
iComMa      & \bf 84.41      & \bf 86.39      & \bf 88.87      & \bf 73.27      & \bf 75.12       & \bf 78.30      & \bf 61.75      & \bf 63.50       & \bf 65.68  \\
\bottomrule
\end{tabular}
}
\end{table*}
\begin{table}[tb]
\caption{Experimental Results of Ablation. "w/o" denotes "without". $\mathrm{mean}_r$ and $\mathrm{mean}_t$ respectively denote the mean rotation error and mean translation error of successful samples. The \underline{poorest} performance is indicated by an underline.}
\label{table_abl}
\centering
{\small
\begin{tabular}{p{2.3cm}|>{\centering}p{1.6cm}*{2}{>{\centering\arraybackslash}p{1.5cm}}|>{\centering}p{1.6cm}*{2}{>{\centering\arraybackslash}p{1.5cm}}}
\toprule
              & \multicolumn{3}{c}{synthetic}     & \multicolumn{3}{c}{$360^\circ$ scene} \\ \midrule
              
              &\footnotesize  failure rate&\footnotesize $\mathrm{mean}_r$ &\footnotesize $\mathrm{mean}_t$  &\footnotesize failure rate &\footnotesize $\mathrm{mean}_r$    &\footnotesize $\mathrm{mean}_t$  \\
\hline
iComMa        & 0.090          & 0.012   & 0.001   & 0.095            & 0.024     & 0.002    \\
w/o Comparing & 0.140         & \underline{0.064}   & \underline{0.006}   & 0.145           & \underline{0.043}     & \underline{0.004}    \\
w/o Matching  & \underline{0.380}         & 0.011   & 0.001   & \underline{0.385}           & 0.030     & 0.002    \\ \bottomrule
\end{tabular}
}
\end{table}

\subsection{Ablation Study}
 \label{sec_abl}
In this section, we designed ablation experiments to demonstrate the effectiveness of the two components of iComMa, namely matching and comparing, and their respective roles. Four synthetic datasets, namely \textit{chair}, \textit{drums}, \textit{ficus}, and \textit{lego}, as well as four complex $360^\circ$ scene datasets, namely \textit{bicycle}, \textit{Playroom}, \textit{DrJohnson}, and \textit{room}, are utilized. We select 5 ground truth camera poses per dataset and apply 10 random transformations to each pose to simulate the initial camera positions. We set a significant initial bias, specifically, the initial rotation angles are within the interval of $\pm [30^\circ,50^\circ]$, while the translation transformations for synthetic datasets are within the interval of $\pm [0,0.6]$, and for $360^\circ$ scene datasets, they are within the interval of $\pm [0,0.2]$.

The threshold values for rotation error and translation error are set to $5^\circ$ and 0.05 respectively. The failure rate and the mean error of successful cases are recorded in Table \ref{table_abl}. From the data in the table, it is evident that the failure rate significantly increases without the matching module, reaching nearly 40\% on both types of datasets. This clearly demonstrates the efficacy of our matching module in effectively countering adverse initialization conditions. Observations from the mean error values of successful cases indicate that relying solely on the matching module without the comparing strategy can still achieve a relatively good pose estimation performance, albeit with slightly less precision. 

\begin{figure}[tb]
    \centering
    {\scriptsize
    \begin{tabular}{cccc|c}
        \multicolumn{4}{c|}{ \textbf{Camera Pose Estimation}} &  \textbf{Detail} \\
        \includegraphics[width=0.18\textwidth]{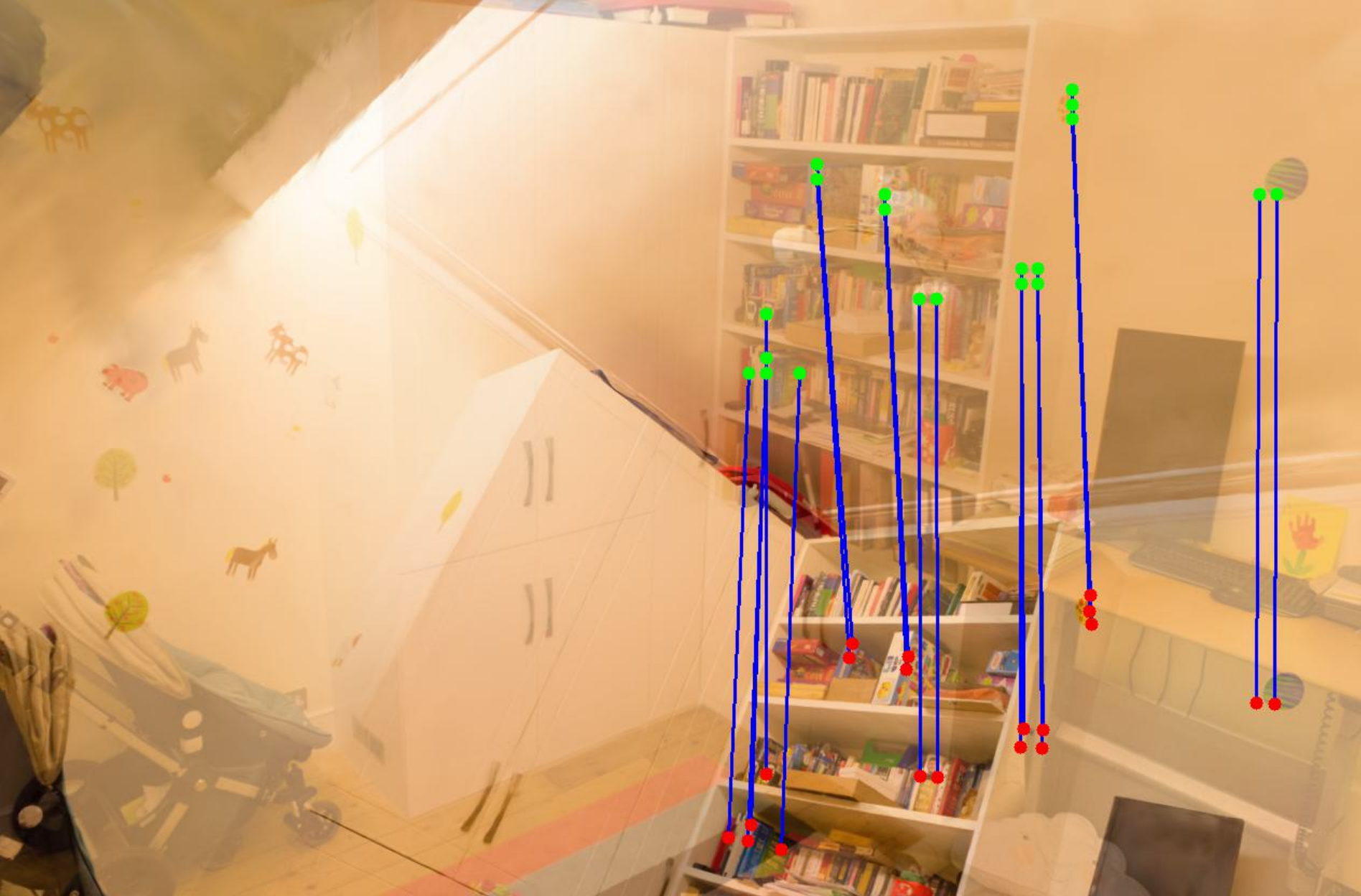} &
        \includegraphics[width=0.18\textwidth]{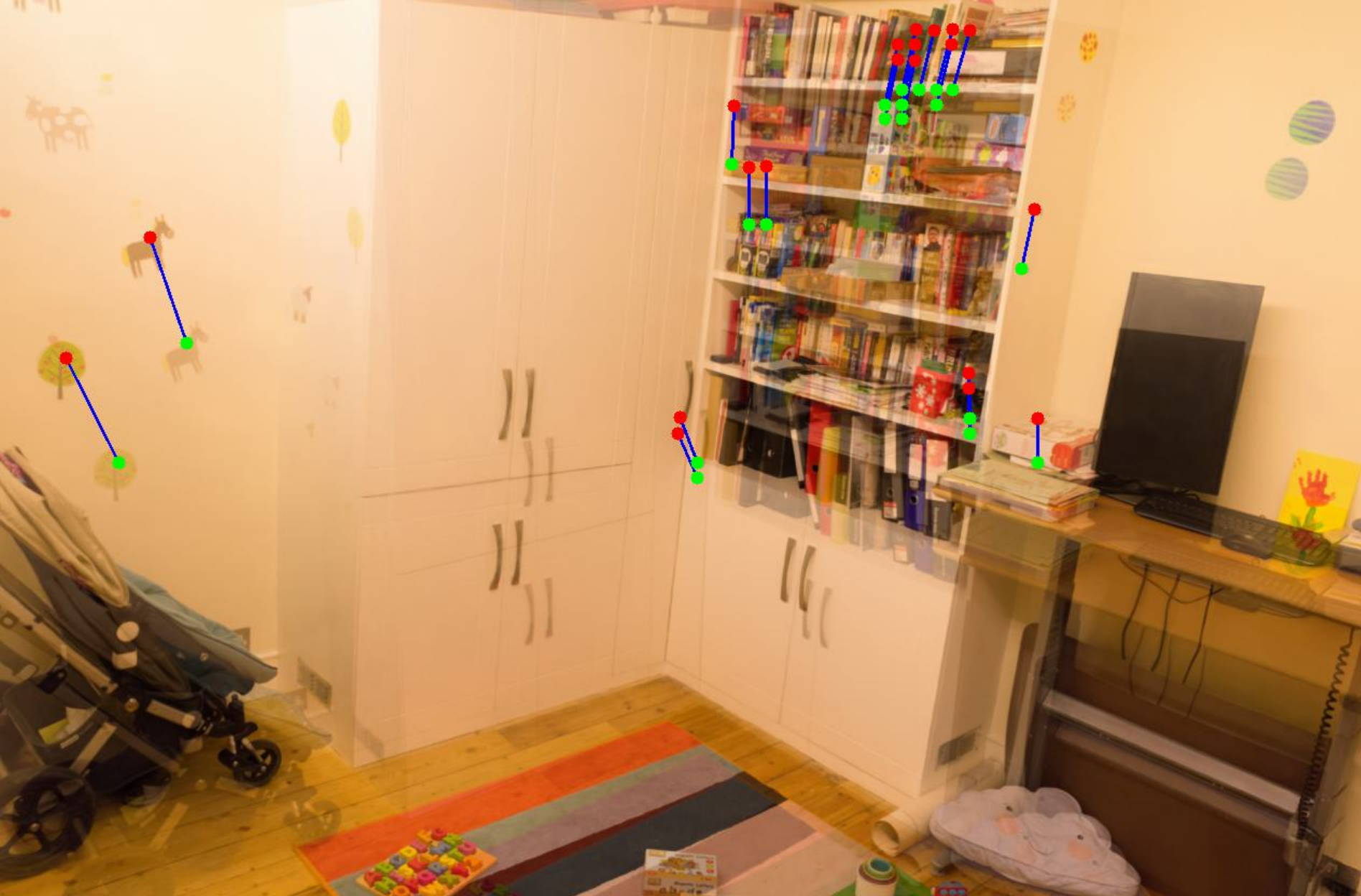} &
        \includegraphics[width=0.18\textwidth]{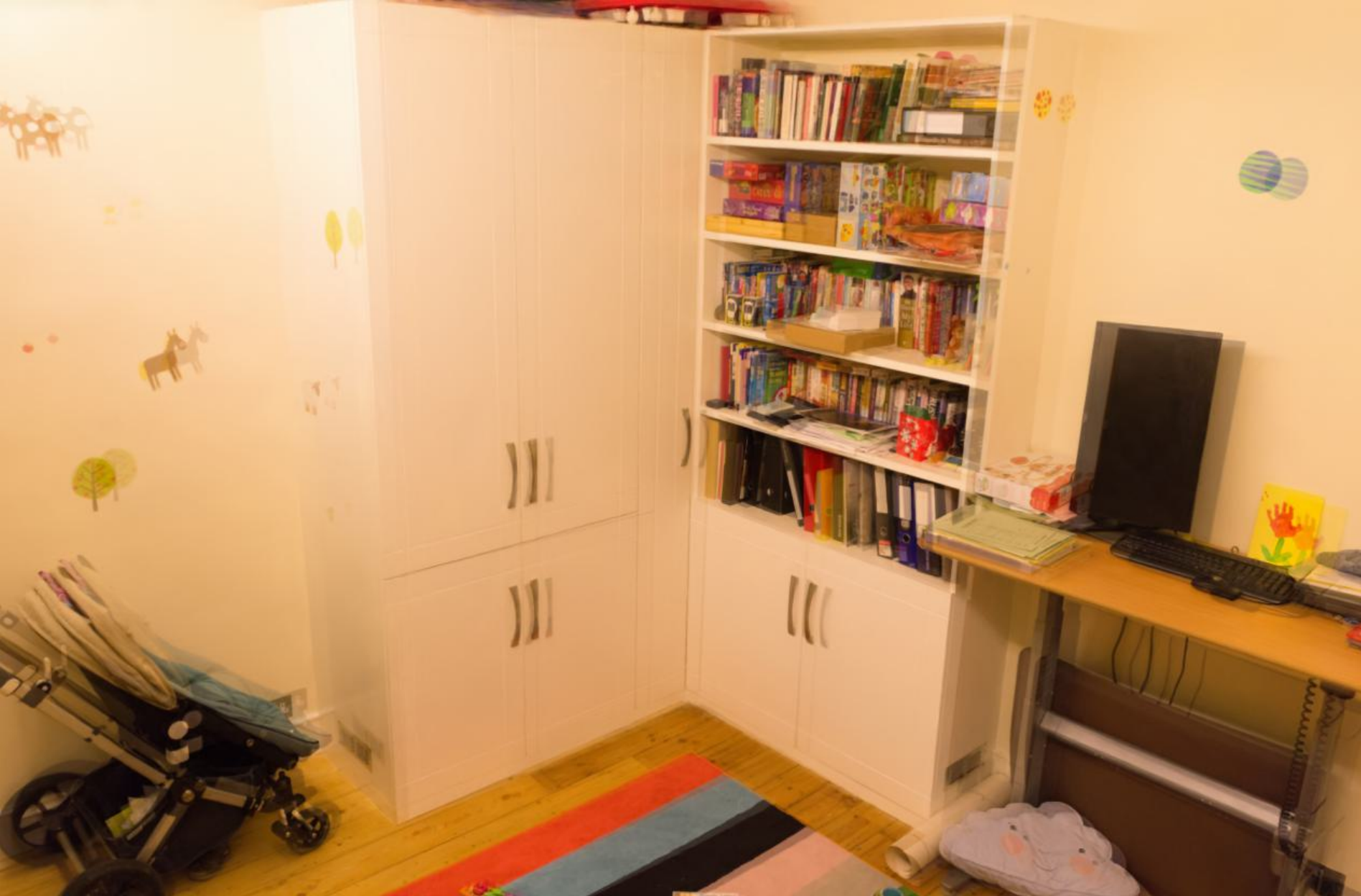} &
        \includegraphics[width=0.18\textwidth]{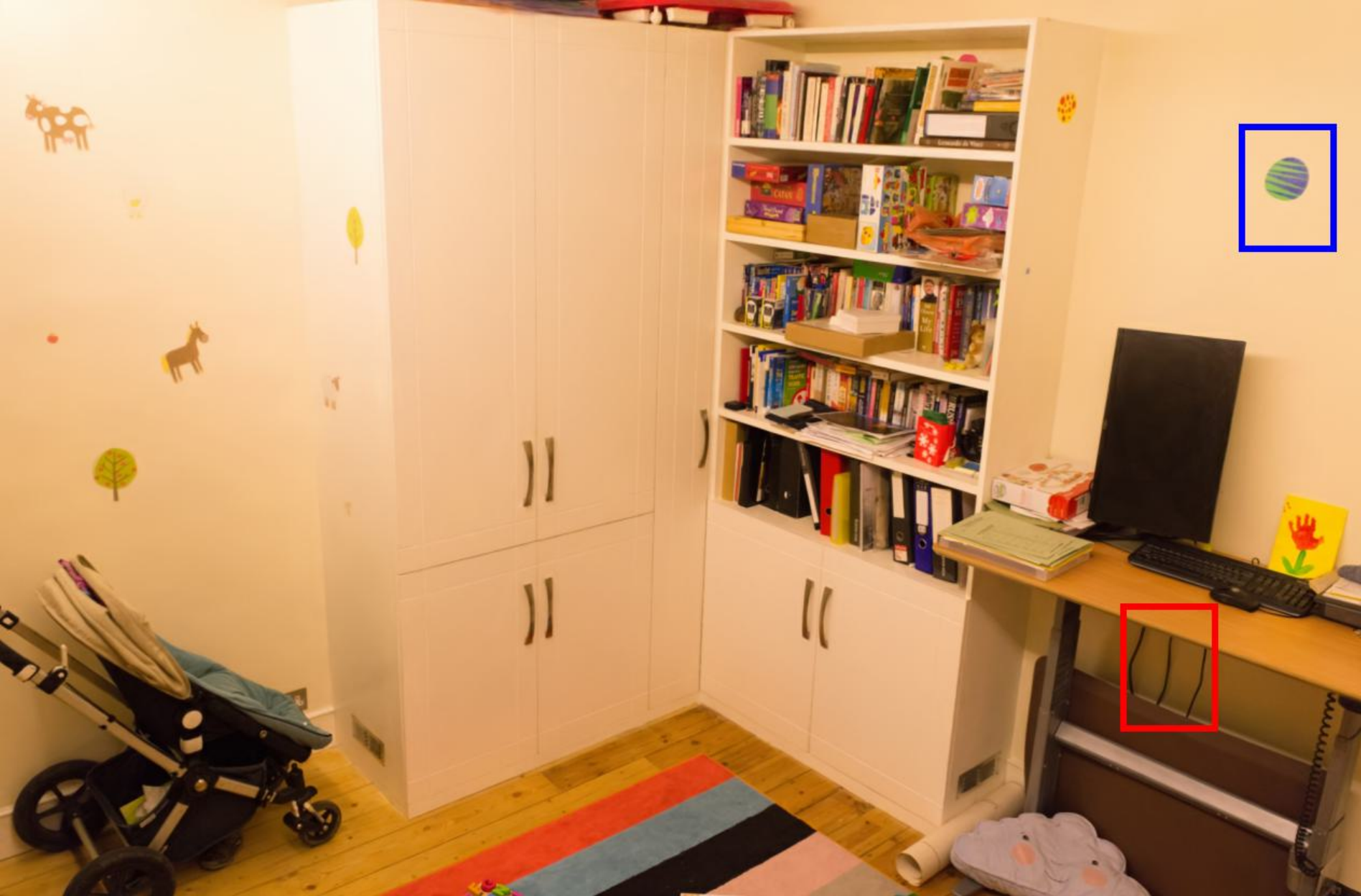} &
        \includegraphics[width=0.18\textwidth]{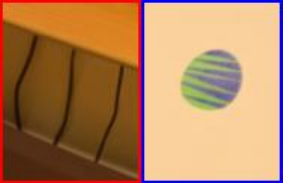} \\
        step=0 & step=20 & step=100 & step=200 &  iComMa\\
        \includegraphics[width=0.18\textwidth]{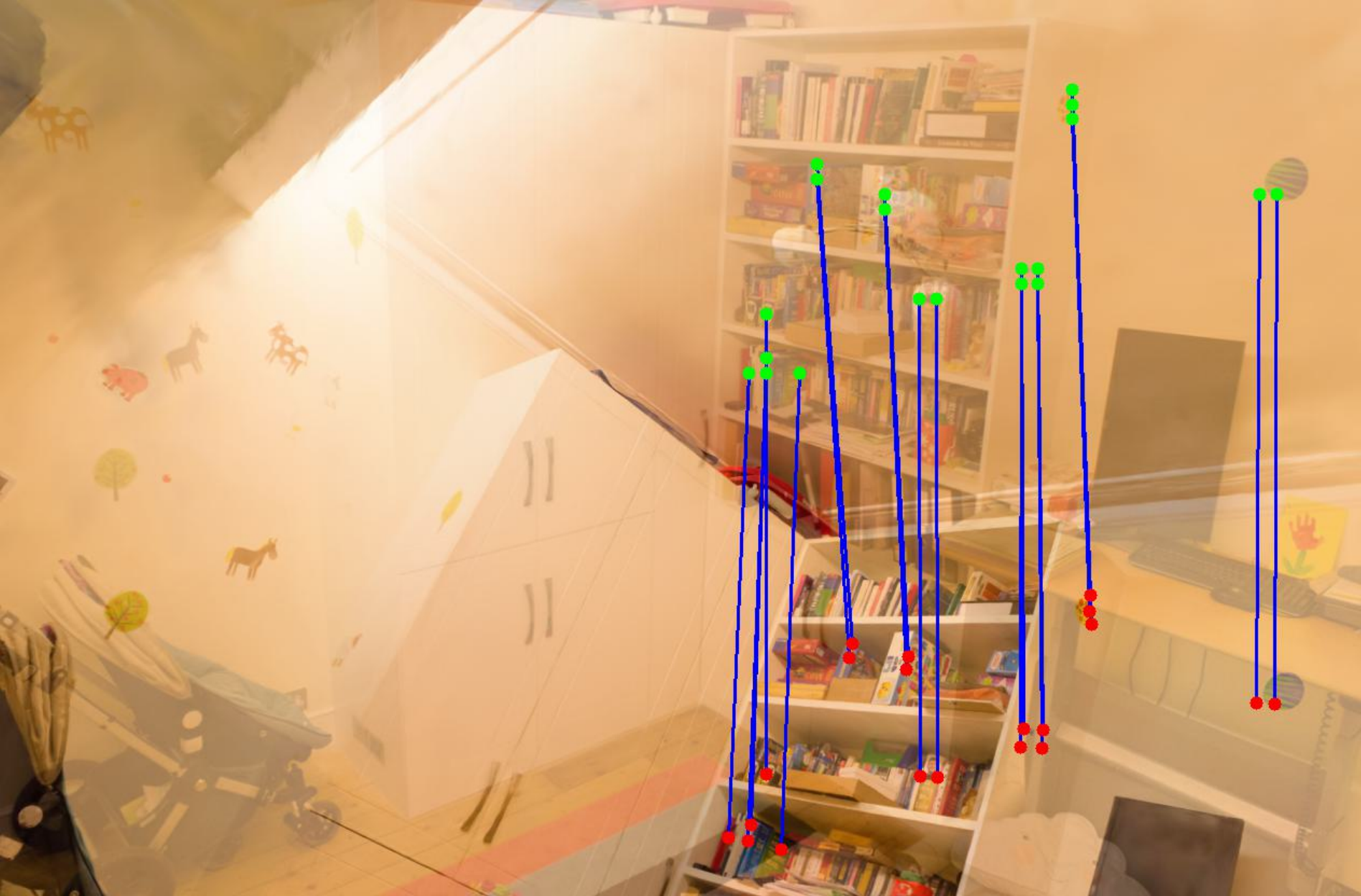} &
        \includegraphics[width=0.18\textwidth]{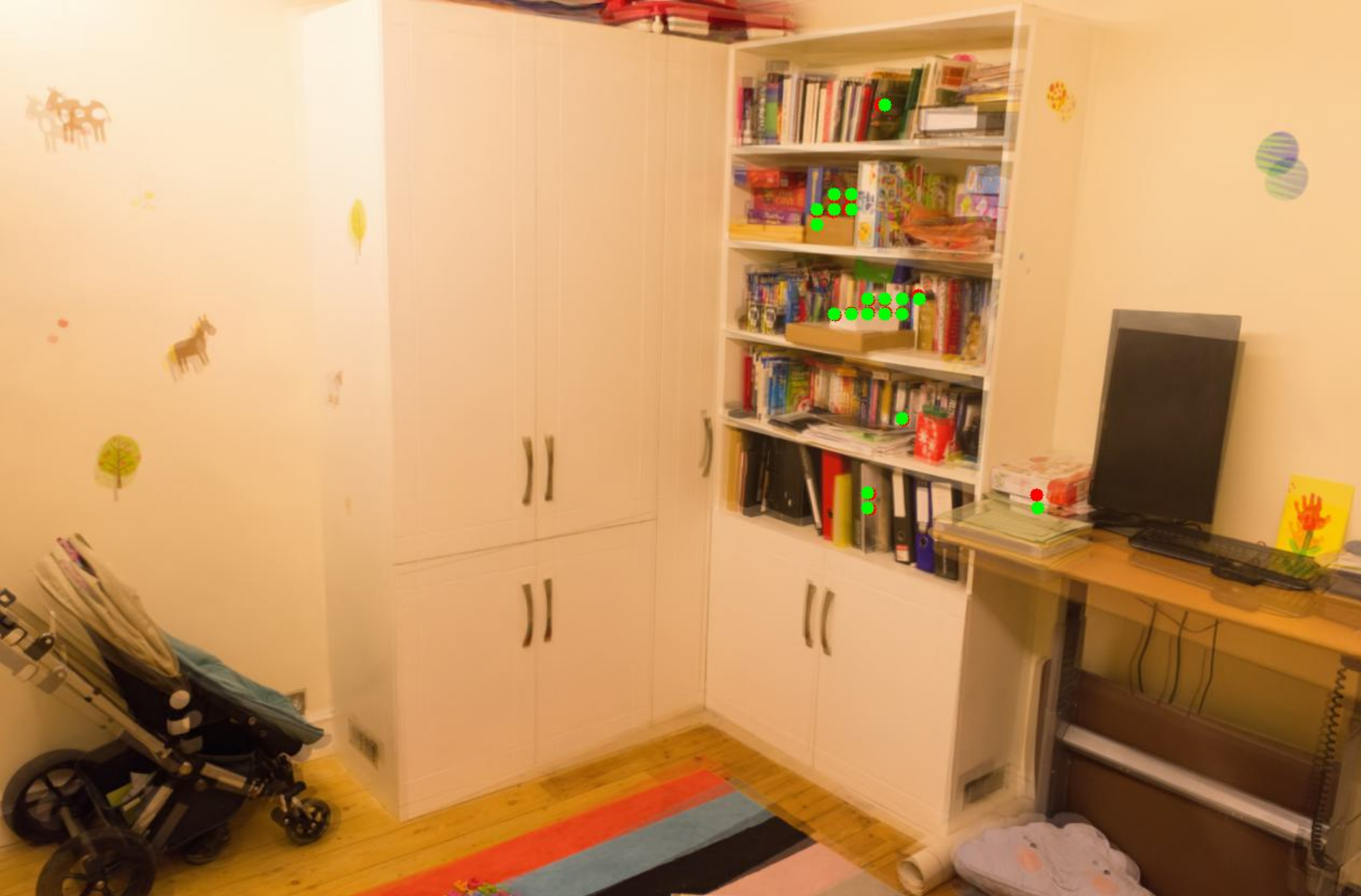} &
        \includegraphics[width=0.18\textwidth]{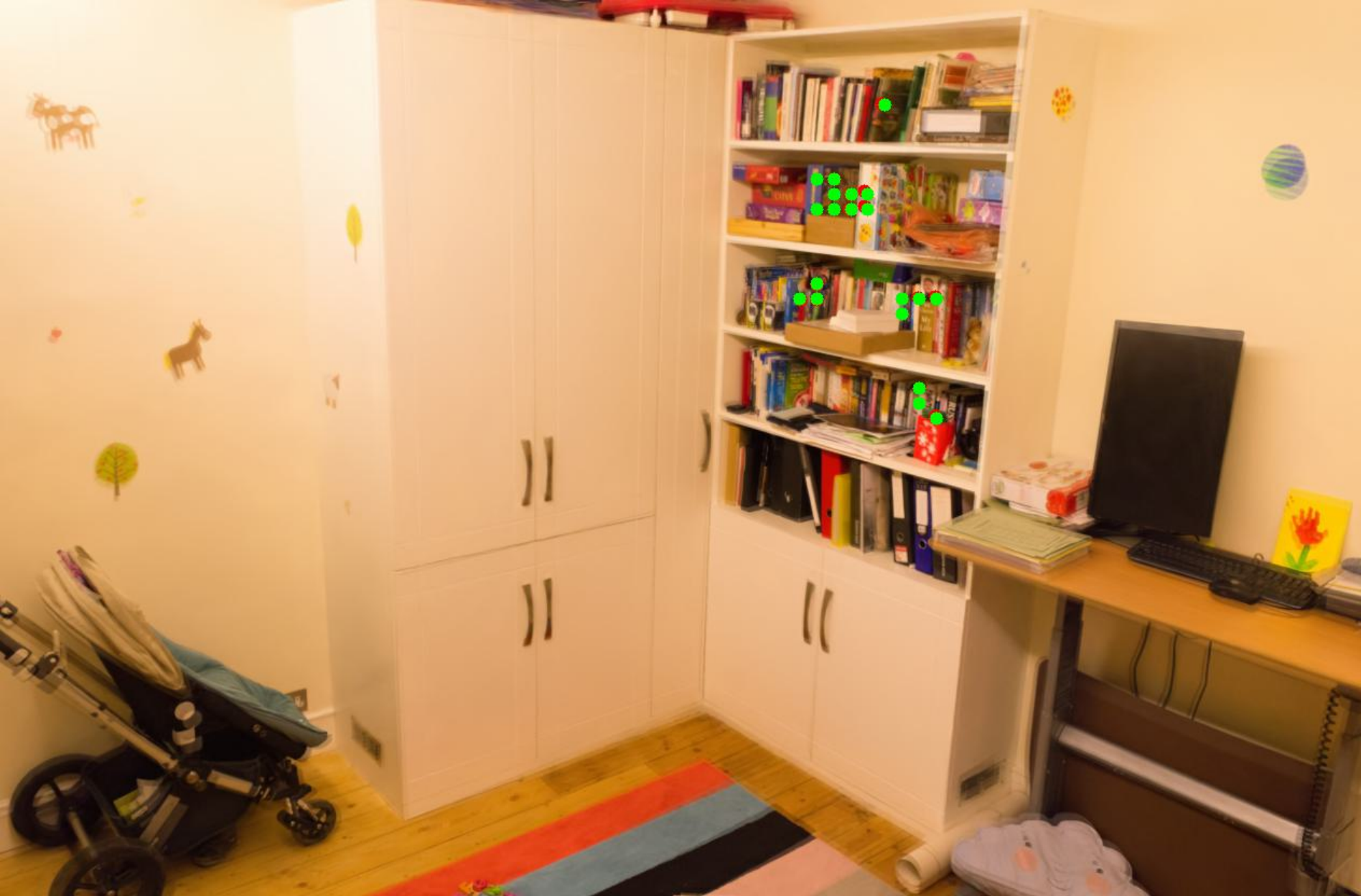} &
        \includegraphics[width=0.18\textwidth]{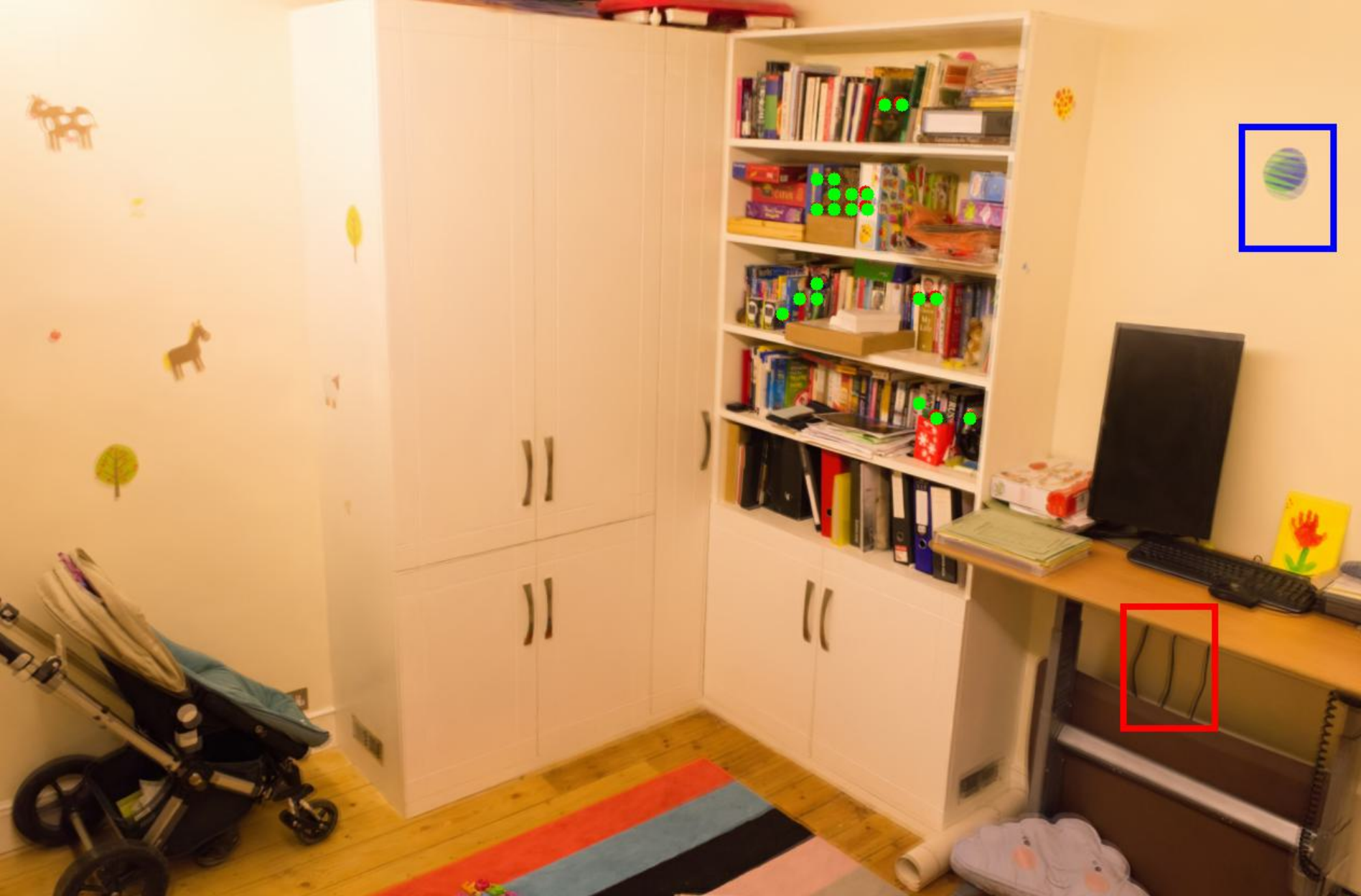} &
        \includegraphics[width=0.18\textwidth]{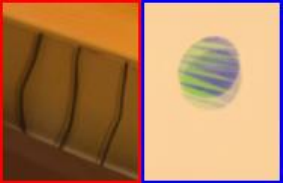} \\
         step=0 & step=200 & step=400 & step=500 &  w/o Comparing\\
        \includegraphics[width=0.18\textwidth]{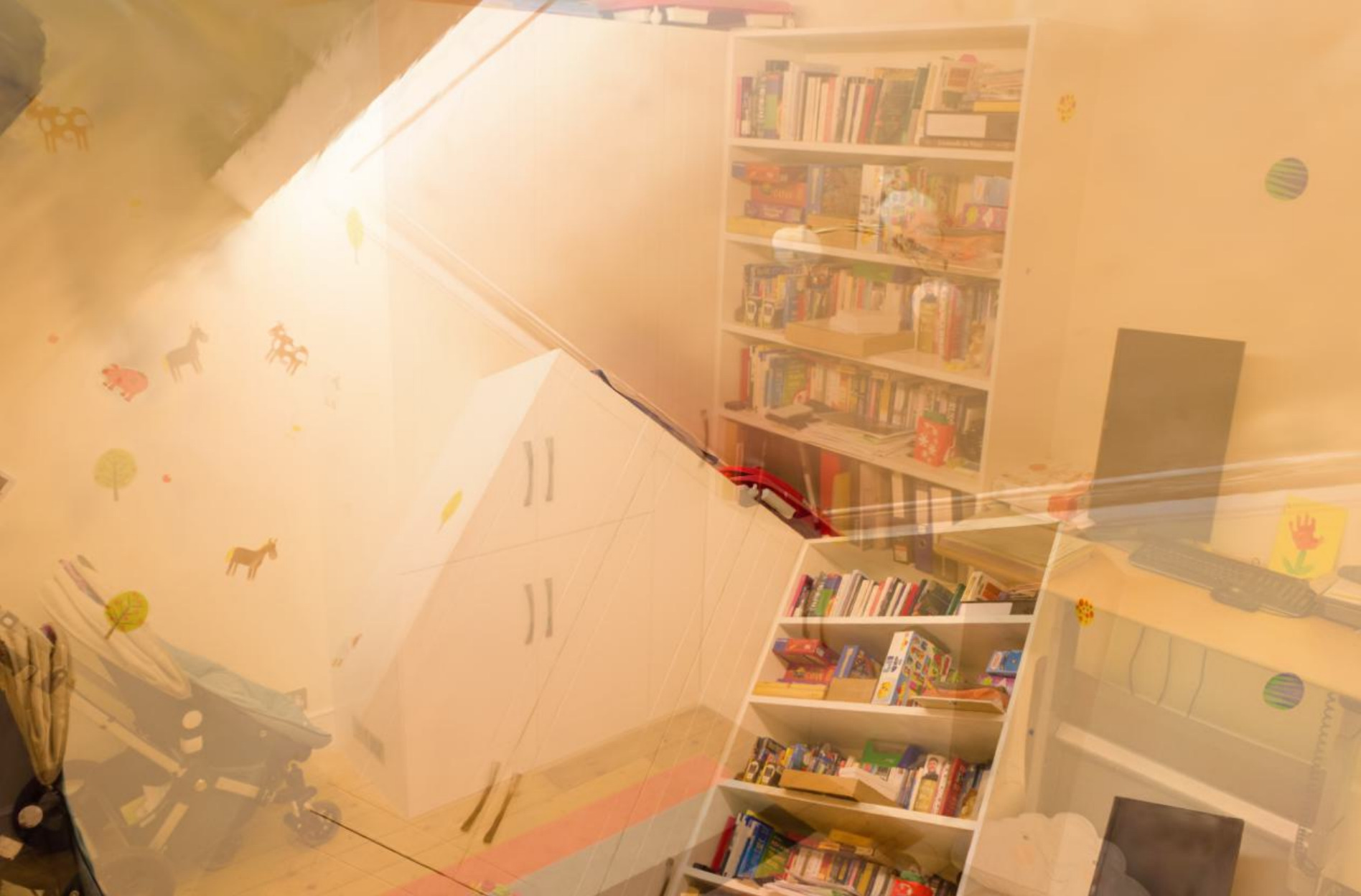} &
        \includegraphics[width=0.18\textwidth]{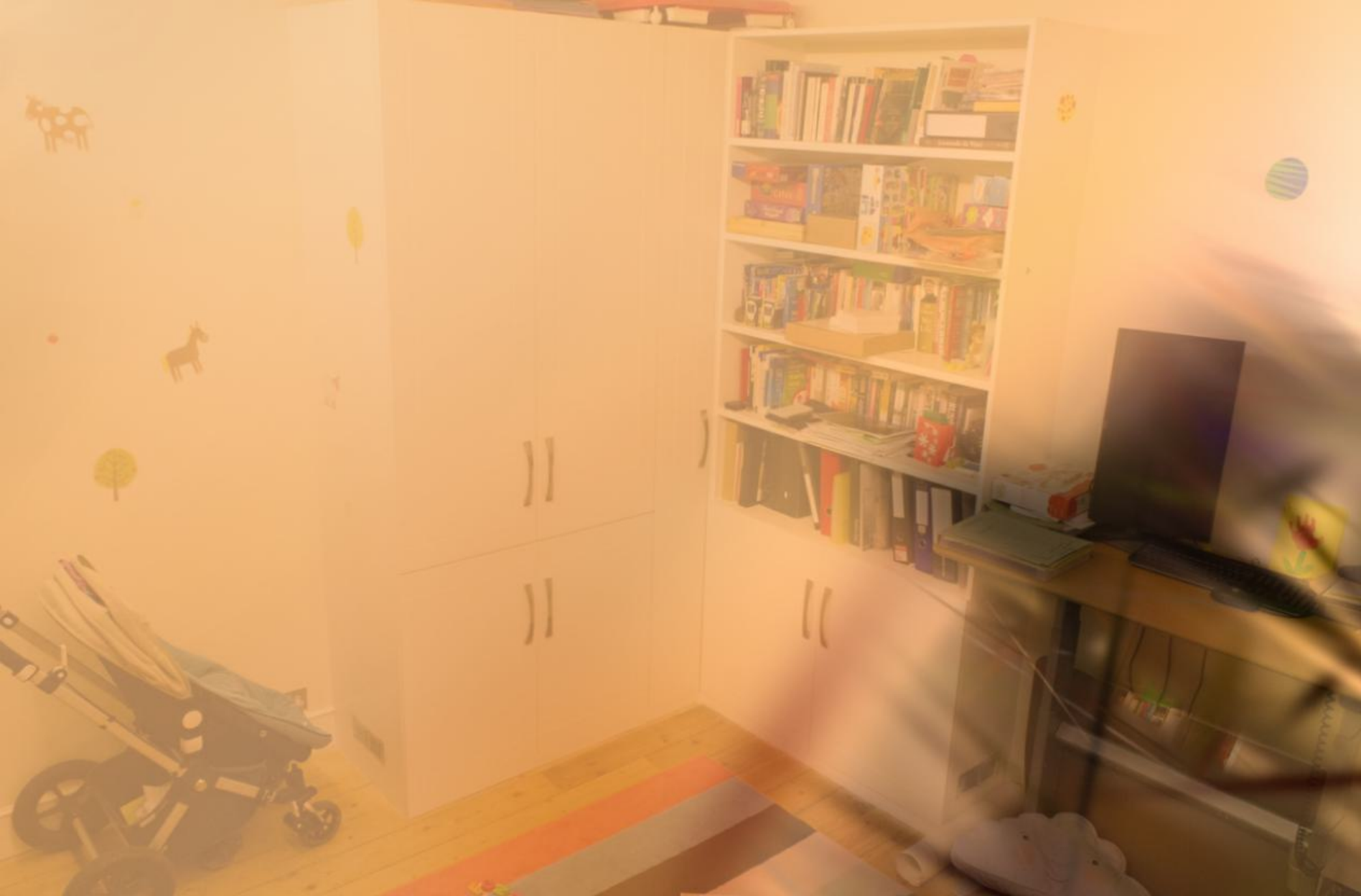} &
        \includegraphics[width=0.18\textwidth]{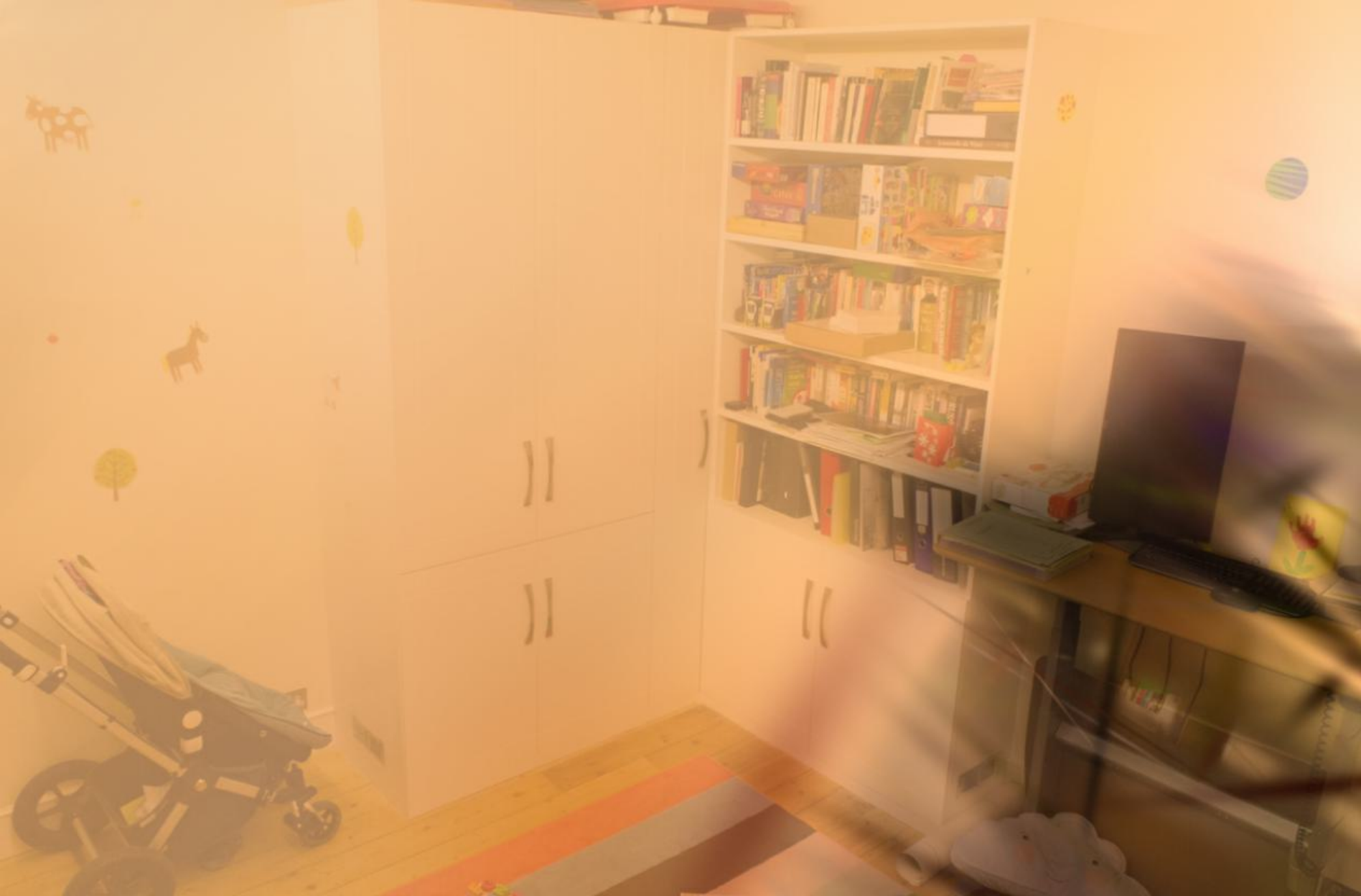} &
        \includegraphics[width=0.18\textwidth]{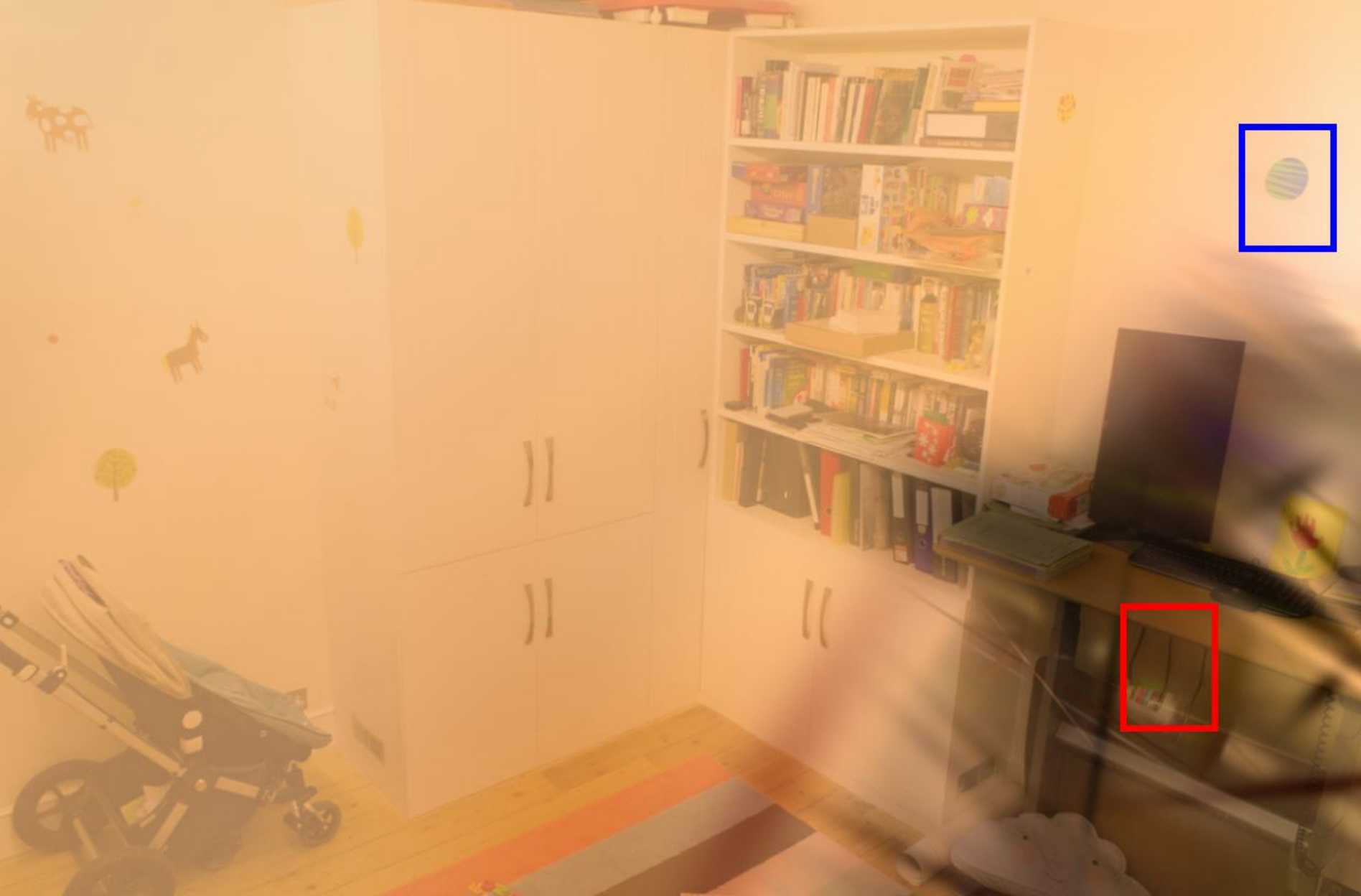} &
        \includegraphics[width=0.18\textwidth]{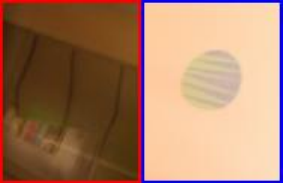} \\
          step=0 & step=200 & step=400 & step=500 &  w/o Matching\\
    \end{tabular}
    }
    \caption{Ablation experiment visualization. The images depict the overlay of the query image and the rendered image, where a higher degree of overlap indicates more accurate pose estimation. The first row presents the results of iComMa, the second row showcases the results of iComMa without comparing, and the third row exhibits the results of iComMa without matching. The last column provides detailed views of the highlighted regions within the boxes in the images of the fourth column.}
    \label{fig_vis_abl}
\end{figure}

Fig. \ref{fig_vis_abl} intuitively illustrates the results of different variants of pose estimation. As shown in the first row of images, iComMa achieves a highly accurate pose estimation result in very few iterations. Furthermore, as demonstrated in the third row, the absence of the matching module fails to cope with large initial pose differences. Lastly, without a comparing strategy, after 200 iterations, a relatively accurate pose is achieved, but subsequent optimization still cannot reach the precision of iComMa.
\section{Conclusion}

In this paper, we propose a novel 6D camera pose estimation method by inverting 3D Gaussian Splatting. This framework integrates traditional geometric matching methods with rendering comparison techniques. A key feature of our approach is the integration of a matching strategy, which significantly enhances the effectiveness in addressing challenging initial states. Additionally, the adoption of render-and-compare method ensures the accuracy of the optimization process in the final stages, which is crucial for downstream applications of pose estimation tasks. The experimental results demonstrate that iComMa effectively balances the robustness and accuracy of camera pose estimation, making it a valuable tool for tackling the challenges of complex pose estimation tasks.
\newpage

%
%
\bibliographystyle{splncs04}
\bibliography{main}
\end{document}